\documentclass[lettersize,journal]{IEEEtran}

\usepackage{amsmath,amssymb,amsfonts,mathrsfs,bbm}
\usepackage{algorithm}
\usepackage[noend]{algpseudocode}
\usepackage{graphicx}
\usepackage{epsfig}
\usepackage{subcaption}
\usepackage{float}
\usepackage{booktabs}
\usepackage{makecell}
\usepackage{threeparttable}
\usepackage{tabularx}
\usepackage{array}
\usepackage{longtable}
\usepackage{multirow}
\usepackage{colortbl}
\usepackage{textcomp}
\usepackage{times}
\usepackage{xspace}
\usepackage{xcolor}
\usepackage{enumerate}
\usepackage{comment}
\usepackage{multicol}
\usepackage{framed}
\usepackage{gensymb}
\usepackage{latexsym}
\usepackage{verbatim}
\usepackage{bbding}

\usepackage{pifont}
\newcommand{\cmark}{\ding{51}} 
\newcommand{\xmark}{\ding{55}} 

\usepackage{cite}
\definecolor{citecolor}{RGB}{119,185,0}
\usepackage[pagebackref,colorlinks,citecolor=citecolor]{hyperref}

\definecolor{light-gray}{gray}{0.9}
\definecolor{cvprblue}{rgb}{0.21,0.49,0.74}
\newcommand{\best}[1]{\textcolor{red}{\textbf{#1}}}
\newcommand{\sbest}[1]{\textcolor{blue}{\textit{#1}}}
\newcommand{\ModelName}{MambaMIL+\xspace}

\begin{document}

\title{\ModelName: Modeling Long-Term Contextual Patterns for Gigapixel Whole Slide Image}
\author{
Qian Zeng\IEEEauthorrefmark{1},
Yihui Wang\IEEEauthorrefmark{1},
Shu Yang,
Yingxue Xu,
Fengtao Zhou,
Jiabo Ma,  
Dejia Cai,\\
Zhengyu Zhang,
Lijuan Qu,
Yu Wang,
Li Liang,
Hao Chen,~\IEEEmembership{Senior Member,~IEEE}%

\thanks{This work was supported by the National Natural Science Foundation of China (No. 62202403), Innovation and Technology Commission (Project No. MHP/002/22 and ITCPD/17-9), Research Grants Council of the Hong Kong Special Administrative Region, China (Project No: T45-401/22-N, R6003-22 and C4024-22GF), Frontier Technology Research for Joint Institutes with Industry Scheme (No. OKT24EG01), and National Key R\&D Program of China (Project No. 2023YFE0204000). (\textit{Corresponding author: Hao Chen.})}
\thanks{Qian Zeng, Yihui Wang, Shu Yang, Yingxue Xu, Fengtao Zhou, Jiabo Ma, and Dejia Cai are with the Department of Computer Science and Engineering, Hong Kong University of Science and Technology, Hong Kong, China
(e-mail: qzengar@connect.ust.hk; ywangrm@connect.ust.hk; syangcw@connect.ust.hk; yxueb@connect.ust.hk; fzhouaf@connect.ust.hk; jmabq@connect.ust.hk; dcaiac@connect.ust.hk).
}
\thanks{Zhengyu Zhang and Li Liang  are with the Department of Pathology, Nanfang Hospital, School of Basic Medical Sciences, Southern Medical University, Guangzhou, China
(e-mail: zzyusmu@163.com; lli@smu.edu.cn). 
}
\thanks{Lijuan Qu is with the 900th Hospital of Joint Logistic Support Force, PLA, Fuzhou, Fujian, China
(e-mail: qljuan6516@sina.com). 
}
\thanks{Yu Wang is with the Department of Pathology, Zhujiang Hospital, Southern Medical University, Guangzhou, China
(e-mail: doctorwylh@163.com). 
}
\thanks{Hao Chen is with the Department of Computer Science and Engineering, the Department of Chemical and Biological Engineering,
the Division of Life Science, the State Key Laboratory of Nervous System Disorders, Hong Kong University of Science and Technology, Hong Kong, China and
HKUST Shenzhen-Hong Kong Collaborative Innovation Research Institute, Futian, Shenzhen, China (e-mail: jhc@cse.ust.hk).}%
\thanks{\IEEEauthorrefmark{1}Equally contributed authors.}
}

\markboth{Journal of \LaTeX\ Class Files,~Vol.~14, No.~8, August~2021}%
{Shell \MakeLowercase{\textit{et al.}}: A Sample Article Using IEEEtran.cls for IEEE Journals}

\maketitle

\begin{abstract}
Whole-slide images (WSIs) are an important data modality in computational pathology, yet their gigapixel resolution and lack of fine-grained annotations challenge conventional deep learning models.
Multiple instance learning (MIL) offers a solution by treating each WSI as a bag of patch-level instances, but effectively modeling ultra-long sequences with rich spatial context remains difficult. 
Recently, Mamba has emerged as a promising alternative for long sequence learning, scaling linearly to thousands of tokens. 
However, despite its efficiency, it still suffers from limited spatial context modeling and memory decay, constraining its effectiveness to WSI analysis. 
To address these limitations, we propose \ModelName, a new MIL framework that explicitly integrates spatial context while maintaining long-range dependency modeling without memory forgetting.
Specifically, \ModelName introduces 
1) \textbf{overlapping scanning}, which restructures the patch sequence to embed spatial continuity and instance correlations; 2) a \textbf{selective stripe position encoder ($\text{S}^{2}\text{PE}$)} that encodes positional information while mitigating the biases of fixed scanning orders; 
and 3) a contextual token selection (CTS) mechanism, which leverages supervisory knowledge to dynamically enlarge the contextual memory for stable long-range modeling. 
Extensive experiments on 20 benchmarks across diagnostic classification, molecular prediction, and survival analysis demonstrate that \ModelName consistently achieves state-of-the-art performance under three feature extractors (ResNet-50, PLIP, and CONCH), highlighting its effectiveness and robustness for large-scale computational pathology.
\end{abstract}

\begin{IEEEkeywords}
Whole Slide Image, Multiple Instance Learning, State Space Model.
\end{IEEEkeywords}

\section{Introduction}

Driven by modern digitization technology, whole slide images (WSIs) have become the foundational data format for automated cancer assessment~\cite{bejnordi2017diagnostic}. 
With the rise of artificial intelligence (AI), pathologists can now leverage off-the-shelf deep learning models to achieve automated and accurate WSI analysis, substantially reducing their workload and accelerating clinical diagnosis~\cite{jiang2023deep,Litjens2017Survey,Komura2018CSBJ}. 
However, the gigapixel-scale resolution of WSIs and the lack of pixel-level annotations pose significant challenges to conventional deep learning frameworks~\cite{huang2022deep,wang2019weakly}.
To address these issues, multiple instance learning (MIL) has emerged as a promising paradigm~\cite{amores2013multiple,shmatko2022artificial}.
In MIL, a WSI is regarded as a ``bag" while the subdivided tissue patches are treated as ``instances''~\cite{amores2013multiple}.
Each bag is assigned a label corresponding to its overall diagnostic outcome, whereas individual instances remain unlabeled, rendering MIL a weakly supervised learning task.

Following the MIL paradigm, a gigapixel WSI is divided into numerous patches, and each patch is transformed into a feature representation using a well-pretrained extractor, such as the ImageNet~\cite{deng2009imagenet} pretrained ResNet-50~\cite{he2016resnet} or foundation models~\cite{uni,huang2023visual,lu2024conch,mstar,gpfm,musk,chief,gigapath,virchow,campanella2025clinical,pathbench}.
This process converts the high-resolution image into an extremely long sequence of low-dimensional features, which is more manageable for MIL methods.
However, many MIL methods often fall short when capturing long-range dependencies among instances.
Although Transformer-based MIL methods introduce the ability to model such long-range interactions, they suffer from quadratic computational cost as the sequence length increases~\cite{shao2021transmil,tang2024rrtmil,vaswani2017attention}.
Moreover, most existing approaches primarily rely on instance-level information under the assumption of i.i.d. (independent and identically distributed) instances, overlooking contextual relationships within WSIs and leaving the actual contribution of spatial context to WSI modeling largely unexplored.

Recently, the emergence of Mamba has provided a promising solution to the aforementioned challenges by enabling efficient gigapixel WSI modeling with linear complexity, leading to notable progress in WSI analysis~\cite{gu2023mamba,dao2024mamba2,yang2024mambamil}. 
However, the vanilla Mamba architecture exhibits two critical limitations when applied to high-resolution WSIs: the lack of spatial context modeling and the problem of memory decay.
Originally proposed for language modeling, Mamba was designed to process 1D sequences and later extended to the vision domain by simply flattening images into token sequences~\cite{zhu2024visionmamba,mambasurvey}. 
While recent efforts introduced 2D scanning mechanisms to better capture spatial continuity, they still suffer from an inductive bias imposed by the predefined scanning order, which may not fully reflect the irregular spatial relationships in WSIs~\cite{zhu2024visionmamba,liu2024vmamba}.
Furthermore, recent studies have revealed that although Mamba is efficient in modeling long sequences, its effective receptive field (ERF) is still limited, leading to inferior performance compared to naive self attention~\cite{ben2024decimamba,ye2025longmamba}.
More importantly, unlike natural images, WSIs exhibit extremely high resolution and complex spatial structures, which further amplify the challenges for Mamba to model long-range dependencies in downstream WSI analysis.

To unlock the potential of Mamba for addressing the aforementioned challenges, we propose \ModelName, a novel framework that effectively models ultra-long WSI sequences while explicitly capturing spatial correlations.

In summary, our contributions are as follows:

\begin{itemize}

\item We provide both qualitative and quantitative evidence that spatial context matters in WSI analysis.
To overcome the spatial-agnostic nature of most MIL frameworks, we introduce an overlapping scanning scheme and a selective stripe position encoder ($\text{S}^{2}\text{PE}$) to explicitly construct and further enhance spatial correlations for WSI modeling.

\item We identify and formalize the exponential memory decay problem in Mamba when modeling ultra-long WSI sequences.
To address this bottleneck, we design a contextual token selection (CTS) mechanism that selectively preserves high-value contextual memory under supervisory guidance for stable long-range contextual modeling. 
Extensive ablation studies validate the superiority of CTS for scalable WSI sequence learning.

\item We conduct comprehensive experiments across 20 tasks, 11 state-of-the-art (SOTA) MIL methods, and 3 feature extractors.
The strong empirical results highlight that Mamba can serve as a powerful and competitive backbone for MIL in WSI analysis.
\end{itemize}

This paper is an extended version of our conference work~\cite{yang2024mambamil}, addressing key limitations of the original approach to WSI analysis. 
We also present extensive additional experiments and insights.

\section{Related Work}

\subsection{Multiple Instance Learning}

Multiple instance learning (MIL) has become a widely adopted paradigm in WSI analysis, as conventional deep learning frameworks struggle to process gigapixel resolution images directly~\cite{amores2013multiple,li2021dsmil,zhang2022dtfd}. 
Broadly, MIL approaches can be divided into two categories: instance-level and embedding-level~\cite{amores2013multiple}.
Instance-level MIL methods aim to assign pseudo labels to individual instances within a slide and train a network accordingly~\cite{kanavati2020weakly,campanella2019clinical,xu2019camel}. 
While intuitive, this strategy typically requires a large number of patches to achieve robust performance, yet only a limited subset of instances is actually used during training, which constrains its effectiveness.
In contrast, embedding-level MIL methods have demonstrated stronger performance and are now the mainstream paradigm~\cite{lu2021clam,ilse2018abmil,li2021dsmil,zhang2022dtfd}. 
In this setting, each patch is first transformed into a fixed-length embedding using a pretrained feature extractor. 
These embeddings are then aggregated into a bag-level representation via a pooling operator, upon which the final slide-level classification is performed. 
Building on this framework, recent Transformer-based MIL models, such as TransMIL~\cite{shao2021transmil} and RRTMIL~\cite{tang2024rrtmil}, further improve performance by leveraging attention mechanisms to capture both global and local correlations among instances.

\subsection{State Space Model}
State space model (SSM) is a sequence model that encodes temporal dynamics through latent states and transition functions, making it well-suited for learning long-range dependencies~\cite{kalman1960ssm}.
Motivated by their strong performance, numerous variants have been proposed~\cite{gu2021s4model,gu2023mamba,dao2024mamba2}.
One successful case is the structured state space model (S4), which can capture long sequences with linear complexity, in contrast to the quadratic complexity of Transformers~\cite{gu2021s4model,vaswani2017attention}. S4 has achieved remarkable progress in diverse domains, including vision and natural language processing~\cite{gu2021efficiently}. 
In computational pathology, S4MIL has further shown that S4 can effectively handle gigapixel WSIs by modeling long sequential patch dependencies with improved efficiency~\cite{fillioux2023s4mil}.
Building upon these advances, the emergence of Mamba has marked another milestone~\cite{gu2023mamba}. 
By introducing a selective state space mechanism with hardware-efficient parallelization, Mamba enables selective information propagation and forgetting, resulting in more effective long-term dependency modeling with linear-time scalability~\cite{gu2023mamba}. 
Recent methods, including MambaMIL~\cite{yang2024mambamil}, PAM~\cite{huang2024pam}, and related variants~\cite{fang2024mammil,zhang20252dmamba}, consistently demonstrate the strength of Mamba in handling high-resolution WSIs, and collectively establish it as a powerful backbone for computational pathology. 
Building on this line, 2DMamba introduces two-dimensional selective scanning to better preserve spatial continuity in gigapixel WSIs, alleviating the limitations of purely 1D token flattening~\cite{zhang20252dmamba}. 
To further enhance efficiency, state space duality (SSD) models extend this line of work by simplifying Mamba’s transition matrix to a single scalar, thereby significantly lowering computational cost while preserving modeling capacity~\cite{dao2024mamba2}.
\section{Method}

This section begins with preliminaries, followed by a detailed description of our method and its key components.

\begin{figure*}[ht]
    \centering
    \includegraphics[width=\linewidth]{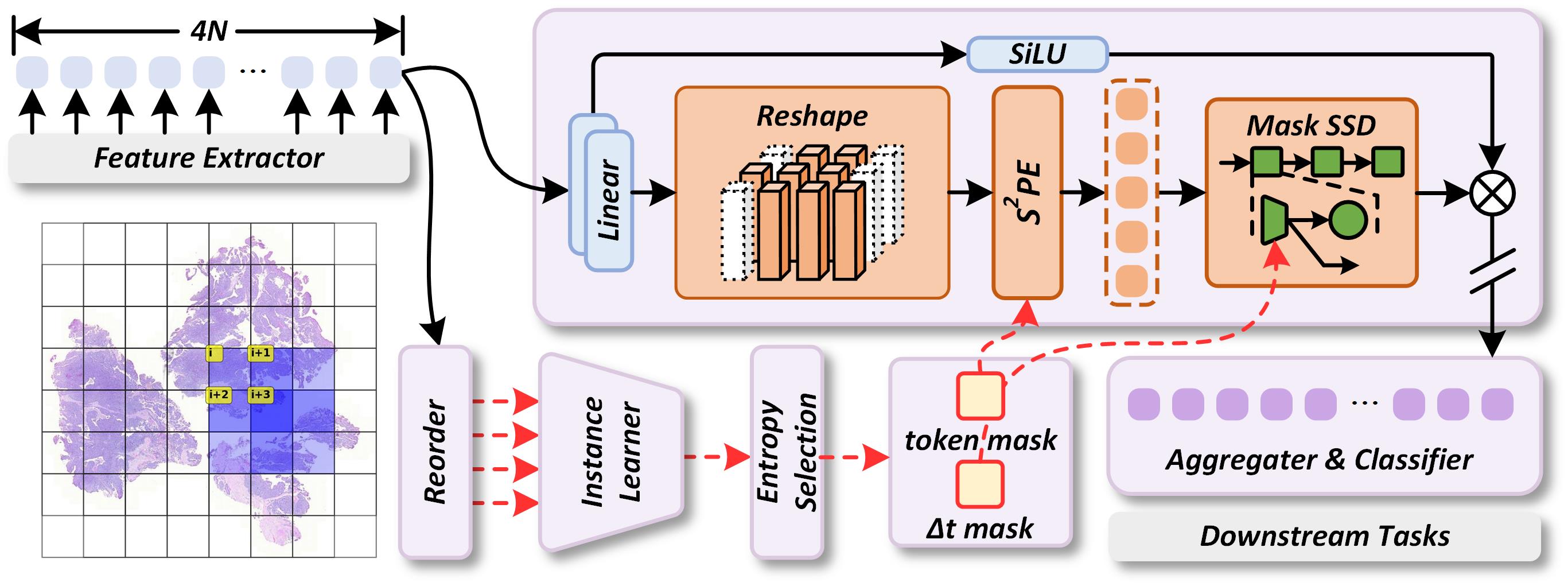}
    \caption{Overview of the proposed \ModelName.
    Overlapping patches are first processed by an offline feature extractor. The resulting overlapping features are reordered and sent to the instance learner to generate instance-level masks. A selective stripe position encoder and contextual token selection are then applied to enhance spatial context modeling and mitigate memory decay in the state space duality model.}
    \label{fig:overview}
\end{figure*}

\subsection{Preliminaries}

\subsubsection{MIL Formulation}

In the MIL paradigm, each WSI $X$ with a corresponding label $Y$ is treated as a \textit{bag}, while the tissue patches obtained after partitioning are regarded as \textit{instances}. 
A bag containing multiple instances is assigned a positive label if it includes at least one positive instance, and a negative label otherwise. 
Formally, the bag with $n$ instances can be represented as
$B=\{(x_1,y_1), \ldots, (x_n,y_n)\},$
where $x_i \in X$ denotes an instance and $y_i \in \{0,1,\ldots,k\}$ denotes its label, with $k$ being the number of classes. The bag label $Y$ is then defined as
\begin{equation}
Y = \begin{cases}
0, & \text{if } \sum y_i = 0, \\
1, & \text{otherwise}.
\end{cases}
\end{equation}
Existing MIL methods embed each instance $x_i$ into a $D$-dimensional feature vector $p_i$ using a pretrained feature extractor $f$, and predict the bag label with a feature aggregator $g$, formulated as:
\begin{equation}
Y = g(p_1, p_2, \ldots, p_n), \quad p_i = f(x_i).
\end{equation}

\subsubsection{State Space Model}

State space models (SSMs) are well-suited for capturing long-range dependencies by maintaining and updating hidden states, in a manner conceptually similar to recurrent neural networks (RNNs). Formally, the continuous-time dynamics can be described as:

\begin{equation}
\begin{cases}
h^\prime(t) = A h(t) + B x(t) \\
y(t) = C h(t),
\end{cases}
\end{equation}
where $A$, $B$, and $C$ denote the continuous system parameters.

In the structured state space model (S4), these continuous parameters are discretized using a timescale parameter $\Delta$, yielding their discrete counterparts $\bar{A}$ and $\bar{B}$:

\begin{equation}
\bar{A} = e^{\Delta A}, \quad 
\bar{B} = (\Delta A)^{-1}\left(e^{\Delta A} - I\right)\cdot \Delta B.
\end{equation}

The final recurrence is expressed as:

\begin{equation}
\begin{cases}
h(t) = \bar{A} h(t-1) + \bar{B} x(t) \\
y(t) = C h(t).
\end{cases}
\label{eq:mamba}
\end{equation}

Building on this discrete formulation, Mamba further enhances the model by introducing a selective propagation mechanism with a hardware-efficient parallel algorithm. 
This design enables effective modeling of long-term dependencies and global context while maintaining linear computational complexity.
In the S4 model, the transition matrix $A$ is constrained to a diagonal form, which allows Mamba to parameterize $A$ as a vector of its diagonal entries rather than a full matrix. 
Based on this idea, the SSD model further simplifies the formulation by reducing $A$ to a single scalar.
This aggressive simplification not only boosts computational efficiency but also makes it easier to couple SSMs with attention mechanisms while still preserving the model’s core structural characteristics.

\begin{figure*}[th]
    \centering
    \includegraphics[width=\linewidth]{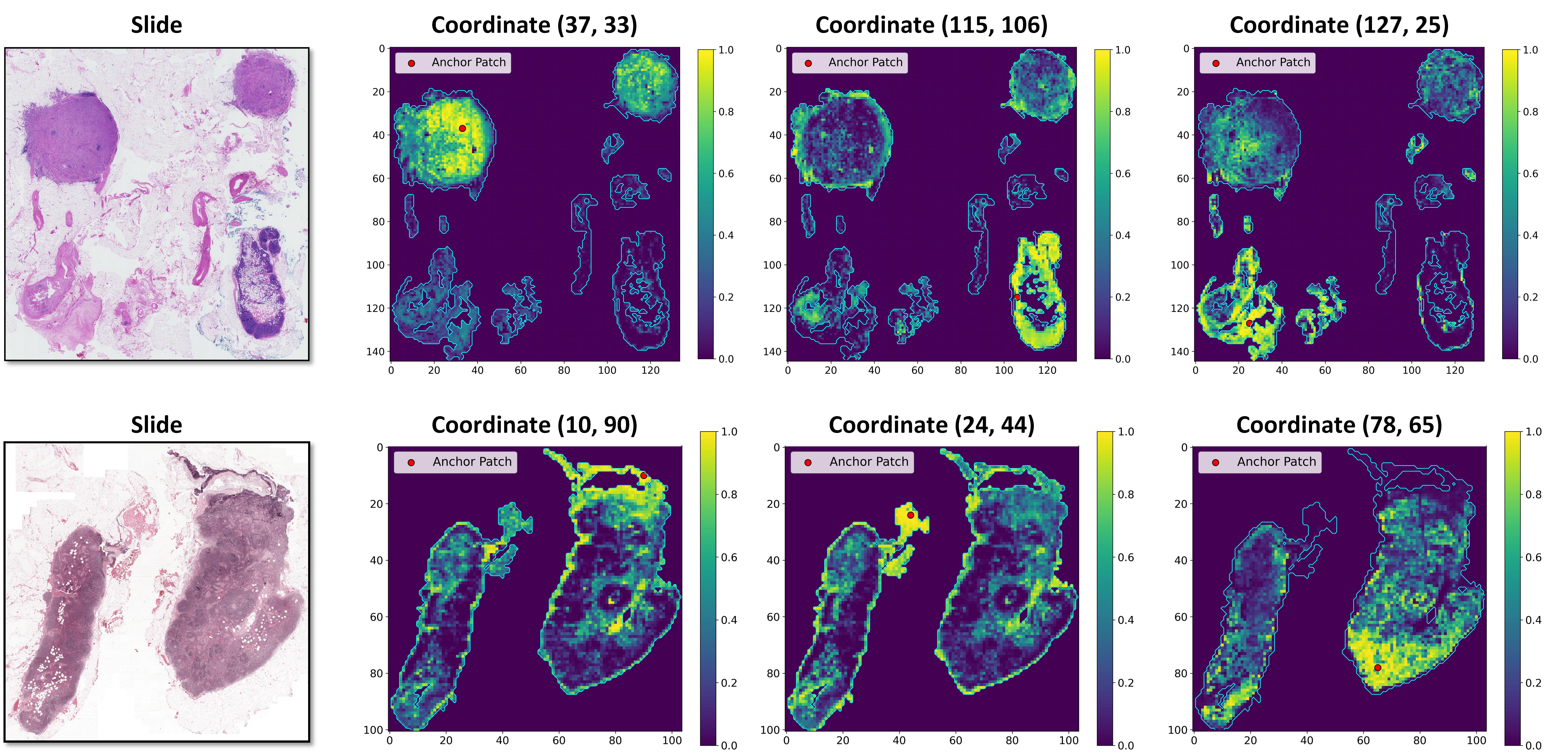}
    \caption{Visualization of the attention map corresponding to randomly selected anchor patches.}
    \label{fig:vis_spatial_map}
\end{figure*}

\begin{table*}[t]
\centering
\scriptsize
\resizebox{\linewidth}{!}{
\setlength{\tabcolsep}{3pt}
\begin{tabular}{@{}c|*{3}{c}|*{3}{c}|*{3}{c}|*{3}{c}|*{3}{c}|*{3}{c}@{}}
\toprule
Dataset & \multicolumn{3}{c|}{BRACS-7} & \multicolumn{3}{c|}{BRCA} & \multicolumn{3}{c|}{Camelyon} & \multicolumn{3}{c|}{NSCLC} & \multicolumn{3}{c|}{UBC-OCEAN} & \multicolumn{3}{c}{AVERAGE} \\
\midrule
Method & AUC & ACC & F1 & AUC & ACC & F1 & AUC & ACC & F1 & AUC & ACC & F1 & AUC & ACC & F1 & AUC & ACC & F1 \\
\midrule
$\mathop{\text{MeanMIL}}\limits_{\text{w/ Overlapping Feature}}$ 
& $\mathop{71.3}\limits_{(+2.1)}$ & $\mathop{23.0}\limits_{(+2.7)}$ & $\mathop{17.5}\limits_{(+3.1)}$ 
& $\mathop{79.0}\limits_{(+5.6)}$ & $\mathop{62.6}\limits_{(+6.4)}$ & $\mathop{64.2}\limits_{(+7.2)}$ 
& $\mathop{70.0}\limits_{(+1.2)}$ & $\mathop{63.3}\limits_{(+0.7)}$ & $\mathop{62.5}\limits_{(+0.8)}$ 
& $\mathop{83.9}\limits_{(+3.4)}$ & $\mathop{77.5}\limits_{(+3.0)}$ & $\mathop{77.5}\limits_{(+3.0)}$
& $\mathop{83.4}\limits_{(+1.6)}$ & $\mathop{42.3}\limits_{(+4.8)}$ & $\mathop{41.2}\limits_{(+5.3)}$ 
& $\mathop{77.5}\limits_{(+2.8)}$ & $\mathop{53.7}\limits_{(+3.5)}$ & $\mathop{52.6}\limits_{(+3.9)}$ \\
\addlinespace
$\mathop{\text{MaxMIL}}\limits_{\text{w/ Overlapping Feature}}$ 
& $\mathop{72.4}\limits_{(+0.8)}$ & $\mathop{23.9}\limits_{(+0.5)}$ & $\mathop{17.0}\limits_{(+1.0)}$ 
& $\mathop{69.2}\limits_{(+4.2)}$ & $\mathop{50.4}\limits_{(+7.6)}$ & $\mathop{45.4}\limits_{(+13.0)}$ 
& $\mathop{81.1}\limits_{(-8.6)}$ & $\mathop{74.1}\limits_{(-6.5)}$ & $\mathop{75.3}\limits_{(-8.6)}$ 
& $\mathop{88.6}\limits_{(+4.1)}$ & $\mathop{80.1}\limits_{(+4.3)}$ & $\mathop{80.0}\limits_{(+4.3)}$ 
& $\mathop{84.9}\limits_{(-0.5)}$ & $\mathop{43.1}\limits_{(+0.1)}$ & $\mathop{41.3}\limits_{(+1.1)}$ 
& $\mathop{79.3}\limits_{(-)}$ & $\mathop{54.3}\limits_{(+1.2)}$ & $\mathop{51.8}\limits_{(+2.1)}$ \\
\addlinespace
$\mathop{\text{CLAM~\cite{lu2021clam}}}\limits_{\text{w/ Overlapping Feature}}$ 
& $\mathop{76.4}\limits_{(+2.0)}$ & $\mathop{29.2}\limits_{(+2.6)}$ & $\mathop{24.5}\limits_{(+3.1)}$ 
& $\mathop{84.1}\limits_{(+1.2)}$ & $\mathop{74.2}\limits_{(-1.9)}$ & $\mathop{74.0}\limits_{(-1.5)}$ 
& $\mathop{89.6}\limits_{(-0.5)}$ & $\mathop{84.1}\limits_{(-0.9)}$ & $\mathop{85.3}\limits_{(-1.5)}$ 
& $\mathop{92.9}\limits_{(+0.3)}$ & $\mathop{86.3}\limits_{(+0.2)}$ & $\mathop{86.3}\limits_{(+0.2)}$
& $\mathop{87.2}\limits_{(+1.2)}$ & $\mathop{56.9}\limits_{(+1.7)}$ & $\mathop{56.2}\limits_{(+2.5)}$ 
& $\mathop{86.0}\limits_{(+0.9)}$ & $\mathop{66.1}\limits_{(+0.3)}$ & $\mathop{65.3}\limits_{(+0.6)}$ \\
\addlinespace
$\mathop{\text{ABMIL~\cite{ilse2018abmil}}}\limits_{\text{w/ Overlapping Feature}}$ 
& $\mathop{74.7}\limits_{(+2.1)}$ & $\mathop{24.7}\limits_{(+0.9)}$ & $\mathop{17.6}\limits_{(+1.4)}$ 
& $\mathop{84.6}\limits_{(+3.7)}$ & $\mathop{72.2}\limits_{(+1.1)}$ & $\mathop{73.3}\limits_{(+1.8)}$ 
& $\mathop{87.9}\limits_{(+0.4)}$ & $\mathop{82.2}\limits_{(+1.8)}$ & $\mathop{83.2}\limits_{(+2.0)}$ 
& $\mathop{92.3}\limits_{(+1.0)}$ & $\mathop{83.3}\limits_{(+2.4)}$ & $\mathop{83.1}\limits_{(+2.5)}$ 
& $\mathop{86.3}\limits_{(+1.6)}$ & $\mathop{51.0}\limits_{(+4.1)}$ & $\mathop{49.3}\limits_{(+5.1)}$ 
& $\mathop{85.2}\limits_{(+1.8)}$ & $\mathop{62.7}\limits_{(+2.1)}$ & $\mathop{61.3}\limits_{(+2.6)}$ \\
\addlinespace
$\mathop{\text{MHIM-ABMIL~\cite{tang2023MHIM}}}\limits_{\text{w/ Overlapping Feature}}$ 
& $\mathop{78.4}\limits_{(+1.0)}$ & $\mathop{29.9}\limits_{(+3.5)}$ & $\mathop{26.2}\limits_{(+3.0)}$ 
& $\mathop{83.7}\limits_{(+4.6)}$ & $\mathop{70.9}\limits_{(-0.5)}$ & $\mathop{72.5}\limits_{(+1.0)}$ 
& $\mathop{89.4}\limits_{(-0.7)}$ & $\mathop{83.5}\limits_{(+1.3)}$ & $\mathop{84.3}\limits_{(+1.3)}$ 
& $\mathop{92.9}\limits_{(+0.5)}$ & $\mathop{85.3}\limits_{(+0.4)}$ & $\mathop{85.1}\limits_{(+0.5)}$ 
& $\mathop{88.6}\limits_{(+2.5)}$ & $\mathop{59.0}\limits_{(+5.3)}$ & $\mathop{60.9}\limits_{(+3.2)}$ 
& $\mathop{86.6}\limits_{(+1.6)}$ & $\mathop{65.7}\limits_{(+2.0)}$ & $\mathop{65.8}\limits_{(+1.8)}$ \\

\addlinespace
$\mathop{\text{DSMIL~\cite{li2021dsmil}}}\limits_{\text{w/ Overlapping Feature}}$ 
& $\mathop{74.0}\limits_{(+1.9)}$ & $\mathop{27.8}\limits_{(-0.8)}$ & $\mathop{22.4}\limits_{(-0.1)}$ 
& $\mathop{84.0}\limits_{(+5.1)}$ & $\mathop{73.9}\limits_{(+3.1)}$ & $\mathop{74.5}\limits_{(+4.3)}$ 
& $\mathop{87.2}\limits_{(+1.8)}$ & $\mathop{78.9}\limits_{(+3.6)}$ & $\mathop{79.7}\limits_{(+2.3)}$ 
& $\mathop{91.1}\limits_{(-0.2)}$ & $\mathop{84.4}\limits_{(-0.9)}$ & $\mathop{83.1}\limits_{(+0.4)}$ 
& $\mathop{86.0}\limits_{(+0.7)}$ & $\mathop{50.9}\limits_{(+3.9)}$ & $\mathop{49.2}\limits_{(+3.6)}$ 
& $\mathop{84.5}\limits_{(+1.8)}$ & $\mathop{63.2}\limits_{(+1.8)}$ & $\mathop{61.8}\limits_{(+2.1)}$ \\

\addlinespace
$\mathop{\text{DTFD~\cite{zhang2022dtfd}}}\limits_{\text{w/ Overlapping Feature}}$ 
& $\mathop{77.5}\limits_{(-0.3)}$ & $\mathop{27.4}\limits_{(-1.3)}$ & $\mathop{22.0}\limits_{(-1.7)}$ 
& $\mathop{83.7}\limits_{(+3.4)}$ & $\mathop{67.1}\limits_{(+7.2)}$ & $\mathop{69.3}\limits_{(+7.3)}$ 
& $\mathop{87.9}\limits_{(-0.3)}$ & $\mathop{79.4}\limits_{(-0.4)}$ & $\mathop{80.4}\limits_{(+1.0)}$ 
& $\mathop{93.7}\limits_{(+0.2)}$ & $\mathop{86.4}\limits_{(-0.4)}$ & $\mathop{86.3}\limits_{(-0.3)}$ 
& $\mathop{86.5}\limits_{(+2.8)}$ & $\mathop{50.3}\limits_{(+6.4)}$ & $\mathop{49.5}\limits_{(+6.8)}$ 
& $\mathop{85.8}\limits_{(+1.2)}$ & $\mathop{62.1}\limits_{(+2.3)}$ & $\mathop{61.5}\limits_{(+2.6)}$ \\

\addlinespace

$\mathop{\text{TransMIL~\cite{shao2021transmil}}}\limits_{\text{w/ Overlapping Feature}}$ 
& $\mathop{74.6}\limits_{(-1.3)}$ 
& $\mathop{29.5}\limits_{(-1.1)}$ 
& $\mathop{28.1}\limits_{(-1.7)}$ 
& $\mathop{82.8}\limits_{(+0.1)}$ 
& $\mathop{69.5}\limits_{(-1.7)}$ 
& $\mathop{70.9}\limits_{(-1.6)}$ 
& $\mathop{89.0}\limits_{(-2.9)}$ 
& $\mathop{82.5}\limits_{(-3.8)}$ 
& $\mathop{82.0}\limits_{(-2.8)}$ 
& $\mathop{91.5}\limits_{(+1.5)}$ 
& $\mathop{80.3}\limits_{(+2.9)}$ 
& $\mathop{79.7}\limits_{(+3.1)}$ 
& $\mathop{87.9}\limits_{(-0.9)}$ 
& $\mathop{61.9}\limits_{(-6.2)}$ 
& $\mathop{61.7}\limits_{(-5.7)}$ 
& $\mathop{85.2}\limits_{(-0.7)}$ 
& $\mathop{64.8}\limits_{(-2.0)}$ 
& $\mathop{64.5}\limits_{(-1.8)}$ \\

\addlinespace
$\mathop{\text{MHIM-TransMIL~\cite{tang2023MHIM}}}\limits_{\text{w/ Overlapping Feature}}$ 
& $\mathop{76.6}\limits_{(-0.5)}$ 
& $\mathop{33.2}\limits_{(+0.8)}$ 
& $\mathop{30.4}\limits_{(+0.8)}$ 
& $\mathop{84.5}\limits_{(+2.6)}$ 
& $\mathop{71.4}\limits_{(-4.3)}$ 
& $\mathop{73.2}\limits_{(-3.1)}$ 
& $\mathop{90.0}\limits_{(-1.2)}$ 
& $\mathop{80.9}\limits_{(-2.1)}$ 
& $\mathop{79.7}\limits_{(-2.1)}$ 
& $\mathop{93.4}\limits_{(-0.3)}$ 
& $\mathop{84.1}\limits_{(-2.9)}$ 
& $\mathop{83.8}\limits_{(-3.5)}$ 
& $\mathop{89.0}\limits_{(+0.2)}$ 
& $\mathop{58.8}\limits_{(-0.8)}$ 
& $\mathop{60.0}\limits_{(-2.4)}$ 
& $\mathop{86.7}\limits_{(+0.2)}$ 
& $\mathop{65.7}\limits_{(-1.9)}$ 
& $\mathop{65.4}\limits_{(-2.1)}$ \\

\addlinespace
$\mathop{\text{RRTMIL~\cite{tang2024rrtmil}}}\limits_{\text{w/ Overlapping Feature}}$ 
& $\mathop{78.7}\limits_{(+0.4)}$ 
& $\mathop{33.8}\limits_{(+1.0)}$ 
& $\mathop{29.3}\limits_{(+2.4)}$ 
& $\mathop{86.4}\limits_{(+0.3)}$ 
& $\mathop{72.7}\limits_{(-)}$ 
& $\mathop{74.0}\limits_{(+0.8)}$ 
& $\mathop{88.8}\limits_{(+1.2)}$ 
& $\mathop{83.2}\limits_{(-0.4)}$ 
& $\mathop{84.1}\limits_{(-1.1)}$ 
& $\mathop{93.4}\limits_{(+0.7)}$ 
& $\mathop{84.4}\limits_{(+1.9)}$ 
& $\mathop{84.2}\limits_{(+1.9)}$ 
& $\mathop{91.5}\limits_{(-0.5)}$ 
& $\mathop{68.1}\limits_{(+0.7)}$ 
& $\mathop{69.6}\limits_{(+0.3)}$ 
& $\mathop{87.7}\limits_{(+0.4)}$ 
& $\mathop{68.4}\limits_{(+0.6)}$ 
& $\mathop{68.2}\limits_{(+0.9)}$ \\
\addlinespace
$\mathop{\text{PAM~\cite{huang2024pam}}}\limits_{\text{w/ Overlapping Feature}}$ 
& $\mathop{80.1}\limits_{(-0.7)}$ 
& $\mathop{40.7}\limits_{(-2.1)}$ 
& $\mathop{39.9}\limits_{(-2.9)}$ 
& $\mathop{85.0}\limits_{(+4.5)}$
& $\mathop{72.6}\limits_{(+4.2)}$ 
& $\mathop{74.8}\limits_{(+3.1)}$ 
& $\mathop{88.4}\limits_{(+0.8)}$ 
& $\mathop{78.3}\limits_{(+2.0)}$ 
& $\mathop{76.7}\limits_{(+1.8)}$ 
& $\mathop{93.8}\limits_{(+0.6)}$ 
& $\mathop{84.3}\limits_{(+2.5)}$ 
& $\mathop{83.9}\limits_{(+2.7)}$
& $\mathop{91.7}\limits_{(+0.5)}$ 
& $\mathop{66.8}\limits_{(+4.9)}$ 
& $\mathop{66.9}\limits_{(+4.3)}$ 
& $\mathop{87.8}\limits_{(+1.2)}$ 
& $\mathop{68.5}\limits_{(+2.3)}$ 
& $\mathop{68.4}\limits_{(+1.8)}$ \\

\addlinespace
$\mathop{\text{MambaMIL~\cite{yang2024mambamil}}}\limits_{\text{w/ Overlapping Feature}}$ 
& $\mathop{79.6}\limits_{(+0.5)}$ & $\mathop{38.4}\limits_{(+1.1)}$ & $\mathop{37.7}\limits_{(+0.4)}$ 
& $\mathop{85.2}\limits_{(+3.9)}$ & $\mathop{74.0}\limits_{(+1.8)}$ & $\mathop{73.7}\limits_{(+4.0)}$ 
& $\mathop{90.1}\limits_{(+0.1)}$ & $\mathop{84.6}\limits_{(+0.2)}$ & $\mathop{85.5}\limits_{(-0.2)}$ 
& $\mathop{94.1}\limits_{(-0.1)}$ & $\mathop{86.9}\limits_{(-0.6)}$ & $\mathop{86.6}\limits_{(-0.5)}$ 
& $\mathop{92.1}\limits_{(+0.7)}$ & $\mathop{68.9}\limits_{(+0.7)}$ & $\mathop{70.7}\limits_{(-0.7)}$ 
& $\mathop{88.2}\limits_{(+1.0)}$ & $\mathop{70.6}\limits_{(+0.6)}$ & $\mathop{70.8}\limits_{(+0.6)}$ \\

\addlinespace
$\mathop{\text{2DMambaMIL~\cite{zhang20252dmamba}}}\limits_{\text{w/ Overlapping Feature}}$ 
& $\mathop{81.1}\limits_{(+1.3)}$ & 
$\mathop{41.4}\limits_{(+2.0)}$ & 
$\mathop{40.6}\limits_{(+1.5)}$ & 
$\mathop{85.2}\limits_{(+2.6)}$ & 
$\mathop{67.4}\limits_{(+6.7)}$ & 
$\mathop{69.6}\limits_{(+6.5)}$ & 
$\mathop{88.7}\limits_{(-2.6)}$ & 
$\mathop{84.5}\limits_{(-1.1)}$ & 
$\mathop{85.3}\limits_{(-0.7)}$ & 
$\mathop{94.1}\limits_{(-0.1)}$ & 
$\mathop{87.3}\limits_{(-0.8)}$ & 
$\mathop{87.2}\limits_{(-0.8)}$ & 
$\mathop{92.8}\limits_{(+0.1)}$ & 
$\mathop{69.0}\limits_{(-1.3)}$ & 
$\mathop{69.8}\limits_{(-1.1)}$ & 
$\mathop{88.4}\limits_{(+0.2)}$ &
$\mathop{69.9}\limits_{(+1.1)}$ & 
$\mathop{70.5}\limits_{(+1.1)}$ \\

\bottomrule
\end{tabular}
}
\caption{Performance improvements of different baselines with overlapping features across diagnostic classification datasets.}
\label{tab:overlap_improvement}
\end{table*}

\subsection{Spatial Context Matters in WSIs}
Spatial context plays a crucial role in understanding the content of a natural image~\cite{wei2021fine,liu2024vmamba,zhu2024visionmamba}; however, for WSIs, whether modeling spatial context is beneficial remains unclear, especially when the two-stage MIL pipeline further leads to feature discontinuity between instances~\cite{gadermayr2024multiple}.
Recent studies suggest that incorporating a correlation assumption among instances can provide more useful information for the MIL problem compared to the i.i.d. (independent and identically distributed) assumption~\cite{shao2021transmil}, but the benefits of modeling spatial dependencies have not been well demonstrated.

A simple way to quantify the correlation between instances in the feature space is by computing the attention score. To do this, we randomly select an anchor patch $x^{anchor}$ from the tissue foreground and calculate the attention score between the features corresponding to the anchor patch and the other features, as given by:
\begin{equation}
    a^{anchor}_{i} = \|p^{anchor}\| \cdot \|p_{i}\|^{T},
\end{equation}
where $\|\cdot\|$ denotes the L2-norm of the feature vectors. 
As shown in Fig.~\ref{fig:vis_spatial_map}, the attention maps corresponding to different selected anchor patches are visualized. 
The cyan contour denotes the foreground of the WSI, and the red points represent the selected anchor patches. 
It is evident that the attention map is highly localized, with patches spatially closer to the anchor exhibiting higher attention. 
Clearly, despite the two-stage MIL paradigm substantially attenuating spatial cues in the embedding space, the extracted features still preserve meaningful spatial patterns to a large extent.
Any perturbation or neglect of spatial order can result in the loss of this information.

Given that spatial context is preserved within instances, effectively leveraging this information becomes crucial.
Most mainstream MIL methods partition WSIs into non-overlapping patches, which are then passed through a feature extractor to generate embeddings for the subsequent aggregator: $\mathcal{P}=\left\{\boldsymbol{p}_0, \boldsymbol{p}_1, \ldots, \boldsymbol{p}_{N-1}\right\} \in \mathbb{R}^{N \times D}$, where $\boldsymbol{p}_i \in \mathbb{R}^D$ denotes an instance feature, $N$ and $D$ denote the length of the sequence and feature dimension. 
However, the i.i.d. assumption~\cite{ilse2018abmil,lu2021clam,li2021dsmil,zhang2022dtfd}, sequential modeling~\cite{huang2024pam} or perturbation operations~\cite{shao2021transmil,yang2024mambamil} in existing MIL methods, often weaken or even disregard the spatial relationships between instances. 
Therefore, we explore the use of overlapping features to explicitly incorporate spatial context between neighboring patches and address the feature discontinuity issue inherent in the two-stage MIL paradigm.

Tab.~\ref{tab:overlap_improvement} presents the improvement gained by applying overlapping features to various MIL methods. 
Notably, the results highlight the significant benefit of incorporating explicit spatial context into the feature space across almost all methods. 
For methods based on the i.i.d. assumption, overlapping features introduce spatial correlation between instances. 
Meanwhile, for methods that follow correlation assumption, the overlapping features further enhance performance by reducing the side effects from sequential modeling or perturbation operations.
This underscores the critical role of spatial context in effectively modeling WSIs.

\subsection{Exponential Memory Decay in Mamba}
\label{sec:exponential memory decay}
Modeling WSIs often involves extremely long sequences, requiring the model to capture long-range dependencies. 
Although Mamba has been shown to be a strong backbone for WSI analysis~\cite{yang2024mambamil,huang2024pam,zhang20252dmamba}, whether it can effectively capture long-range dependencies in such ultra-large images remains unexplored.
To analyze the long-context modeling capability of Mamba, we revisit the hidden state updates when processing long sequences. 
Given a WSI with $N$ patches, we obtain the token sequence:
$\mathcal{X} = \{\boldsymbol{x}_0, \boldsymbol{x}_1, \dots , \boldsymbol{x}_{N-1}\}$.
Following Equation~\ref{eq:mamba}, the hidden state of the $i$-th token can be reformulated as a weighted accumulation of all previous tokens:
\begin{equation}
h_i = \sum_{j=1}^i \left( \prod_{k=j+1}^i \bar{A}_k \right) \bar{B}_j x_j,
\label{eq:recursive_mamba_eq}
\end{equation}
where $h_i$ and $x_i$ denote the $i$-th hidden state and token embedding. Since $\bar{A}$ is discretized through a timescale parameter $\Delta$, the product term is defined as:
\begin{equation}
\prod_{k=j+1}^{i} \bar{A}_k 
= \prod_{k=j+1}^{i} \exp(A \Delta_k) 
= \exp\!\left( A \sum_{k=j+1}^{i} \Delta_k \right),
\label{eq:decay}
\end{equation}
where $A<0$ and $\Delta>0$, meaning that the contribution of previous tokens exponentially decays toward zero as the sequence extends.
This behavior is formally defined as \textit{exponential memory decay} in Mamba, whereby earlier tokens have progressively diminishing impact in sequences.

In the context of WSIs, the sequences can span tens of thousands of tokens, a length that often leads to memory collapse. 
As the sequence grows, the model struggles to preserve critical contextual information due to exponential memory decay, causing earlier yet important tokens to lose their influence on the hidden state. 
Therefore, effectively capturing and preserving this information is essential to ensure that the hidden state retains the contextual details for long-range modeling.



\subsection{Overview of \ModelName}
Building on our analysis, we introduce \ModelName, a MIL framework designed to effectively model long sequences in WSIs by incorporating spatial context and alleviating the issue of exponential memory decay in Mamba.
As shown in Fig.~\ref{fig:overview}, \ModelName applies overlapping scanning to prevent spatial discontinuity, incorporates a selective stripe position encoder ($\text{S}^2\text{PE}$) for efficient position encoding, and utilizes contextual token selection to prioritize informative tokens by leveraging supervisory knowledge.
Details are explained below. 

\subsubsection{Overlapping Scanning}
\begin{figure}[ht]
    \centering
    \includegraphics[width=0.9\linewidth]{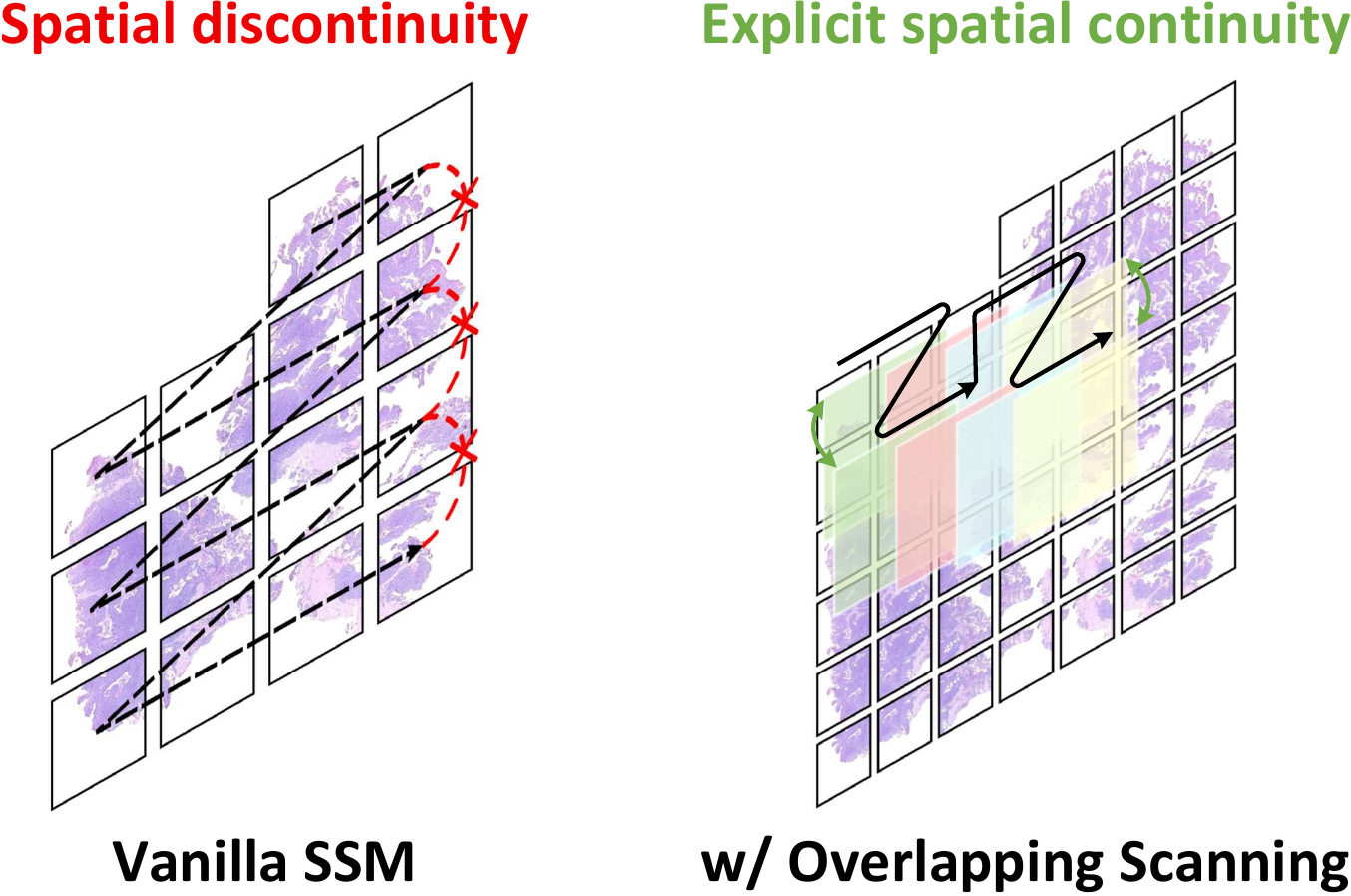}
    \caption{Comparison of different designs: (1) Vanilla SSM flattening images into 1D sequences; (2) SSM with overlapping scanning preserving spatial continuity across patches.}
    \label{fig:overlap_scanning}
\end{figure}

In our preliminary experiments, overlapping features were found to be beneficial, as they explicitly encode spatial relationships between adjacent patches. 
Therefore, we also adopt this design in our model. 
More importantly, Mamba is particularly compatible with overlapping features, since its linear complexity allows it to scale in sequence length with substantially lower computational overhead. 
We refer to this scanning mode as overlapping scanning, where the feature sequence $\mathcal{P}=\left\{\boldsymbol{p}_0, \boldsymbol{p}_1, \ldots, \boldsymbol{p}_{4N-1}\right\} \in \mathbb{R}^{4N \times D}$ is constructed with overlapping patches and scanned accordingly.
As shown in Fig.~\ref{fig:overlap_scanning}, unlike the vanilla row-wise scan, overlapping scanning maintains spatial continuity along the vertical axis, producing a more structural sequential representation.
Even though overlapping scanning still adopts a 1D scanning scheme, it proves unexpectedly competitive. Specifically, our preliminary results (see Tab.~\ref{tab:overlap_improvement}) show that 1D scanning performs on par with 2D scanning—and becomes even stronger once overlapping features are introduced. 
To uncover the underlying reasons, we theoretically revisit both scanning strategies:
we first give the 2D scanning hidden state formulation and then reform the 1D scanning recurrence on the 2D input via a row-major linearization.

\begin{figure*}[ht]
    \centering
    \includegraphics[width=\linewidth]{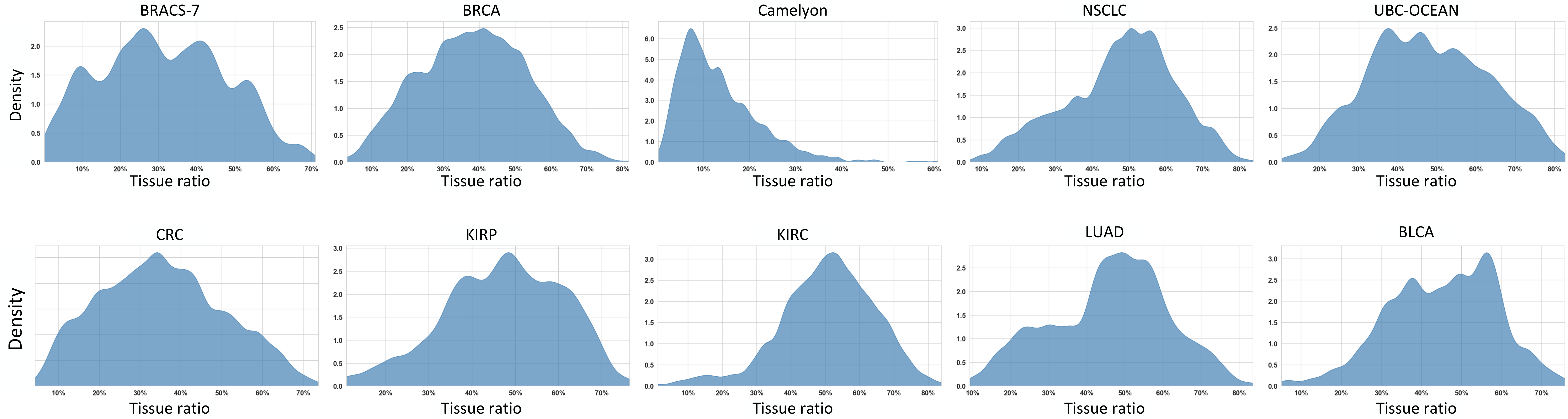}
    \caption{Tissue ratio distributions in the reconstructed two-dimensional map across different datasets.}
    \label{fig:MambaMIL_Tissue_Ratio_Statistic}
\end{figure*}

\paragraph{2D Scanning}
For a 2D input feature map, the hidden state at location \((i,j)\) is
\begin{equation}
\label{eq:2dmamba}
h_{i,j} =
\sum_{u=1}^{i} \sum_{v=1}^{j}
\Big[
\Phi(u,v;i,j)\, \bar{B}_{u,v}\, x_{u,v}
\Big],
\end{equation}
with the 2D transition kernel:
\begin{equation}
\Phi(u,v;i,j) \triangleq
\Big(\prod_{p=u+1}^{i} \bar{A}_{p}\Big)
\Big(\prod_{q=v+1}^{j} \bar{A}_{q}\Big),
\end{equation}
where \(\bar{A}_{p}\) and \(\bar{A}_{q}\) are state-transition parameters along horizontal and vertical directions, respectively; \(\bar{B}_{u,v}\) projects the input \(x_{u,v}\) at \((u,v)\).
Empty products are defined to be the identity.

\paragraph{Vanilla 1D Scanning}
Following Eq.~\ref{eq:mamba}, we map \((i,j)\) to a 1D time index by the row-major linearization:
\begin{equation}
    \ell(i,j) \triangleq (i-1)W + j \in \{1,\dots, HW\},
\end{equation}
where \(H\) and \(W\) are the map height and width. Denote \(\tilde{A}_t\triangleq \bar{A}_{\ell^{-1}(t)}\), \(\tilde{B}_t\triangleq \bar{B}_{\ell^{-1}(t)}\), and \(\tilde{x}_t\triangleq x_{\ell^{-1}(t)}\), the 1D Mamba recurrence on the 2D map then reads:
\begin{equation}
\label{eq:1d-on-2d-compact}
h_{i,j} \;=\;
\sum_{t=1}^{\ell(i,j)}
\Bigg(
\prod_{k=t+1}^{\ell(i,j)} \tilde{A}_{k}
\Bigg)\,
\tilde{B}_{t}\,\tilde{x}_{t}.
\end{equation}
Equivalently, splitting the predecessors into all previous rows and the current row, we obtain:
\begin{equation}
\label{eq:1d-on-2d-split}
\begin{aligned}
h_{i,j}
&=
\underbrace{\sum_{u=1}^{i-1}\sum_{v=1}^{W}
\Bigg(\prod_{k=\ell(u,v)+1}^{\ell(i,j)} \tilde{A}_{k}\Bigg)
\bar{B}_{u,v}\,x_{u,v}}_{\text{contributions from rows }1\ldots i-1}
\\
&\quad+\;
\underbrace{\sum_{v=1}^{j}
\Bigg(\prod_{k=\ell(i,v)+1}^{\ell(i,j)} \tilde{A}_{k}\Bigg)
\bar{B}_{i,v}\,x_{i,v}}_{\text{contributions within row }i,\ \text{cols }1\ldots j}\!
\end{aligned}
\end{equation}
Equations~\eqref{eq:2dmamba}–\eqref{eq:1d-on-2d-split} show that, although 2D scanning allows spatially adjacent tokens to be processed in a closer manner, it still suffers from the sequential bias and actually reduces the number of accessible preceding tokens compared to vanilla 1D Mamba.
Additionally, 2D scanning scheme requires the patch sequence to be reconstructed into a rectangular grid, which inevitably introduces large blank regions.
To quantify this effect, we measure the tissue-token ratio within the reconstructed 2D map. 
Notably, a considerable number of WSIs exhibit ratios below 50\%, with some dropping to less than 10\%, as shown in Fig.~\ref{fig:MambaMIL_Tissue_Ratio_Statistic}.
These limitations are further intensified in WSIs, where extremely long sequences and large non-tissue regions accelerate memory decay and erode contextual information.
Consequently, overlapping scanning adopts a 1D scanning scheme—one that, when introducing explicit spatial context, can outperform 2D scanning in practice.

\begin{figure}[ht]
    \centering
    \includegraphics[width=\linewidth]{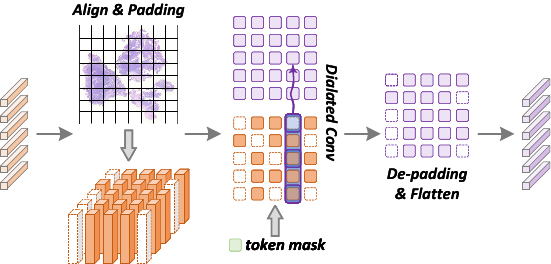}
    \caption{Overview of the selective stripe position encoder.}
    \label{fig:position_encoder}
\end{figure}

\subsubsection{Selective Stripe Position Encoder}
Positional encoding (PE) plays a crucial role in MIL frameworks as WSIs are typically flattened into sequential embeddings that discard their original spatial layout.
Architectures such as Transformers, by default, cannot effectively capture positional relations without additional encoding. 
Similarly, although Mamba processes WSIs in a streaming manner, it was originally designed for language modeling, where tokens naturally follow a one-dimensional order. 
Thus, explicitly incorporating positional encoding is a natural extension to enrich the representation with two-dimensional spatial structure, bridging the gap between sequential modeling and the intrinsic topology of histopathology images.

Inspired by the pyramid position encoding generator (PPEG)~\cite{shao2021transmil}, we extend the use of convolutional kernels to dynamically generate position encodings that adapt to the local neighborhood of tissue tokens.
To align with our architectural design, we introduce the Selective Stripe Position Encoder (S$^2$PE), as shown in Fig.~\ref{fig:position_encoder}. 
In this module, sequential tokens are padded and reshaped into a two-dimensional map first, thereby preserving the intrinsic spatial arrangement. 
Before applying positional encoding, irrelevant tokens are masked to prevent spurious contextual interactions—a detailed description of the masking strategy will be provided in the subsequent section.
Because Mamba processes the image sequentially in a row-by-row manner, it inherently models horizontal dependencies. 
We therefore apply additional positional encoding only along the vertical axis to avoid redundant or conflicting cues across spatial dimensions.
To further accommodate overlapping patches and enhance positional awareness, S$^2$PE employs a lightweight one-dimensional dilated convolution, which efficiently captures broader spatial and contextual contexts without excessive computational cost. 
The S$^2$PE can be formulated as:
\begin{equation}
\begin{aligned}
    x'_i &= \mathcal{T}^{-1}\!\Bigg( 
        \operatorname{DilatedConv}_{1\mathrm{D}} \Big( 
            \mathcal{T}\big( [M] \odot x_i \big) 
        \Big) \Bigg), \\
    \mathcal{T}&: \mathbb{R}^{4N \times D} \;\to\; \mathbb{R}^{H \times W \times D},
\end{aligned}
\end{equation}
where $x_i \in \mathbb{R}^{D}$ denotes the representation of the $i$-th token,  
$x'_i$ is its updated representation after positional encoding,  
$[M] \in \{0,1\}^{4N}$ is a binary mask that filters out irrelevant tokens,  
$\odot$ denotes element-wise multiplication,  
$\mathcal{T}$ is the transformation that reshapes a 1D sequence into a 2D feature map of size $H \times W$, the shape of the WSI,  
$\mathcal{T}^{-1}$ is the inverse transformation that flattens the map back into a sequence,  
and $\operatorname{DilatedConv}_{1\mathrm{D}}$ represents a one-dimensional dilated convolution.

\subsubsection{Contextual Token Selection}
While overlapping scanning enhances the spatial and contextual continuity between tokens, it also increases the sequence length by four times. 
Mamba, though effective in modeling long sequences with much lower computational cost, becomes limited when handling ultra-long sequences due to the exponential memory decay problem (discussed in Sec.~\ref{sec:exponential memory decay}), especially in WSIs where regions of interest may be spatially distant. 

\begin{table}[htbp]
\centering
\scriptsize
\resizebox{\linewidth}{!}{
\setlength{\tabcolsep}{3pt}
\begin{tabular}{@{}l|*{3}{c}|*{3}{c}|*{3}{c}@{}}
\toprule
Dataset & \multicolumn{3}{c}{BRACS-7} & \multicolumn{3}{|c}{BRCA} & \multicolumn{3}{|c}{Camelyon}\\
\midrule
Metric & AUC & ACC & F1 & AUC & ACC & F1 & AUC & ACC & F1\\
\midrule
MambaMIL & 79.6 & 38.4 & 37.7 & 85.2 & 74.0 & 73.7 & 90.1 & 84.6 & 85.5 \\
\rowcolor{gray!20} w/ MHIM & 80.5 & 40.7 & 40.9 & 85.4 & 73.6 & 76.5 & 90.3 & 84.7 & 84.9  \\
\midrule
MambaMIL$^\star$ & 80.1 & 39.6 & 38.0 & 89.0 & 75.8 & 77.7 & 90.2 & 84.8 & 85.3 \\
\rowcolor{gray!20} w/ MHIM & 79.7 & 37.6 & 36.5 & 89.1 & 77.2 & 78.6 & 88.9 & 84.6 & 85.6  \\
\bottomrule
\end{tabular}}
\caption{Performance comparison of MambaMIL~\cite{yang2024mambamil} and MHIM-MambaMIL~\cite{tang2023MHIM} using ResNet-50 as the feature extractor. The star symbol ($^\star$) denotes the variant with the overlapping scanning scheme.}
\label{tab:comparison_MHIM}
\end{table}

A straightforward approach for extending memory is to prune specific tokens based on their attention scores, similar to MHIM-MIL~\cite{tang2023MHIM}, which essentially performs hard instance mining. 
However, transferring this method to Mamba may also lead to undesirable results. 
To verify this effect in Mamba, we compare MambaMIL with MHIM-MambaMIL, which incorporates the pruning mechanism from MHIM-MIL. 
As shown in Tab.~\ref{tab:comparison_MHIM}, the results indicate that while MHIM-MambaMIL leads to improvements in some cases, the overall performance gain is not statistically significant ($ P > 0.05$ across two baselines, three datasets, and five runs).
This can be attributed to the fact that pruning a large number of tokens in Mamba may disrupt the temporal structure within the memory.
Furthermore, relying solely on attention scores without appropriate supervision may introduce unstable performance and boundary issues.

Given these challenges, it is natural to consider discarding the memory of irrelevant or background tokens in the hidden states to extend Mamba’s applicability to large WSIs.
To this end, we propose Contextual Token Selection (CTS), which aims to identify informative tokens while filtering out irrelevant ones. 
Instead of relying on attention scores or the learned $\Delta_k$ values which may be noisy and lack supervision, we utilize a supervised auxiliary branch, termed instance learner $h_\theta$, where $\theta$ denotes its learnable parameters. 
Using simple yet effective operations such as max or mean pooling, the instance learner makes decisions at the instance level. 
Coupled with supervisory knowledge, it provides a much more reliable indicator of token importance.
Specifically, token importance can be quantified via entropy, where tokens with higher entropy values are regarded as less informative:
\begin{equation}
    \mathbb{S}_r = \left\{ x \;\middle|\; x \in \mathcal{X}, \; H\!\left(\hat{Y} \mid x\right) \geq \alpha_{r} \right\},
    \label{eq:cts}
\end{equation}
where $\mathbb{S}_r$ denotes the set of less important tokens selected by entropy, $H(\cdot)$ is the softmax entropy function, $\hat{Y} = g_\theta(x)$ represents the logits from the instance learner, and $\alpha_{r}$ is the entropy threshold corresponding to the top-$r$ percentile (i.e., the highest $r$-ratio of entropy values among all tokens). 
Subsequently, the instance-level mask can be derived as follows:
\begin{equation}
    [M]_{i} =
    \begin{cases}
        0, & \text{if } x_i \in \mathbb{S}_{r}, \\
        1, & \text{otherwise}.
    \end{cases}
\end{equation}
where $[M]_{i}$ denotes the mask for the $i$-th token.
By masking $\Delta_k \in \mathbb{S}_r$ corresponding to irrelevant tokens, the effective memory can be extended while preserving contextual information, and the hidden state update can be expressed as:

\begin{equation}
    h_i =
    \begin{cases}
        h_{i-1}, & \text{if } x_i \in \mathbb{S}_{r}, \\[6pt]
        \bar{A}_i h_{i-1} + \bar{B}_i x_i, & \text{otherwise},
    \end{cases}
\end{equation}

\begin{figure*}[!ph]
    \centering
    \includegraphics[width=\linewidth]{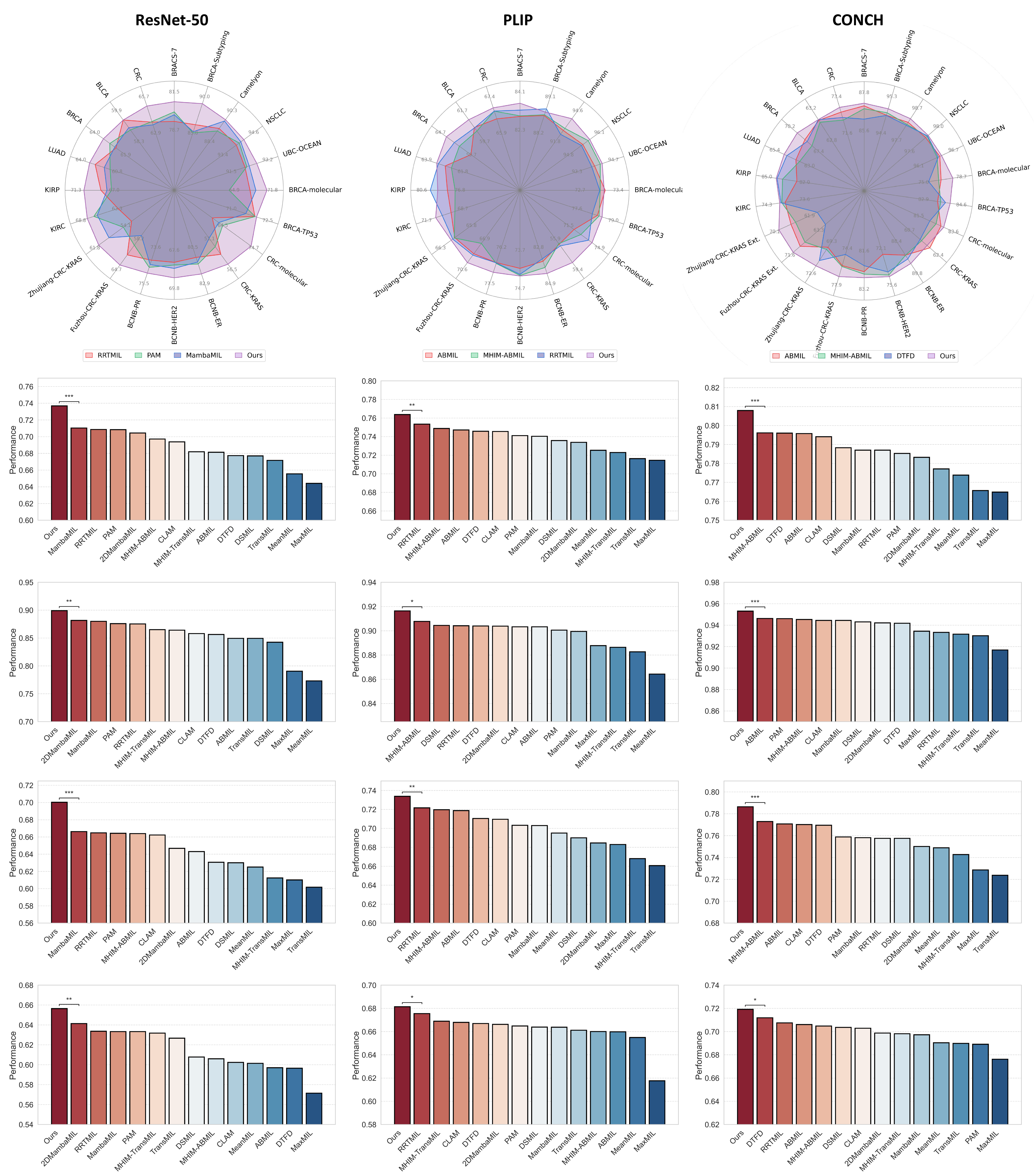}
    \caption{Performance comparison of MIL methods using three different feature extractors (ResNet-50, PLIP, and CONCH, from left to right). 
    The first row presents the performance of the top four MIL methods across 20 benchmarks.
    The second row reports the average performance of all MIL methods across all tasks, while the third, fourth, and fifth rows further break down the results for diagnostic classification, molecular prediction, and survival analysis tasks, respectively. 
    Ext., external test set. 
    The symbol * denotes statistical significance in paired t-test between \ModelName and the second performing method (*** indicates $p\leq0.001$, ** indicates $p\leq0.01$, * indicates $p\leq0.05$).}
    \label{fig:vis_results}
\end{figure*}

\begin{table*}[ht]
\centering
\scriptsize
\resizebox{\linewidth}{!}{
\setlength{\tabcolsep}{3pt}
\begin{tabular}{@{}l|*{3}{c}|*{3}{c}|*{3}{c}|*{3}{c}|*{3}{c}|*{3}{c}@{}}
\toprule
Dataset & \multicolumn{3}{c}{BRACS-7} & \multicolumn{3}{|c}{BRCA-Subtyping} & \multicolumn{3}{|c}{Camelyon} & \multicolumn{3}{|c}{NSCLC} & \multicolumn{3}{|c}{UBC-OCEAN} & \multicolumn{3}{|c}{AVERAGE}\\
\midrule
Metric & AUC & ACC & F1 & AUC & ACC & F1 & AUC & ACC & F1 & AUC & ACC & F1 & AUC & ACC & F1 & AUC & ACC & F1\\
\midrule
\multicolumn{19}{c}{ResNet-50}\\
\midrule
MeanMIL     & 71.3$_{1.2}$ & 23.0$_{2.1}$ & 17.5$_{2.5}$ & 79.0$_{3.4}$ & 62.6$_{4.1}$ & 64.2$_{4.3}$ & 70.0$_{2.3}$ & 63.3$_{2.3}$ & 62.5$_{3.5}$ & 83.9$_{1.8}$ & 77.5$_{1.6}$ & 77.5$_{1.6}$ & 83.4$_{3.2}$ & 42.3$_{2.0}$ & 41.2$_{4.0}$ & 77.5 & 53.7 & 52.6 \\
MaxMIL      & 72.4$_{2.8}$ & 23.9$_{2.1}$ & 17.0$_{1.9}$ & 69.2$_{6.1}$ & 50.4$_{0.9}$ & 45.4$_{1.9}$ & 81.1$_{13.7}$ & 74.1$_{12.0}$ & 75.3$_{12.7}$ & 88.6$_{3.2}$ & 80.1$_{3.3}$ & 80.0$_{3.3}$ & 84.9$_{2.4}$ & 43.1$_{1.3}$ & 41.3$_{2.3}$ &79.3 & 54.3 & 51.8  \\
CLAM~\cite{lu2021clam}       & 76.4$_{3.2}$ & 29.2$_{2.6}$ & 24.5$_{3.5}$ & 84.1$_{2.1}$ & \sbest{74.2$_{3.5}$} & 74.0$_{3.3}$ & 89.6$_{2.4}$ & 84.1$_{2.7}$ & 85.3$_{2.9}$ & 92.9$_{1.2}$ & 86.3$_{1.7}$ & 86.3$_{1.7}$ & 87.2$_{2.9}$ & 56.9$_{7.9}$ & 56.2$_{7.2}$ &86.0 & 66.1 & 65.3 \\
ABMIL~\cite{ilse2018abmil}      & 74.7$_{2.0}$ & 24.7$_{2.2}$ & 17.6$_{2.1}$ & 84.6$_{2.0}$ & 72.2$_{1.9}$ & 73.3$_{2.3}$ & 87.9$_{5.9}$ & 82.2$_{4.5}$ & 83.2$_{4.7}$ & 92.3$_{1.0}$ & 83.3$_{2.1}$ & 83.1$_{2.2}$ & 86.3$_{3.6}$ & 51.0$_{6.5}$ & 49.3$_{7.8}$ &85.2 & 62.7 & 61.3 \\
MHIM-ABMIL~\cite{tang2023MHIM} &78.4$_{2.4}$ &29.9$_{1.8}$ &26.2$_{2.1}$ &83.7$_{1.3}$ &70.9$_{1.1}$ &72.5$_{1.4}$ &89.4$_{3.1}$ &83.5$_{2.7}$ &84.3$_{2.7}$ &92.9$_{2.3}$ &85.3$_{2.6}$ &85.1$_{2.8}$ &88.6$_{2.8}$ &59.0$_{5.9}$ &60.9$_{6.6}$ &86.6 &65.7 &65.8 \\
DSMIL~\cite{li2021dsmil} & 74.0$_{1.8}$ & 27.8$_{4.4}$ & 22.4$_{5.1}$ & 84.0$_{4.3}$ & 73.9$_{6.8}$ & 74.5$_{6.5}$ & 87.2$_{4.6}$ & 78.9$_{6.4}$ & 79.7$_{7.2}$ & 91.1$_{3.2}$ & 84.4$_{6.5}$ & 83.1$_{7.2}$ & 86.0$_{4.0}$ & 50.9$_{5.8}$ & 49.2$_{6.8}$ & 84.5 & 63.2 & 61.8 \\
DTFD~\cite{zhang2022dtfd}        & 77.5$_{3.0}$ & 27.4$_{1.7}$ & 22.0$_{2.2}$ & 83.7$_{1.6}$ & 67.1$_{6.0}$ & 69.3$_{7.1}$ & 87.9$_{4.0}$ & 79.4$_{4.9}$ & 80.4$_{5.1}$ & 93.7$_{1.3}$ & 86.4$_{2.1}$ & 86.3$_{2.3}$ & 86.5$_{3.3}$ & 50.3$_{2.7}$ & 49.5$_{3.7}$ &85.8 & 62.1 & 61.5 \\
TransMIL~\cite{shao2021transmil}    & 74.6$_{2.1}$ & 29.5$_{5.1}$ & 28.1$_{7.1}$ & 82.8$_{4.5}$ & 69.5$_{5.3}$ & 70.9$_{4.6}$ & 89.0$_{5.0}$ & 82.5$_{3.9}$ & 82.0$_{3.6}$ & 91.5$_{3.3}$ & 80.3$_{5.6}$ & 79.7$_{6.1}$ & 87.9$_{1.5}$ & 61.9$_{2.7}$ & 61.7$_{2.9}$ &85.2 & 64.8 & 64.5 \\
MHIM-TransMIL~\cite{tang2023MHIM} &76.6$_{1.4}$ &33.2$_{5.2}$ &30.4$_{6.8}$ &84.5$_{1.3}$ &71.4$_{6.0}$ &73.2$_{5.1}$ &90.0$_{3.6}$ &80.9$_{8.1}$ &79.7$_{10.0}$ &93.4$_{1.6}$ &84.1$_{1.6}$ &83.8$_{1.8}$ &89.0$_{2.1}$ &58.8$_{8.0}$ &60.0$_{8.8}$ &86.7 &65.7 &65.4\\
RRTMIL~\cite{tang2024rrtmil}      & 78.7$_{1.2}$ & 33.8$_{3.9}$ & 29.3$_{3.3}$ & \sbest{86.4$_{1.8}$} & 72.7$_{5.2}$ & 74.0$_{4.3}$ & 88.8$_{3.8}$ & 83.2$_{4.8}$ & 84.1$_{4.3}$ & 93.4$_{1.3}$ & 84.4$_{3.0}$ & 84.2$_{3.1}$ & 91.5$_{2.3}$ & 68.1$_{6.2}$ & 69.6$_{6.3}$ &87.7 & 68.4 & 68.2 \\
PAM~\cite{huang2024pam} & 80.1$_{1.9}$ & 40.7$_{4.5}$ & 39.9$_{4.7}$ & 85.0$_{3.7}$ & 72.6$_{5.4}$ & \sbest{74.8$_{5.8}$} & 88.4$_{2.8}$ & 78.3$_{5.3}$ & 76.7$_{6.7}$ &  93.8$_{0.5}$ & 84.3$_{1.8}$ & 83.9$_{1.9}$ & 91.7$_{1.6}$ & 66.8$_{3.7}$ & 66.9$_{4.4}$ &87.8 & 68.5 & 68.4 \\
MambaMIL~\cite{yang2024mambamil} & 79.6$_{3.7}$ & 38.4$_{5.1}$ & 37.7$_{6.6}$ & 85.2$_{4.3}$ & 74.0$_{3.5}$ & 73.7$_{4.8}$ & \sbest{90.1$_{1.2}$} & \sbest{84.6$_{1.4}$} & \sbest{85.5$_{1.4}$} & 94.1$_{1.4}$ & 86.9$_{4.7}$ & 86.6$_{5.2}$ & 92.1$_{2.3}$ & 68.9$_{2.3}$ & 70.7$_{2.4}$ &88.2 & \sbest{70.6} & \sbest{70.8} \\
2DMambaMIL~\cite{zhang20252dmamba} & \sbest{81.1$_{2.3}$} & \sbest{41.4$_{1.9}$} & \sbest{40.6$_{2.0}$} & 85.2$_{2.9}$ & 67.4$_{6.4}$ & 69.6$_{6.7}$ & 88.7$_{1.8}$ & 84.5$_{1.8}$ & 85.3$_{2.0}$ & \sbest{94.1$_{0.8}$} & \sbest{87.3$_{1.0}$} & \sbest{87.2$_{1.0}$} & \sbest{92.8$_{1.5}$} & \sbest{69.0$_{5.4}$} & 69.8$_{5.7}$ & \sbest{88.4} & 69.9 & 70.5\\
\ModelName  & \best{81.5$_{3.1}$} & \best{43.2$_{4.9}$} & \best{42.5$_{5.9}$} & \best{90.0$_{3.3}$} & \best{80.5$_{3.9}$} & \best{81.8$_{4.0}$} & \best{90.3$_{1.0}$} & \best{86.7$_{2.0}$} & \best{86.9$_{1.8}$} & \best{94.6$_{1.4}$} & \best{87.3$_{1.5}$} & \best{87.3$_{1.5}$}  & \best{93.2$_{1.9}$} & \best{69.7$_{3.6}$ }& \best{71.0$_{3.3}$} &\best{89.9} & \best{73.5} & \best{73.8} \\
\midrule
\multicolumn{19}{c}{PLIP}\\
\midrule
MeanMIL     & 77.6$_{3.6}$ & 35.7$_{4.3}$ & 34.0$_{3.2}$ & 87.8$_{2.0}$ & 74.2$_{3.5}$ & 75.5$_{1.5}$ & 80.5$_{4.3}$ & 70.8$_{2.7}$ & 71.7$_{3.0}$ & 93.5$_{1.3}$ & 84.8$_{1.8}$ & 84.6$_{1.8}$ & 92.7$_{1.9}$ & 68.4$_{5.1}$ & 70.0$_{5.5}$ &86.4 & 66.8 & 67.2 \\
MaxMIL      & 75.1$_{4.5}$ & 30.8$_{4.9}$ & 27.0$_{4.7}$ & 87.1$_{1.4}$ & 73.1$_{6.2}$ & 76.4$_{5.7}$ & 92.9$_{0.5}$ & 86.9$_{1.7}$ & 87.6$_{1.4}$ & 94.6$_{0.9}$ & 85.9$_{3.0}$ & 85.9$_{3.1}$ & 94.3$_{0.8}$ & 71.9$_{3.2}$ & 73.3$_{3.6}$ &88.8 & 69.7 & 70.0 \\
CLAM~\cite{lu2021clam}        & 81.3$_{3.0}$ & 41.0$_{4.2}$ & 39.5$_{4.6}$ & 88.7$_{1.8}$ & 71.9$_{6.8}$ & 74.7$_{6.2}$ & 92.4$_{0.9}$ & 85.8$_{2.0}$ & 86.5$_{2.2}$ & 95.0$_{0.5}$ & 87.1$_{1.8}$ & 87.1$_{1.8}$ & 94.3$_{0.6}$ & 73.0$_{4.2}$ & 72.9$_{4.3}$ &90.3 & 71.8 & 72.1 \\
ABMIL~\cite{ilse2018abmil}      & 82.3$_{2.6}$ & 43.1$_{4.8}$ & 43.0$_{5.1}$ & 88.2$_{1.6}$ & 75.7$_{5.4}$ & 76.6$_{3.0}$ & 92.5$_{1.2}$ & 86.5$_{2.0}$ & 87.7$_{1.6}$ & 95.0$_{0.4}$ & 88.1$_{1.0}$ & 88.1$_{1.0}$ & 93.7$_{1.7}$ & 74.5$_{3.4}$ & 74.1$_{3.5}$ &90.3 & 73.6 & 73.9 \\
MHIM-ABMIL~\cite{tang2023MHIM} &82.4$_{2.7}$ &41.2$_{5.3}$ &40.1$_{5.4}$ &88.3$_{2.3}$ &78.6$_{5.8}$ &78.6$_{4.9}$ &92.9$_{0.4}$ &87.0$_{1.4}$ &87.0$_{1.2}$ &\sbest{95.8$_{0.7}$} &87.8$_{1.9}$ &87.8$_{2.0}$ & \sbest{94.4$_{0.5}$} &73.5$_{3.2}$ &74.3$_{3.0}$ &\sbest{90.7} &73.6 &73.5 \\
DSMIL~\cite{li2021dsmil}      & 82.3$_{2.0}$ & 41.7$_{4.5}$ & 39.5$_{5.4}$ & 88.0$_{1.7}$ & 75.8$_{2.0}$ & 77.5$_{2.7}$ & 92.2$_{1.7}$ & 85.3$_{3.3}$ & 85.3$_{3.7}$ & 95.6$_{0.8}$ & 87.1$_{2.1}$ & 86.9$_{2.3}$ & 94.2$_{1.3}$ & 75.1$_{1.7}$ & 75.4$_{2.0}$ & 90.4 & 73.0 & 72.9 \\
DTFD~\cite{zhang2022dtfd}        & 81.1$_{2.5}$ & 36.2$_{3.1}$ & 34.2$_{3.3}$ & 88.2$_{2.1}$ & 74.9$_{4.5}$ & 77.3$_{3.4}$ & \sbest{93.3$_{1.2}$} & \sbest{87.5$_{1.2}$} & \sbest{88.4$_{1.2}$} & 95.0$_{1.3}$ & 84.5$_{8.4}$ & 83.7$_{9.9}$ & 94.4$_{0.9}$ & \sbest{75.9$_{3.7}$} & \sbest{75.4$_{2.8}$} &90.4 & 71.8 & 72.0\\
TransMIL~\cite{shao2021transmil}    & 75.5$_{3.7}$ & 30.4$_{6.6}$ & 26.8$_{9.1}$ & 85.5$_{2.8}$ & 73.6$_{6.1}$ & 75.2$_{5.0}$ & 91.7$_{2.1}$ & 83.9$_{5.1}$ & 83.1$_{6.8}$ & 94.9$_{0.9}$ & 87.0$_{1.3}$ & 86.8$_{1.3}$ & 93.8$_{0.9}$ & 70.6$_{4.5}$ & 70.5$_{4.7}$ &88.3 & 69.1 & 68.5 \\
MHIM-TransMIL~\cite{tang2023MHIM} &78.4$_{2.2}$ &35.3$_{4.7}$ &33.2$_{5.0}$ &83.7$_{1.3}$ &70.9$_{1.1}$ &72.5$_{1.4}$ &92.5$_{2.0}$ &83.7$_{3.6}$ &82.6$_{5.0}$ &95.2$_{0.8}$ &86.3$_{3.7}$ &86.1$_{3.9}$ &93.4$_{2.1}$ &70.1$_{7.7}$ &69.5$_{8.7}$ &88.6 &69.3 &68.8 \\
RRTMIL~\cite{tang2024rrtmil}      & \sbest{83.1$_{1.6}$} & 42.4$_{4.1}$ & 40.4$_{4.7}$ & \best{89.1$_{1.4}$} & 73.5$_{7.6}$ & 76.1$_{7.4}$ & 91.8$_{1.2}$ & 86.3$_{0.9}$ & 87.0$_{1.5}$ & 94.8$_{0.7}$ & \sbest{88.8$_{1.3}$} & \sbest{88.8$_{1.3}$} & 93.3$_{1.2}$ & 73.7$_{4.4}$ & 73.8$_{5.2}$ &90.4 & 73.0 & 73.2 \\
PAM~\cite{huang2024pam}         & 82.0$_{1.0}$ & 42.1$_{5.0}$ & 41.2$_{4.8}$ & 87.2$_{2.8}$ & 76.6$_{3.4}$ & 78.1$_{3.1}$ & 91.2$_{2.3}$ & 78.2$_{8.0}$ & 75.2$_{11.5}$ & 95.5$_{0.9}$ & 87.6$_{1.2}$ & 87.6$_{1.2}$ & 94.4$_{0.9}$ & 70.7$_{3.8}$ & 71.5$_{3.1}$ &90.1 & 71.0 & 70.7 \\
MambaMIL~\cite{yang2024mambamil}    & 81.8$_{2.2}$ & 42.0$_{4.2}$ & 40.8$_{5.1}$ & 87.7$_{1.4}$ & 76.1$_{4.5}$ & 77.3$_{2.6}$ & 91.0$_{3.0}$ & 82.9$_{4.8}$ & 82.4$_{5.5}$ & 95.3$_{0.5}$ & 86.7$_{1.9}$ & 86.5$_{2.1}$ & 94.0$_{0.4}$ & 72.7$_{5.7}$ & 72.8$_{4.3}$ &90.0 & 72.1 & 72.0  \\
2DMambaMIL~\cite{zhang20252dmamba} & 82.8$_{0.4}$ & \sbest{45.2$_{4.2}$} & \sbest{43.0$_{3.3}$} & 88.4$_{1.5}$ & \sbest{78.8$_{6.8}$} & \sbest{78.7$_{5.6}$} & 90.8$_{3.3}$ & 85.5$_{2.1}$ & 86.1$_{1.8}$ & 95.5$_{0.9}$ & 87.6$_{0.8}$ & 87.5$_{0.9}$ & 94.4$_{0.8}$ & 75.3$_{3.1}$ & 75.4$_{3.1}$ & 90.4 & \sbest{74.5} & \sbest{74.2} \\
\ModelName        & \best{84.1$_{2.0}$} & \best{45.3$_{3.0}$} & \best{43.1$_{3.1}$} & \sbest{88.7$_{2.0}$} & \best{79.8$_{4.2}$} & \best{78.8$_{3.0}$} & \best{94.6$_{2.4}$} & \best{87.8$_{1.3}$} & \best{88.9$_{0.8}$} & \best{96.1$_{0.6}$} & \best{89.2$_{0.9}$} & \best{89.2$_{1.0}$} & \best{94.7$_{1.1}$} & \best{75.9$_{3.9}$} & \best{75.6$_{3.2}$} &\best{91.6} & \best{75.6} & \best{75.1} \\
\midrule
\multicolumn{19}{c}{CONCH}\\
\midrule
MeanMIL     & 83.7$_{1.4}$ & 44.5$_{3.6}$ & 43.6$_{3.1}$ & 94.1$_{0.4}$ & 81.0$_{7.1}$ & 83.9$_{5.4}$ & 89.6$_{2.6}$ & 79.9$_{3.5}$ & 81.5$_{3.9}$ & 96.7$_{1.2}$ & 90.0$_{0.4}$ & 89.9$_{0.6}$ & 95.4$_{1.5}$ & 77.1$_{2.8}$ & 76.2$_{4.5}$ & 91.9 & 74.5 & 75.0 \\
MaxMIL      & 83.1$_{1.6}$ & 43.9$_{6.0}$ & 43.1$_{5.4}$ & 93.8$_{1.2}$ & 80.0$_{2.6}$ & 83.3$_{2.3}$ & 97.5$_{1.5}$ & 86.5$_{6.6}$ & 84.6$_{9.1}$ & 97.4$_{0.6}$ & 90.3$_{1.3}$ & 90.3$_{1.3}$ & 96.4$_{0.5}$ & 81.3$_{2.8}$ & 81.5$_{2.5}$ & 93.6 & 76.4 & 76.6 \\
CLAM~\cite{lu2021clam}       & 86.9$_{2.0}$ & 43.8$_{5.5}$ & 40.5$_{5.9}$ & 94.5$_{1.5}$ & 87.6$_{2.7}$ & \sbest{87.5$_{1.4}$} & 97.5$_{1.3}$ & 92.9$_{1.7}$ & 93.2$_{1.8}$ & \sbest{98.0$_{0.6}$} & \sbest{92.7$_{1.8}$} & \best{93.0$_{1.8}$} & 96.4$_{0.8}$ & 79.3$_{5.1}$ & 79.4$_{3.5}$ & 94.7 & 79.3 & 78.7 \\
ABMIL~\cite{ilse2018abmil}      & 87.5$_{1.9}$ & 49.7$_{8.4}$ & \sbest{48.6$_{8.2}$} & 94.4$_{0.9}$ & 83.4$_{4.5}$ & 85.1$_{3.2}$ & \sbest{98.1$_{0.9}$} & 93.8$_{2.0}$ & \sbest{94.4$_{1.7}$} & 97.6$_{1.2}$ & 92.4$_{1.4}$ & 92.4$_{1.4}$ & 96.5$_{0.6}$ & 81.7$_{2.9}$ & 81.3$_{3.3}$ & \sbest{94.8} & \sbest{80.2} & \sbest{80.4} \\
MHIM-ABMIL~\cite{tang2023MHIM} &87.1$_{0.9}$ &49.3$_{5.0}$ &47.9$_{5.6}$ &\sbest{94.8$_{0.6}$} &84.6$_{5.2}$ &86.1$_{0.0}$ &97.8$_{0.7}$ &90.9$_{4.2}$ &90.0$_{5.3}$ &97.9$_{0.7}$ &92.3$_{1.4}$ &92.2$_{1.4}$ &96.1$_{0.6}$ &81.8$_{4.6}$ &81.8$_{3.8}$ &94.7 &79.8 &79.6 \\
DSMIL~\cite{li2021dsmil}       & 86.3$_{1.7}$ & 48.5$_{4.2}$ & 47.5$_{2.6}$ & 94.7$_{1.1}$ & 86.0$_{4.1}$ & 86.9$_{2.3}$ & 97.8$_{1.1}$ & 92.7$_{1.4}$ & 93.0$_{1.2}$ & 97.6$_{0.9}$ & 92.8$_{1.3}$ & 92.9$_{1.3}$ & 96.1$_{0.8}$ & 79.6$_{7.5}$ & 79.7$_{6.6}$ & 94.5 & 79.9 & 80.0 \\
DTFD~\cite{zhang2022dtfd}       & 85.6$_{1.9}$ & 47.8$_{5.9}$ & 46.6$_{5.7}$ & 94.4$_{2.1}$ & 85.5$_{6.8}$ & 85.9$_{4.2}$ & 97.7$_{1.1}$ & 92.2$_{1.6}$ & 92.1$_{2.2}$ & 97.9$_{0.5}$ & 91.6$_{1.2}$ & 91.6$_{1.2}$ & 96.3$_{1.1}$ & \sbest{82.1$_{5.0}$} & \best{82.2$_{4.0}$} & 94.4 & 79.9 & 79.7 \\
TransMIL~\cite{shao2021transmil}    & 82.7$_{2.4}$ & 33.8$_{5.7}$ & 29.9$_{7.0}$ & 93.2$_{1.6}$ & 86.6$_{3.7}$ & 85.4$_{2.2}$ & 97.4$_{0.9}$ & 93.3$_{2.3}$ & 93.4$_{2.4}$ & 97.6$_{0.5}$ & 91.9$_{1.1}$ & 91.9$_{1.1}$ & 95.2$_{1.0}$ & 75.8$_{5.1}$ & 76.1$_{3.9}$ & 93.2 & 76.3 & 75.3 \\
MHIM-TransMIL~\cite{tang2023MHIM} &82.2$_{1.6}$ &42.4$_{2.2}$ &41.1$_{2.3}$ &93.4$_{0.8}$ &79.7$_{10.9}$ &80.6$_{9.7}$ &98.0$_{0.8}$ &85.3$_{6.8}$ &82.6$_{9.0}$ &97.3$_{0.9}$ &90.3$_{2.9}$ &90.2$_{2.9}$ &95.9$_{0.9}$ &82.0$_{6.0}$ &81.0$_{5.6}$ &93.4 &75.9 &75.1 \\
RRTMIL~\cite{tang2024rrtmil} & 83.0$_{1.4}$ & 45.8$_{2.1}$ & 44.3$_{1.9}$ & 94.0$_{1.0}$ & \sbest{87.6$_{2.2}$} & 85.3$_{3.8}$ & 97.3$_{1.5}$ & 90.4$_{3.9}$ & 89.6$_{4.4}$ & 97.7$_{0.5}$ & 92.4$_{1.7}$ & 92.5$_{1.7}$ & 95.7$_{0.8}$ & 79.5$_{5.0}$ & 79.6$_{3.9}$ & 93.5 & 79.1 & 78.2 \\
PAM~\cite{huang2024pam}         & \sbest{87.6$_{1.7}$} & \sbest{50.8$_{5.7}$} & 48.6$_{6.0}$ & 94.4$_{1.1}$ & 85.8$_{3.8}$ & 87.4$_{2.3}$ & 97.9$_{1.0}$ & \sbest{94.0$_{2.4}$} & 93.9$_{2.7}$ & 97.9$_{0.5}$ & 91.3$_{2.0}$ & 91.2$_{2.1}$ & 96.3$_{0.8}$ & 78.0$_{3.8}$ & 78.5$_{3.7}$ & 94.8 & 80.0 & 79.9 \\
MambaMIL~\cite{yang2024mambamil}    & 86.9$_{1.3}$ & 49.4$_{6.0}$ & 48.5$_{5.6}$ & 94.4$_{1.5}$ & 84.7$_{4.4}$ & 86.3$_{2.6}$ & 97.6$_{0.9}$ & 89.8$_{5.1}$ & 89.0$_{6.6}$ & 97.9$_{0.7}$ & 92.5$_{1.6}$ & 92.6$_{1.6}$ & 96.4$_{0.6}$ & 81.9$_{2.7}$ & 81.7$_{2.5}$ & 94.7 & 79.6 & 79.6 \\
2DMambaMIL~\cite{zhang20252dmamba}  & 86.9$_{2.8}$ & 46.9$_{6.4}$ & 45.6$_{6.1}$ & 94.1$_{1.5}$ & 86.5$_{2.9}$ & 87.2$_{2.0}$ & 97.0$_{0.8}$ & 91.5$_{4.7}$ & 91.1$_{6.0}$ & 97.6$_{0.7}$ & 91.8$_{2.0}$ & 91.7$_{2.1}$ & \sbest{96.5$_{1.0}$} & 81.8$_{4.7}$ & 81.8$_{4.3}$ & 94.4 & 79.7 & 79.5 \\
\ModelName        & \best{87.9$_{1.5}$} & \best{51.3$_{4.1}$} & \best{49.1$_{2.8}$} & \best{95.3$_{1.3}$} & \best{90.0$_{1.3}$} & \best{89.0$_{1.9}$} & \best{98.7$_{0.9}$} & \best{94.4$_{2.7}$} & \best{95.0$_{2.2}$} & \best{98.0$_{0.7}$} & \best{92.9$_{0.7}$} & \sbest{92.7$_{0.7}$} & \best{96.7$_{0.7}$} & \best{82.4$_{3.4}$} & \sbest{81.9$_{3.5}$} & \best{95.3} & \best{82.2} & \best{81.5} \\

\bottomrule
\end{tabular}
}
\caption{Performance comparison of our method with MIL baselines on the diagnostic classification benchmark. The values highlighted in \best{red} and \sbest{blue} denote the best and second-best performances, respectively.}
\label{tab:comparison_subtyping}
\end{table*}

\begin{table*}[htbp]
\centering
\scriptsize
\resizebox{\linewidth}{!}{
\setlength{\tabcolsep}{3pt}
\begin{tabular}{@{}l|*{3}{c}|*{3}{c}|*{3}{c}|*{3}{c}|*{3}{c}@{}}
\toprule
Dataset & \multicolumn{3}{c}{BRCA-Molecular} & \multicolumn{3}{|c}{BRCA-TP53} & \multicolumn{3}{|c}{BCNB-ER} & \multicolumn{3}{|c}{BCNB-HER2} & \multicolumn{3}{|c}{BCNB-PR} \\
\midrule
Metric & AUC & ACC & F1 & AUC & ACC & F1 & AUC & ACC & F1 & AUC & ACC & F1 & AUC & ACC & F1 \\
\midrule
\multicolumn{16}{c}{ResNet-50}\\
\midrule
MeanMIL & 67.3$_{2.7}$ & 28.1$_{1.2}$ & 24.5$_{2.7}$ & 68.5$_{3.2}$ & 56.2$_{6.4}$ & 51.3$_{10.8}$ & 73.2$_{4.8}$ & 54.6$_{3.3}$ & 53.0$_{5.8}$ & 62.3$_{6.1}$ & 50.0$_{0.0}$ & 42.4$_{0.1}$ & 66.9$_{6.5}$ & 51.0$_{1.0}$ & 45.2$_{1.9}$ \\
MaxMIL & 61.9$_{5.5}$ & 27.0$_{5.6}$ & 22.5$_{7.6}$ & 65.8$_{3.5}$ & 52.7$_{3.1}$ & 46.5$_{7.2}$ & 79.2$_{3.0}$ & 57.7$_{4.9}$ & 57.3$_{7.2}$ & 60.6$_{4.6}$ & 50.0$_{0.0}$ & 42.4$_{0.1}$ & 69.0$_{7.3}$ & 52.8$_{4.4}$ & 47.9$_{7.7}$ \\
CLAM~\cite{lu2021clam} & 68.9$_{6.6}$ & 28.0$_{2.0}$ & 25.2$_{2.9}$ & 71.0$_{6.1}$ & 63.8$_{6.1}$ & 63.6$_{6.3}$ & 78.8$_{4.2}$ & 64.4$_{4.5}$ & 65.8$_{4.9}$ & 66.6$_{5.4}$ & 50.5$_{0.9}$ & 44.1$_{2.7}$ & 72.3$_{5.3}$ & 56.0$_{6.1}$ & 53.6$_{8.8}$ \\
ABMIL~\cite{ilse2018abmil} & 66.6$_{6.4}$ & 28.4$_{4.9}$ & 24.6$_{6.8}$ & 69.2$_{3.9}$ & 59.2$_{7.6}$ & 55.6$_{12.7}$ & 80.8$_{3.9}$ & 64.5$_{4.1}$ & 65.9$_{4.3}$ & 65.7$_{3.7}$ & 50.0$_{0.0}$ & 42.4$_{0.1}$ & 72.5$_{5.9}$ & 56.2$_{4.9}$ & 54.2$_{8.2}$ \\
MHIM-ABMIL~\cite{tang2023MHIM} &69.5$_{5.8}$ &30.0$_{3.2}$ &26.2$_{2.0}$ &71.4$_{5.0}$ &64.7$_{3.2}$ &64.9$_{3.8}$ &80.1$_{4.3}$ & 66.6$_{6.7}$ & 68.0$_{6.8}$ & 66.4$_{5.3}$ & 53.5$_{2.8}$ & 49.8$_{5.6}$ & 74.2$_{5.7}$ & 60.2$_{5.8}$ & 59.4$_{9.0}$ \\
DSMIL~\cite{li2021dsmil} & 67.8$_{3.2}$ & 28.3$_{2.3}$ & 25.2$_{2.2}$ & 69.6$_{5.7}$ & 58.7$_{9.8}$ & 53.6$_{14.6}$ & 77.4$_{4.4}$ & 56.0$_{3.8}$ & 55.0$_{5.4}$ & 63.1$_{4.4}$ & 49.9$_{0.2}$ & 42.3$_{0.1}$ & 65.7$_{2.1}$ & 51.9$_{2.0}$ & 47.2$_{4.2}$ \\
DTFD~\cite{zhang2022dtfd}  & 65.0$_{5.7}$ & 24.2$_{2.8}$ & 18.9$_{4.3}$ & 67.5$_{7.2}$ & 55.0$_{5.2}$ & 50.0$_{9.7}$ & 81.1$_{3.5}$ & 66.5$_{2.3}$ & 68.2$_{2.6}$ & 66.1$_{6.1}$ & 54.7$_{4.8}$ & 51.8$_{8.5}$ & 73.9$_{5.3}$ & 58.8$_{4.2}$ & 58.4$_{6.5}$ \\
TransMIL~\cite{shao2021transmil} & 63.1$_{3.5}$ & 24.4$_{4.3}$ & 17.9$_{6.8}$ & 67.8$_{3.6}$ & 60.7$_{8.0}$ & 58.5$_{11.0}$ & 76.9$_{3.9}$ & 66.2$_{5.0}$ & 63.2$_{5.2}$ & 54.2$_{5.0}$ & 51.7$_{3.7}$ & 46.2$_{6.7}$ & 63.9$_{10.8}$ & 54.4$_{6.9}$ & 52.1$_{9.6}$ \\
MHIM-TransMIL~\cite{tang2023MHIM} &64.2$_{4.2}$ &26.8$_{4.9}$ &20.7$_{8.0}$ &68.7$_{6.0}$ &64.1$_{3.4}$ &64.3$_{3.8}$ &76.7$_{3.7}$ & 60.1$_{7.4}$ & 59.1$_{9.6}$ & 51.2$_{7.8}$ & 51.6$_{3.3}$ & 45.6$_{6.5}$ & 66.0$_{8.9}$ & 56.7$_{4.9}$ & 55.3$_{7.8}$ \\
RRTMIL~\cite{tang2024rrtmil} & 68.7$_{5.2}$ & 31.6$_{3.7}$ & 27.4$_{5.5}$ & 72.3$_{6.1}$ & 64.8$_{6.3}$ & 63.6$_{7.6}$ & 80.5$_{4.5}$ & 64.3$_{6.1}$ & 65.5$_{5.6}$ & 67.6$_{5.5}$ & \sbest{55.6$_{6.5}$} & \sbest{52.3$_{9.5}$} & 73.6$_{6.0}$ & \sbest{61.6$_{5.8}$} & \sbest{61.7$_{8.2}$} \\
PAM~\cite{huang2024pam} & 64.9$_{2.7}$ & 31.1$_{3.3}$ & 25.7$_{5.6}$ & \sbest{72.3$_{4.5}$} & 64.6$_{4.8}$ & 62.2$_{7.6}$ & 81.5$_{5.6}$ & 63.9$_{5.0}$ & 65.7$_{5.6}$ & 68.0$_{3.5}$ & 53.7$_{3.0}$ & 50.9$_{5.9}$ & \sbest{74.7$_{3.2}$} & 59.1$_{4.6}$ & 58.8$_{7.5}$ \\
MambaMIL~\cite{yang2024mambamil} & 69.8$_{4.0}$ & \sbest{35.7$_{2.9}$} & \sbest{31.8$_{4.1}$} & 71.0$_{6.7}$ & \sbest{65.1$_{4.0}$} & \sbest{65.8$_{4.4}$} & 81.2$_{5.5}$ & 68.9$_{6.6}$ & \sbest{70.4$_{6.0}$} & \sbest{68.5$_{5.9}$} & 54.8$_{4.2}$ & 51.9$_{8.1}$ & 74.3$_{5.0}$ & 58.7$_{2.9}$ & 58.7$_{3.4}$ \\
2DMambaMIL~\cite{zhang20252dmamba} & \sbest{71.2$_{1.2}$} & 34.2$_{6.6}$ & 30.4$_{8.1}$ & 68.2$_{5.2}$ & 63.3$_{6.9}$ & 60.9$_{10.5}$ & \sbest{81.5$_{3.8}$} & \sbest{69.1$_{6.2}$} & 70.2$_{5.4}$ & 64.2$_{4.9}$ & 51.8$_{1.6}$ & 47.3$_{3.9}$ & 68.5$_{11.9}$ & 58.9$_{7.2}$ & 56.2$_{9.6}$ \\
\ModelName & \best{71.8$_{4.6}$} & \best{34.2$_{4.7}$} & \best{32.1$_{6.7}$} & \best{72.5$_{3.1}$} & \best{68.0$_{2.1}$} & \best{68.3$_{1.7}$} & \best{82.9$_{4.0}$} &\best{ 69.9$_{5.7}$} & \best{70.8$_{3.6}$} & \best{69.8$_{5.2}$} & \best{55.7$_{4.1}$} & \best{54.2$_{6.1}$} & \best{75.5$_{2.0}$} & \best{66.1$_{4.2}$} & \best{66.9$_{4.2}$}\\
\midrule
\multicolumn{16}{c}{PLIP}\\
\midrule
MeanMIL & 72.5$_{3.1}$ & 34.3$_{2.0}$ & 32.4$_{2.6}$ & 77.3$_{2.5}$ & 67.3$_{2.9}$ & 68.2$_{3.0}$ & 80.1$_{2.3}$ & 66.0$_{4.2}$ & 67.3$_{3.9}$ & 69.4$_{3.5}$ & 55.0$_{2.9}$ & 53.3$_{5.8}$ & 74.0$_{4.6}$ & 62.3$_{5.1}$ & 62.6$_{6.1}$ \\
MaxMIL & 69.8$_{7.1}$ & 31.7$_{7.3}$ & 29.1$_{9.1}$ & 76.0$_{5.5}$ & 64.4$_{3.7}$ & 63.8$_{3.8}$ & 82.4$_{3.2}$ & 68.0$_{7.0}$ & 69.3$_{7.6}$ & 72.4$_{3.9}$ & 62.5$_{0.8}$ & 63.1$_{1.4}$ & 74.7$_{3.6}$ & 60.5$_{7.3}$ & 59.2$_{10.5}$ \\
CLAM~\cite{lu2021clam} & 72.6$_{2.9}$ & 34.6$_{3.8}$ & 32.4$_{3.8}$ & 77.8$_{5.8}$ & 70.9$_{5.2}$ & 70.3$_{5.7}$ & 81.9$_{3.4}$ & 72.1$_{4.9}$ & 71.3$_{5.0}$ & 74.1$_{3.3}$ & 57.0$_{5.7}$ & 53.7$_{8.5}$ & 76.2$_{3.4}$ & 64.0$_{2.0}$ & 65.0$_{1.8}$ \\
ABMIL~\cite{ilse2018abmil} &\sbest{ 73.1$_{3.1}$} & 36.5$_{4.7}$ & 33.8$_{4.3}$ & 78.3$_{5.1}$ & 70.7$_{4.4}$ & 70.0$_{3.8}$ & 83.2$_{3.7}$ & \best{72.9$_{5.3}$} & \sbest{71.4$_{4.2}$} & 73.7$_{2.5}$ & 57.5$_{3.9}$ & 56.6$_{5.7}$ & 76.2$_{2.8}$ & 61.4$_{6.4}$ & 61.0$_{9.1}$ \\
MHIM-ABMIL~\cite{tang2023MHIM} &72.8$_{2.8}$ &37.7$_{4.5}$ &\sbest{36.0$_{6.1}$} & \sbest{78.5$_{5.3}$} &70.9$_{5.1}$ &70.8$_{5.1}$ & \sbest{84.1$_{4.2}$} & 68.1$_{6.0}$ & 70.3$_{6.8}$ & 74.6$_{3.2}$ & 61.9$_{5.6}$ & 60.7$_{5.1}$ & 76.3$_{3.0}$ & 62.4$_{6.7}$ & 61.7$_{7.2}$ \\
DSMIL~\cite{li2021dsmil}  & 72.2$_{3.9}$ & 36.1$_{3.1}$ & 32.7$_{3.6}$ & 78.4$_{4.9}$ & 68.6$_{7.3}$ & 67.8$_{8.0}$ & 82.4$_{2.9}$ & 70.5$_{1.0}$ & 72.1$_{1.8}$ & 73.5$_{3.6}$ & 58.9$_{3.5}$ & 59.0$_{4.7}$ & 75.4$_{2.0}$ & 60.4$_{8.6}$ & 57.7$_{12.3}$ \\
DTFD~\cite{zhang2022dtfd} & 72.2$_{3.5}$ & 35.0$_{5.5}$ & 32.8$_{5.6}$ & 77.9$_{5.7}$ & 69.8$_{10.1}$ & 68.5$_{12.0}$ & 83.7$_{3.7}$ & 69.4$_{6.5}$ & 69.8$_{6.3}$ & \sbest{74.6$_{4.2}$} & \sbest{63.9$_{6.3}$} & \sbest{64.0$_{5.9}$} & \sbest{76.4$_{3.8}$} & 59.3$_{5.8}$ & 58.6$_{8.7}$ \\
TransMIL~\cite{shao2021transmil} & 70.7$_{1.6}$ & 36.1$_{5.4}$ & 31.8$_{6.4}$ & 74.6$_{4.1}$ & 65.8$_{6.2}$ & 66.0$_{6.2}$ & 77.9$_{5.2}$ & 64.4$_{9.5}$ & 63.5$_{11.2}$ & 64.9$_{5.1}$ & 57.3$_{6.7}$ & 55.7$_{8.6}$ & 71.9$_{1.6}$ & 62.2$_{4.9}$ & 61.4$_{6.0}$ \\
MHIM-TransMIL~\cite{tang2023MHIM} &69.5$_{5.0}$ &33.6$_{2.4}$ &28.5$_{5.4}$ &75.0$_{2.8}$ &64.9$_{2.0}$ &65.1$_{2.2}$ &81.2$_{2.9}$ & 65.8$_{3.8}$ & 67.4$_{3.9}$ & 71.6$_{3.5}$ & 62.1$_{3.5}$ & 61.3$_{2.1}$ & 75.6$_{4.2}$ & 62.1$_{7.1}$ & 60.9$_{8.9}$ \\
RRTMIL~\cite{tang2024rrtmil}  & 72.7$_{3.6}$ & 34.3$_{8.1}$ & 31.3$_{8.6}$ & 77.6$_{6.0}$ & 67.2$_{4.6}$ & 67.0$_{5.0}$ & 82.8$_{3.0}$ & 70.3$_{5.4}$ & 69.2$_{5.6}$ & 74.4$_{3.8}$ & 62.8$_{3.7}$ & 63.6$_{4.2}$ & 76.2$_{3.1}$ & 61.1$_{8.1}$ & 58.8$_{10.4}$ \\
PAM~\cite{huang2024pam}  & 72.3$_{3.3}$ & 35.8$_{5.0}$ & 31.2$_{6.9}$ & 78.2$_{5.4}$ & 69.4$_{4.6}$ & 68.2$_{6.9}$ & 83.4$_{3.2}$ & 72.2$_{4.1}$ & 72.3$_{4.2}$ & 71.3$_{5.5}$ & 61.2$_{4.0}$ & 60.8$_{4.2}$ & 74.2$_{4.2}$ & \sbest{66.6$_{4.7}$} & \sbest{66.6$_{5.0}$} \\
MambaMIL~\cite{yang2024mambamil} & 71.6$_{2.9}$ & \sbest{37.9$_{5.9}$} & 35.3$_{6.4}$ & 75.7$_{5.2}$ & \best{71.3$_{5.3}$} & 70.7$_{5.7}$ & 82.4$_{4.2}$ & 67.0$_{7.5}$ & 67.4$_{8.5}$ & 73.5$_{4.1}$ & 66.3$_{5.8}$ & 65.0$_{4.6}$ & 73.3$_{2.6}$ & 62.6$_{1.3}$ & 63.9$_{1.3}$ \\
2DMambaMIL~\cite{zhang20252dmamba} & 72.4$_{3.7}$ & 35.5$_{5.2}$ & 34.2$_{5.5}$ & 76.2$_{3.3}$ & 70.7$_{3.2}$ & \sbest{70.9$_{3.6}$} & 82.4$_{4.5}$ & 64.7$_{9.4}$ & 64.0$_{11.4}$ & 67.2$_{1.8}$ & 56.0$_{4.6}$ & 55.2$_{6.6}$ & 74.3$_{2.6}$ & 65.2$_{3.5}$ & 65.3$_{4.0}$ \\
\ModelName & \best{73.3$_{2.8}$} & \best{38.0$_{3.8}$} & \best{37.5$_{4.2}$} & \best{79.0$_{5.0}$} & \sbest{71.0$_{5.3}$} & \best{71.2$_{5.7}$} & \best{84.9$_{3.4}$} & \sbest{71.9$_{7.1}$} & \best{71.5$_{5.9}$} & \best{74.7$_{4.8}$} & \best{65.1$_{5.4}$} & \best{65.2$_{4.6}$} & \best{77.5$_{2.5}$} & \best{68.9$_{2.2}$} & \best{69.3$_{1.1}$} \\
\midrule
\multicolumn{16}{c}{CONCH}\\
\midrule
MeanMIL & 76.2$_{2.9}$ & 43.7$_{2.0}$ & 43.9$_{3.2}$ & \sbest{83.7$_{3.9}$} & 75.6$_{4.2}$ & \sbest{76.2$_{4.7}$} & 87.1$_{2.7}$ & 71.4$_{5.0}$ & 72.8$_{5.0}$ & 72.2$_{3.8}$ & 59.4$_{2.5}$ & 59.9$_{3.2}$ & 81.4$_{3.2}$ & 68.2$_{5.0}$ & 69.3$_{5.3}$ \\
MaxMIL & 73.3$_{2.6}$ & 41.2$_{2.3}$ & 37.1$_{3.8}$ & 82.9$_{3.4}$ & 72.0$_{4.0}$ & 71.4$_{5.2}$ & 88.8$_{3.0}$ & 76.7$_{6.0}$ & 76.3$_{4.4}$ & 73.0$_{4.3}$ & 60.0$_{7.9}$ & 58.7$_{8.5}$ & 81.0$_{5.2}$ & 67.4$_{4.8}$ & 68.0$_{5.8}$ \\
CLAM~\cite{lu2021clam} & \sbest{77.6$_{2.4}$} & 43.4$_{3.1}$ & 41.3$_{3.1}$ & 82.2$_{3.5}$ & 73.7$_{4.7}$ & 73.9$_{5.5}$ & 88.4$_{3.8}$ & 72.1$_{5.0}$ & 72.3$_{5.1}$ & 74.1$_{2.2}$ & 59.4$_{4.7}$ & 58.8$_{6.4}$ & 81.4$_{5.9}$ & 64.0$_{3.9}$ & 64.9$_{5.6}$ \\
ABMIL~\cite{ilse2018abmil} & 76.9$_{2.6}$ & 41.2$_{3.6}$ & 38.9$_{3.3}$ & 82.9$_{5.2}$ & \sbest{75.6$_{4.1}$} & 76.1$_{5.1}$ & 88.4$_{3.0}$ & 76.4$_{1.5}$ & 77.0$_{3.3}$ & 72.1$_{2.5}$ & 60.5$_{5.6}$ & 59.8$_{6.9}$ & 82.4$_{3.8}$ & \sbest{72.5$_{5.0}$} & \sbest{73.3$_{4.2}$} \\
MHIM-ABMIL~\cite{tang2023MHIM} &76.5$_{2.3}$ &41.6$_{3.6}$ &40.3$_{4.5}$ &83.4$_{4.5}$ &75.0$_{5.2}$ &75.3$_{5.2}$ &88.5$_{3.4}$ & 72.1$_{6.4}$ & 74.4$_{6.8}$ & 75.4$_{4.5}$ & 61.9$_{5.6}$ & 60.7$_{5.1}$ & \sbest{82.7$_{3.2}$} & 70.5$_{6.3}$ & 71.3$_{5.6}$ \\
DSMIL~\cite{li2021dsmil} & 76.2$_{2.4}$ & 41.0$_{1.6}$ & 40.9$_{1.7}$ & 82.9$_{4.7}$ & 74.3$_{5.5}$ & 75.4$_{6.2}$ & 88.7$_{3.5}$ & 76.6$_{3.6}$ & 77.5$_{4.3}$ & \sbest{75.9$_{4.3}$} & 56.5$_{5.7}$ & 54.6$_{6.8}$ & 81.2$_{3.7}$ & 68.9$_{4.4}$ & 70.4$_{3.8}$ \\
DTFD~\cite{zhang2022dtfd} & 75.0$_{4.0}$ & 41.2$_{3.5}$ & 38.8$_{4.4}$ & 84.0$_{4.1}$ & 74.9$_{4.8}$ & 76.1$_{4.5}$ & \sbest{89.0$_{3.5}$} & \sbest{76.6$_{3.6}$} & \sbest{77.6$_{4.3}$} & 74.9$_{5.6}$ & \sbest{63.5$_{8.3}$} & \sbest{62.6$_{7.1}$} & 81.6$_{3.9}$ & 66.3$_{7.0}$ & 67.9$_{7.3}$ \\
TransMIL~\cite{shao2021transmil} & 74.7$_{3.1}$ & 39.9$_{3.2}$ & 38.5$_{3.6}$ & 82.2$_{4.2}$ & 73.9$_{6.1}$ & 71.6$_{8.6}$ & 87.2$_{3.4}$ & 69.9$_{8.9}$ & 69.8$_{7.9}$ & 68.6$_{3.1}$ & 55.6$_{4.2}$ & 52.5$_{6.8}$ & 79.3$_{4.4}$ & 65.8$_{4.6}$ & 66.4$_{5.2}$ \\
MHIM-ABMIL~\cite{tang2023MHIM} &76.5$_{2.3}$ &41.6$_{3.6}$ &40.3$_{4.5}$ &83.4$_{4.5}$ &75.0$_{5.2}$ &75.3$_{5.2}$ &86.8$_{3.4}$ & 74.1$_{7.7}$ & 74.4$_{6.2}$ & 75.4$_{4.5}$ & 62.1$_{7.6}$ & 59.8$_{9.7}$ & 79.6$_{5.8}$ & 66.6$_{6.5}$ & 66.9$_{6.9}$ \\
RRTMIL~\cite{tang2024rrtmil} & 76.3$_{1.2}$ & 42.9$_{5.1}$ & 40.8$_{5.9}$ & 83.0$_{2.9}$ & 75.1$_{3.9}$ & 74.0$_{4.7}$ & 86.3$_{4.0}$ & 73.2$_{7.8}$ & 72.6$_{6.1}$ & 74.2$_{3.7}$ & 56.6$_{5.1}$ & 54.3$_{8.5}$ & 81.5$_{5.0}$ & 65.9$_{3.6}$ & 67.6$_{4.2}$ \\
PAM~\cite{huang2024pam} & 77.5$_{2.5}$ & 41.2$_{3.4}$ & 39.6$_{4.5}$ & 83.3$_{4.7}$ & 75.4$_{6.0}$ & 75.6$_{5.8}$ & 87.8$_{4.0}$ & 67.5$_{5.0}$ & 70.3$_{5.9}$ & 75.3$_{3.8}$ & 61.9$_{2.7}$ & 61.0$_{2.3}$ & 80.1$_{1.7}$ & 67.5$_{4.2}$ & 68.7$_{3.9}$ \\
MambaMIL~\cite{yang2024mambamil} & 75.3$_{3.6}$ & 44.7$_{2.1}$ & 44.2$_{2.4}$ & 82.1$_{5.4}$ & 72.7$_{5.8}$ & 72.1$_{5.1}$ & 86.6$_{2.9}$ & 73.9$_{7.5}$ & 74.7$_{6.6}$ & 73.5$_{4.1}$ & 61.9$_{6.2}$ & 61.5$_{7.5}$ & 73.3$_{5.0}$ & 69.7$_{4.8}$ & 70.5$_{4.4}$ \\
2DMambaMIL~\cite{zhang20252dmamba} & 76.6$_{3.1}$ & \sbest{45.5$_{3.0}$} & \sbest{44.2$_{0.0}$} & 80.5$_{3.5}$ & 73.9$_{1.6}$ & 74.5$_{2.6}$ & 87.0$_{3.4}$ & 72.7$_{6.1}$ & 73.7$_{5.5}$ & 72.1$_{3.1}$ & 60.5$_{5.6}$ & 59.8$_{7.0}$ & 82.4$_{3.3}$ & 72.6$_{3.4}$ & 73.1$_{3.6}$ \\
\ModelName & \best{78.7$_{4.5}$} & \best{47.3$_{2.5}$} & \best{46.7$_{3.8}$} & \best{84.6$_{3.7}$} & \best{75.7$_{4.7}$} & \best{76.5$_{5.3}$} & \best{89.8$_{2.5}$} & \best{79.5$_{3.6}$} & \best{78.1$_{4.2}$} & \best{75.6$_{3.6}$} & \best{65.2$_{6.2}$} & \best{64.8$_{6.2}$} & \best{83.2$_{3.3}$} & \best{73.5$_{3.3}$} & \best{74.1$_{3.3}$}\\
\bottomrule
\end{tabular}
}
\caption{Performance comparison of our method with MIL baselines on BRCA-Molecular, BRCA-TP53, BCNB-ER, BCNB-HER2,and BCNB-PR. The values highlighted in \best{red} and \sbest{blue} denote the best and second-best performances, respectively.}
\label{tab:comparison_molecular_1}
\end{table*}

\begin{table*}[htbp]
\centering
\scriptsize
\resizebox{0.85\linewidth}{!}{
\setlength{\tabcolsep}{3pt}
\begin{tabular}{@{}l|ccc |ccc| ccc |ccc@{}}
\toprule
Dataset  & \multicolumn{3}{c}{CRC-Molecular}  & \multicolumn{3}{c}{CRC-KRAS} & \multicolumn{3}{c}{Fuzhou-CRC-KRAS} & \multicolumn{3}{|c}{Zhujiang-CRC-KRAS}\\
\midrule
Metric & AUC & ACC & F1 & AUC & ACC & F1& AUC & ACC & F1& AUC & ACC & F1\\
\midrule
\multicolumn{13}{c}{ResNet-50}\\
\midrule
MeanMIL & 61.9$_{5.2}$ & 30.2$_{2.9}$ & 24.3$_{4.7}$ & 52.0$_{5.2}$ & 50.3$_{0.5}$ & 37.5$_{6.7}$ &  53.7$_{5.1}$ & 49.9$_{0.2}$ & 35.5$_{4.7}$ & 56.8$_{5.2}$ & 50.3$_{2.6}$ & 40.3$_{3.9}$ \\
MaxMIL & 59.0$_{3.6}$ & 28.3$_{3.0}$ & 22.6$_{5.4}$ & 53.9$_{5.9}$ & 52.0$_{2.9}$ & 42.9$_{6.2}$ & 55.8$_{5.5}$ & 51.0$_{1.6}$ & 42.7$_{8.7}$ & 43.8$_{7.9}$ & 50.4$_{0.9}$ & 37.4$_{2.6}$ \\
CLAM~\cite{lu2021clam} & 62.6$_{5.8}$ & 30.4$_{4.5}$ & 24.7$_{8.2}$ & 54.2$_{5.2}$ & 50.0$_{0.0}$ & 34.1$_{1.2}$ & \sbest{62.9$_{6.0}$} & 56.9$_{4.9}$ & 54.7$_{7.3}$ & 58.9$_{7.0}$ & 50.5$_{0.7}$ & 38.0$_{2.4}$ \\
ABMIL~\cite{ilse2018abmil} &59.7$_{6.6}$ & 26.4$_{1.4}$ & 18.1$_{3.1}$ & 53.6$_{7.1}$ & 50.0$_{0.0}$ & 34.1$_{1.2}$& 52.1$_{7.1}$ & 50.0$_{1.0}$ & 33.7$_{0.5}$ & 58.4$_{6.4}$ & 50.3$_{0.7}$ & 35.7$_{3.1}$ \\
MHIM-ABMIL~\cite{tang2023MHIM} &61.4$_{4.9}$ &27.6$_{3.0}$ &20.9$_{6.1}$ &52.0$_{3.5}$ &50.8$_{1.1}$ &35.8$_{1.8}$ & 62.3$_{4.5}$ &\sbest{58.4$_{2.5}$} &\sbest{57.7$_{2.4}$} &\sbest{60.3$_{6.6}$} &49.0$_{1.4}$ &37.2$_{2.4}$ \\
DSMIL~\cite{li2021dsmil} &58.0$_{3.8}$ & 28.7$_{3.1}$ & 21.6$_{5.3}$ & \sbest{54.8$_{7.9}$} & 52.1$_{3.1}$ & 43.4$_{7.5}$ & 52.1$_{8.3}$ & 52.5$_{3.2}$ & 42.0$_{11.1}$ & 58.7$_{4.6}$ & 51.0$_{1.4}$ & 38.3$_{2.9}$ \\
DTFD~\cite{zhang2022dtfd}&57.9$_{4.8}$ & 27.0$_{2.2}$ & 19.0$_{4.2}$ & 48.4$_{5.8}$ & 49.3$_{2.4}$ & 35.9$_{2.7}$ & 55.8$_{5.8}$ & 50.9$_{2.0}$ & 40.9$_{8.8}$ & 52.0$_{6.3}$ & 50.6$_{1.3}$ & 39.5$_{6.7}$ \\
TransMIL~\cite{shao2021transmil}&59.1$_{3.2}$ & 29.2$_{5.5}$ & 27.0$_{7.3}$ & 51.8$_{3.5}$ & 50.3$_{1.5}$ & 42.8$_{4.9}$  & 58.1$_{6.6}$ & 52.3$_{5.6}$ & 42.2$_{11.7}$ & 46.5$_{3.4}$ & 47.1$_{4.9}$ & 42.1$_{4.9}$ \\
MHIM-TransMIL~\cite{tang2023MHIM}&62.0$_{3.9}$ &32.4$_{5.2}$ &28.0$_{9.8}$ &53.6$_{9.0}$ &49.8$_{0.4}$ &35.5$_{1.6}$ & 57.9$_{4.6}$ &55.4$_{3.0}$ &50.4$_{9.3}$ &50.9$_{3.3}$ &49.5$_{2.9}$ &41.1$_{6.7}$ \\
RRTMIL~\cite{tang2024rrtmil}&64.5$_{7.5}$ & 33.1$_{7.7}$ & 29.7$_{12.9}$ & 54.5$_{3.2}$ & 49.7$_{0.5}$ & 34.0$_{1.2}$ & 62.5$_{4.8}$ & 55.8$_{6.1}$ & 48.5$_{13.1}$ & 54.1$_{4.7}$ & 49.5$_{2.0}$ & 42.9$_{8.8}$ \\
PAM~\cite{huang2024pam}&67.3$_{5.7}$ & \sbest{41.4$_{4.5}$} & \sbest{41.0$_{5.0}$} & 52.9$_{5.2}$ & \sbest{52.3$_{3.5}$} & \sbest{46.1$_{8.5}$} & 60.4$_{7.4}$ & 53.0$_{2.2}$ & 47.3$_{8.3}$ & 56.1$_{5.4}$ & 51.0$_{1.8}$ & 40.3$_{5.5}$ \\
MambaMIL~\cite{yang2024mambamil}&66.2$_{3.8}$ & 36.0$_{5.8}$ & 33.3$_{8.5}$ & 51.1$_{5.2}$ & 50.6$_{3.5}$ & 38.7$_{8.9}$ & 58.2$_{9.4}$ & 57.7$_{4.1}$ & 53.5$_{9.7}$ & 59.4$_{3.9}$ & \sbest{54.1$_{4.5}$} & \sbest{49.3$_{9.4}$} \\
2DMambaMIL~\cite{zhang20252dmamba} &\sbest{67.4$_{4.1}$} & 37.4$_{8.7}$ & 34.5$_{10.7}$ & 51.1$_{2.4}$ & 51.2$_{1.5}$ & 37.8$_{0.4}$ & 56.4$_{5.0}$ & 52.4$_{5.0}$ & 47.6$_{8.9}$ & 53.6$_{5.4}$ & 50.3$_{0.7}$ & 38.1$_{3.9}$ \\
\ModelName  &\best{74.7$_{5.4}$} & \best{46.1$_{5.0}$} & \best{45.2$_{7.0}$} & \best{56.5$_{5.6}$} & \best{54.3$_{3.6}$} & \best{48.4$_{8.8}$} & \best{64.7$_{4.0}$} & \best{60.9$_{2.6}$} & \best{60.0$_{3.0}$} & \best{61.8$_{2.7}$} & \best{56.0$_{8.3}$} & \best{51.2$_{11.5}$} \\
\midrule
\multicolumn{13}{c}{PLIP}\\
\midrule
MeanMIL & 73.3$_{4.2}$ & 46.1$_{3.9}$ & 47.3$_{5.4}$ & 54.8$_{6.5}$ & 51.9$_{1.8}$ & 46.8$_{6.9}$ & 62.4$_{6.2}$ & 54.5$_{5.8}$ & 50.7$_{9.0}$  & 61.7$_{3.5}$ & 55.1$_{5.4}$ & 47.5$_{12.4}$ \\
MaxMIL & 64.0$_{4.8}$ & 32.2$_{6.2}$ & 28.9$_{10.2}$ & 54.1$_{9.3}$ & 52.9$_{5.0}$ & 49.4$_{7.1}$& 64.3$_{7.2}$ & 56.1$_{7.1}$ & 50.2$_{12.0}$ & 57.0$_{5.2}$ & 51.9$_{4.9}$ & 44.7$_{8.0}$  \\
CLAM~\cite{lu2021clam} & 71.9$_{5.2}$ & 48.6$_{3.7}$ & 48.5$_{4.1}$ & 54.2$_{3.1}$ & 52.5$_{1.9}$ & 51.8$_{1.9}$ & 67.5$_{5.8}$ & 61.6$_{5.0}$ & 60.7$_{5.6}$  & 62.4$_{3.8}$ & 55.1$_{4.8}$ & 48.9$_{12.2}$ \\
ABMIL~\cite{ilse2018abmil}&71.5$_{4.0}$ & 45.1$_{7.5}$ & 45.7$_{9.5}$ & 56.0$_{3.9}$ & \sbest{56.0$_{4.2}$} & 51.7$_{10.5}$ & 68.9$_{4.0}$ & 63.7$_{3.8}$ & 62.2$_{5.1}$  & 65.8$_{4.2}$ & 60.2$_{4.9}$ & 59.9$_{5.0}$  \\
MHIM-ABMIL~\cite{tang2023MHIM}&72.8$_{4.7}$ &45.6$_{4.6}$ &46.1$_{5.4}$ &55.9$_{5.3}$ &55.5$_{2.4}$ &\sbest{53.9$_{4.0}$}  &66.9$_{4.4}$ &59.7$_{8.6}$ &55.0$_{13.6}$ &65.9$_{4.3}$ &58.9$_{5.8}$ &56.3$_{7.2}$ \\
DSMIL~\cite{li2021dsmil}&69.5$_{6.1}$ & 39.0$_{10.3}$ & 36.4$_{15.2}$ & 52.9$_{6.8}$ & 51.0$_{3.5}$ & 46.1$_{7.1}$ & 64.4$_{7.3}$ & 57.3$_{6.3}$ & 53.5$_{11.2}$ & 52.4$_{12.1}$ & 54.3$_{8.5}$ & 43.1$_{14.0}$ \\
DTFD~\cite{zhang2022dtfd} &67.5$_{4.0}$ & 38.2$_{4.1}$ & 37.2$_{5.3}$ & 57.2$_{3.4}$ & 53.9$_{3.9}$ & 49.8$_{10.0}$& \sbest{70.3$_{3.1}$} & \sbest{64.0$_{3.5}$} & \sbest{62.6$_{4.5}$}  & 59.4$_{6.5}$ & 52.4$_{4.3}$ & 46.7$_{9.9}$  \\
TransMIL~\cite{shao2021transmil}&62.9$_{4.7}$ & 32.2$_{3.1}$ & 30.0$_{3.3}$ & 54.9$_{10.4}$ & 51.9$_{9.5}$ & 47.9$_{12.0}$  & 64.8$_{4.9}$ & 58.6$_{4.4}$ & 57.0$_{4.9}$  & 52.1$_{4.2}$ & 50.2$_{2.3}$ & 41.0$_{6.6}$  \\
MHIM-TransMIL~\cite{tang2023MHIM}&63.0$_{4.0}$ &38.0$_{6.8}$ &36.4$_{8.3}$ &55.7$_{7.0}$ &52.1$_{5.0}$ &45.8$_{7.8}$  &60.8$_{4.7}$ &55.2$_{5.2}$ &51.4$_{9.7}$ &49.0$_{8.1}$ &51.6$_{5.3}$ &45.9$_{9.5}$ \\
RRTMIL~\cite{tang2024rrtmil}&\sbest{74.3$_{3.9}$} & 48.1$_{4.0}$ & \best{49.1$_{3.9}$} & 56.3$_{6.1}$ & 53.7$_{3.9}$ & 48.7$_{7.8}$ & 69.2$_{2.6}$ & 62.1$_{5.6}$ & 58.0$_{10.6}$ & \sbest{66.1$_{3.9}$} & \sbest{62.7$_{7.4}$} & \sbest{61.9$_{7.2}$}  \\
PAM~\cite{huang2024pam} &69.9$_{4.8}$ & 40.6$_{0.7}$ & 40.8$_{0.9}$ & \sbest{59.3$_{8.0}$} & 54.2$_{8.3}$ & 47.1$_{11.6}$& 65.0$_{5.9}$ & 59.2$_{3.2}$ & 56.0$_{6.0}$  & 59.2$_{4.7}$ & 54.3$_{4.0}$ & 43.5$_{10.5}$ \\
MambaMIL~\cite{yang2024mambamil} & 70.7$_{4.7}$ & 46.9$_{4.5}$ & 45.9$_{4.8}$ & 56.4$_{2.5}$ & 52.3$_{2.2}$ & 48.3$_{5.1}$& 65.5$_{5.5}$ & 60.1$_{4.1}$ & 57.5$_{7.2}$  & 63.7$_{8.1}$ & 59.2$_{4.2}$ & 56.8$_{6.3}$  \\
2DMambaMIL~\cite{zhang20252dmamba}&73.8$_{4.2}$ & \sbest{48.7$_{2.0}$} & 48.0$_{2.0}$ & 54.0$_{2.9}$ & 50.3$_{1.9}$ & 41.6$_{4.0}$ & 58.1$_{9.2}$ & 53.9$_{5.1}$ & 44.1$_{12.3}$ & 57.8$_{6.7}$ & 52.5$_{4.2}$ & 43.1$_{9.0}$  \\
\ModelName  & \best{74.9$_{5.2}$} & \best{48.9$_{5.6}$} & \sbest{48.8$_{4.6}$} & \best{59.4$_{9.2}$} & \best{56.5$_{6.0}$ }& \best{55.7$_{5.7}$}& \best{70.6$_{1.0}$} & \best{64.5$_{4.1}$} & \best{63.6$_{5.1}$}  & \best{66.3$_{4.7}$} & \best{64.0$_{4.0}$} & \best{63.3$_{4.0}$}  \\
\midrule
\multicolumn{13}{c}{CONCH}\\
\midrule
MeanMIL & 80.5$_{2.5}$ & 54.1$_{4.9}$ & 54.1$_{4.0}$ & 57.0$_{8.6}$ & 53.9$_{3.9}$ & 52.0$_{3.6}$& 70.6$_{2.4}$ & 62.1$_{5.6}$ & 59.8$_{9.4}$  & 65.3$_{3.7}$ & 58.3$_{6.1}$ & 53.6$_{10.5}$ \\
MaxMIL &77.0$_{1.6}$ & 50.0$_{4.6}$ & 49.7$_{5.1}$ & 55.9$_{10.2}$ & 54.7$_{7.4}$ & 53.7$_{7.1}$ &  67.3$_{6.6}$ & 61.8$_{5.3}$ & 60.6$_{6.0}$  & 56.6$_{2.3}$ & 55.4$_{2.2}$ & 51.4$_{5.0}$ \\
CLAM~\cite{lu2021clam}&83.1$_{3.0}$ & 56.6$_{6.1}$ & 56.6$_{5.8}$ & 59.9$_{5.3}$ & 56.4$_{3.2}$ & 55.6$_{3.7}$ & 75.6$_{2.1}$ & 66.8$_{8.4}$ & 64.9$_{11.3}$ & 71.1$_{3.1}$ & 63.7$_{5.1}$ & 62.9$_{5.3}$ \\
ABMIL~\cite{ilse2018abmil}&\sbest{83.2$_{3.0}$} & 56.4$_{2.8}$ & \sbest{56.8$_{2.7}$} & \best{62.4$_{5.4}$} & \sbest{58.8$_{5.0}$} & \sbest{58.4$_{5.3}$} & 76.2$_{3.0}$ & 69.3$_{2.2}$ & \sbest{68.8$_{2.6}$}  & 69.3$_{3.9}$ & \best{66.2$_{5.5}$} & \sbest{65.0$_{4.7}$} \\
MHIM-ABMIL~\cite{tang2023MHIM}&82.4$_{4.0}$ &55.1$_{8.4}$ &54.9$_{9.3}$ &60.9$_{8.3}$ &56.7$_{7.4}$ &55.2$_{7.3}$ &\sbest{76.3$_{1.7}$} &\sbest{69.7$_{3.1}$} &68.6$_{3.1}$ &69.7$_{4.8}$ &63.9$_{6.6}$ &63.0$_{7.2}$ \\
DSMIL~\cite{li2021dsmil} &81.7$_{4.5}$ & 56.1$_{8.1}$ & 55.3$_{7.4}$ & 56.4$_{6.6}$ & 54.0$_{4.5}$ & 53.7$_{4.5}$& 71.7$_{3.8}$ & 67.0$_{2.4}$ & 66.8$_{2.2}$  & 66.9$_{6.0}$ & 62.1$_{7.5}$ & 59.1$_{12.3}$ \\
DTFD~\cite{zhang2022dtfd}&81.5$_{3.4}$ & 55.1$_{7.2}$ & 54.7$_{6.0}$ & 60.7$_{5.0}$ & 57.5$_{5.1}$ & 55.5$_{7.6}$ & 74.4$_{2.4}$ & 67.0$_{2.1}$ & 66.8$_{2.2}$  & \sbest{71.6$_{3.9}$} & 65.4$_{5.5}$ & 64.5$_{5.8}$ \\
TransMIL~\cite{shao2021transmil}&77.1$_{4.8}$ & 46.8$_{6.8}$ & 43.1$_{8.3}$ & 50.9$_{6.0}$ & 49.0$_{3.8}$ & 41.3$_{6.8}$ & 70.5$_{2.2}$ & 59.3$_{6.0}$ & 53.0$_{12.3}$ & 60.8$_{7.7}$ & 57.3$_{5.9}$ & 55.0$_{6.0}$ \\
MHIM-TransMIL~\cite{tang2023MHIM}&76.2$_{2.7}$ &47.1$_{6.4}$ &44.5$_{5.7}$ &57.9$_{3.6}$ &54.2$_{4.8}$ &51.6$_{7.8}$ &72.4$_{4.7}$ &68.5$_{2.7}$ &67.8$_{3.2}$ &64.6$_{1.1}$ &58.0$_{8.4}$ &51.7$_{15.0}$ \\
RRTMIL~\cite{tang2024rrtmil}&80.4$_{2.6}$ & 54.0$_{4.6}$ & 54.9$_{4.4}$ & 57.0$_{7.0}$ & 55.1$_{5.4}$ & 54.7$_{5.4}$ & 74.6$_{3.0}$ & 68.0$_{6.3}$ & 66.5$_{7.6}$  & 68.6$_{6.7}$ & 61.3$_{6.0}$ & 59.1$_{9.2}$ \\
PAM~\cite{huang2024pam} &82.7$_{2.6}$ & 55.9$_{3.9}$ & 56.6$_{4.5}$ & 54.9$_{5.3}$ & 54.3$_{1.9}$ & 53.9$_{2.4}$& 74.5$_{2.4}$ & 66.7$_{2.9}$ & 65.6$_{3.4}$  & 66.8$_{4.7}$ & 61.6$_{6.9}$ & 56.9$_{14.0}$ \\
MambaMIL~\cite{yang2024mambamil}&82.0$_{1.6}$ & \sbest{56.9$_{5.5}$} & 56.1$_{5.2}$ & 52.7$_{8.9}$ & 52.4$_{7.1}$ & 50.6$_{7.7}$ & 75.7$_{1.8}$ & 68.8$_{1.9}$ & 68.7$_{2.0}$  & 69.5$_{4.2}$ & 63.5$_{3.2}$ & 61.7$_{3.6}$ \\
2DMambaMIL~\cite{zhang20252dmamba}&82.6$_{3.3}$ & 54.7$_{2.8}$ & 55.5$_{3.3}$ & 55.5$_{4.0}$ & 51.9$_{1.6}$ & 49.2$_{3.4}$ & 71.6$_{1.9}$ & 64.9$_{6.9}$ & 62.7$_{8.7}$  & 66.8$_{7.8}$ & 61.5$_{9.6}$ & 55.3$_{15.7}$ \\
\ModelName &\best{83.6$_{3.6}$} & \best{58.2$_{3.0}$} & \best{59.3$_{3.9}$} & \sbest{61.9$_{6.4}$} & \best{59.8$_{5.0}$} & \best{59.3$_{5.0}$}& \best{77.9$_{3.6}$} & \best{70.1$_{5.4}$} & \best{69.8$_{5.6}$}  & \best{72.6$_{2.4}$} & \sbest{66.0$_{4.4}$} & \best{65.2$_{4.8}$} \\
\bottomrule
\end{tabular}}
\caption{Performance comparison of our method with MIL baselines on the CRC-Molecular and the CRC-KRAS datasets across three cohorts: the public TCGA cohort and the in-house Fuzhou-CRC-KRAS and Zhujiang-CRC-KRAS cohorts. The values highlighted in \best{red} and \sbest{blue} denote the best and second-best performances, respectively.}
\label{tab:comparison_molecular_2}
\end{table*}

\begin{table}[ht]
\centering
\resizebox{\linewidth}{!}{
\setlength{\tabcolsep}{3pt}
\begin{tabular}{@{}l|*{6}{c}@{}}
\toprule
Dataset & KIRC & KIRP & LUAD & BRCA & BLCA & CRC \\ 
\midrule
\multicolumn{7}{c}{ResNet-50}\\
\midrule
MeanMIL & 60.2\(_{ 11.3}\) & 66.2\(_{ 4.5}\) & 59.4\(_{ 3.2}\) & 58.7\(_{ 4.6}\) & 57.4\(_{ 5.4}\) & 59.2\(_{5.4}\)\\ 
MaxMIL & 63.4\(_{ 7.3}\) & 57.7\(_{ 6.0}\) & 57.0\(_{ 2.9}\) & 56.7\(_{ 6.1}\) & 52.1\(_{ 3.2}\) & 55.9\(_{8.2}\)\\ 
CLAM~\cite{lu2021clam} & 59.0\(_{ 10.8}\) & 65.4\(_{ 6.2}\) & 60.7\(_{ 3.3}\) & 61.3\(_{ 2.5}\) & 58.9\(_{ 6.0}\) & 56.2\(_{7.5}\)\\ 
ABMIL~\cite{ilse2018abmil} & 59.3\(_{ 10.9}\) & 65.6\(_{ 6.6}\) & 59.9\(_{ 2.5}\) & 58.8\(_{ 4.2}\) & 58.3\(_{ 6.4}\) & 56.4\(_{7.9}\) \\ 
MIHM-ABMIL~\cite{tang2023MHIM} &59.0$_{11.5}$ &67.5$_{5.6}$ &60.4$_{3.5}$ &60.5$_{2.6}$ &59.4$_{6.7}$ &56.9$_{6.3}$ \\
DSMIL~\cite{li2021dsmil} & 63.2\(_{ 10.0}\) & 65.0\(_{ 7.1}\) & 59.4\(_{ 3.2}\) & 59.7\(_{ 5.1}\) & 58.5\(_{ 5.4}\) & 58.9\(_{5.1}\)\\ 
DTFD~\cite{zhang2022dtfd} & 59.5\(_{ 6.8}\) & 66.7\(_{ 13.7}\) & 57.7\(_{ 3.0}\) & 60.2\(_{ 5.0}\) & 56.8\(_{ 6.6}\) & 57.1\(_{4.6}\)\\ 
TransMIL~\cite{shao2021transmil} & 64.9\(_{ 5.9}\) & 68.8\(_{ 7.1}\) & 61.9\(_{ 3.8}\) & 62.5\(_{ 6.0}\) & 59.4\(_{ 3.2}\) & 58.6\(_{3.9}\)\\ 
MIHM-TransMIL~\cite{tang2023MHIM} &64.4$_{5.5}$ &70.8$_{6.1}$ &62.8$_{2.1}$ &62.9$_{4.5}$ &59.3$_{4.9}$ &58.9$_{3.9}$ \\
RRTMIL~\cite{tang2024rrtmil} & 63.3\(_{ 5.7}\) & 68.6\(_{ 5.9}\) & \sbest{63.2\(_{ 4.3}\)} & 61.9\(_{ 4.4}\) & \sbest{59.9\(_{ 4.4}\)} & 63.3\(_{3.4}\)\\ 
PAM~\cite{huang2024pam} & \sbest{67.6\(_{ 4.8}\)} & 67.0\(_{ 5.0}\) & 60.8\(_{ 3.7}\) & 62.9\(_{ 3.7}\) & 58.3\(_{ 4.3}\) & 63.4\(_{5.3}\)\\ 
MambaMIL~\cite{yang2024mambamil} & 67.0\(_{ 7.6}\) & 67.7\(_{ 4.2}\) & 61.8\(_{ 3.3}\) & 62.1\(_{ 6.8}\) & 58.7\(_{ 4.2}\) & 62.9\(_{4.2}\)\\ 
2DMambaMIL~\cite{zhang20252dmamba} & 66.5$_{6.6}$ & \sbest{69.2$_{4.6}$} & 62.9$_{4.0}$ & \sbest{63.6$_{5.2}$} & 59.1$_{2.3}$ & \sbest{63.6\(_{5.7}\)}\\
\ModelName &  \best{68.8\(_{ 6.3}\)} &  \best{71.3\(_{ 5.3}\)} &  \best{64.0\(_{ 3.2}\)} &  \best{64.1\(_{ 5.0}\)} &  \best{60.1\(_{ 4.9}\)} & \best{65.7\(_{4.7}\)}\\ 
\midrule
\multicolumn{7}{c}{PLIP}\\
\midrule
MeanMIL & 68.3$_{8.6}$ & 75.2$_{7.6}$ & 62.5$_{4.7}$ & 61.3$_{6.4}$ & 59.6$_{3.4}$ & 66.1$_{2.4}$ \\
MaxMIL & 67.3$_{5.9}$ & 65.2$_{11.5}$ & 61.1$_{2.7}$ & 59.8$_{5.3}$ & 56.7$_{4.2}$ & 60.6$_{6.2}$\\
CLAM~\cite{lu2021clam} & 69.6$_{9.3}$ & 79.2$_{10.7}$ & 62.5$_{3.8}$ & 62.1$_{5.7}$ & 60.5$_{3.2}$ & 66.9$_{1.5}$\\
ABMIL~\cite{ilse2018abmil} & 69.3$_{9.4}$ & 78.0$_{9.5}$ & 62.7$_{4.3}$ & 59.7$_{6.8}$ & 60.3$_{3.6}$ & 65.9$_{1.3}$\\
MIHM-ABMIL~\cite{tang2023MHIM} &68.7$_{8.3}$ &76.8$_{7.8}$ &61.8$_{4.8}$ &62.1$_{5.8}$ &59.7$_{2.5}$ &67.0$_{1.1}$ \\
DSMIL~\cite{li2021dsmil} & 68.6$_{8.0}$ & 78.8$_{7.4}$ & 61.4$_{3.8}$ & 63.8$_{5.4}$ & 60.1$_{1.9}$ & 66.6$_{2.5}$\\
DTFD~\cite{zhang2022dtfd} & \sbest{70.1$_{7.4}$} & 78.8$_{10.0}$ & 61.4$_{3.8}$ & 63.8$_{5.4}$ & 59.9$_{4.0}$ & 66.2$_{1.5}$\\
TransMIL~\cite{shao2021transmil} & 67.2$_{6.5}$ & 78.8$_{5.3}$ & 62.3$_{4.2}$ & 62.2$_{5.7}$ & 60.2$_{1.9}$ & 66.1$_{2.7}$\\
MIHM-TransMIL~\cite{tang2023MHIM} &70.8$_{3.3}$ &78.5$_{7.4}$ &62.5$_{2.9}$ &63.7$_{5.8}$ &61.1$_{1.4}$ &64.9$_{1.4}$ \\
RRTMIL~\cite{tang2024rrtmil} & 70.7$_{4.5}$ & \best{80.6$_{5.8}$} & \sbest{63.7$_{3.8}$} & 63.1$_{5.8}$ & 60.3$_{4.3}$ & \sbest{67.0$_{2.1}$}\\
PAM~\cite{huang2024pam} & 70.0$_{5.9}$ & 74.8$_{9.7}$ & 63.2$_{5.0}$ & 63.2$_{5.9}$ & 61.0$_{2.9}$ & 66.7$_{2.9}$\\
MambaMIL~\cite{yang2024mambamil} & 68.3$_{6.2}$ & 77.4$_{9.6}$ & 63.3$_{4.4}$ & 64.2$_{6.1}$ & 59.9$_{1.7}$ & 64.8$_{3.9}$\\
2DMambaMIL~\cite{zhang20252dmamba} & 68.6$_{5.7}$ & 76.6$_{12.9}$ & 62.5$_{5.1}$ & \sbest{64.3$_{6.7}$} & \sbest{61.5$_{3.1}$} & 66.3$_{2.7}$\\
\ModelName & \best{71.7$_{4.4}$} & \sbest{79.5$_{6.3}$} & \best{64.2$_{4.5}$} & \best{64.7$_{7.8}$} & \best{61.7$_{2.7}$} & \best{67.4$_{2.6}$}\\
\midrule
\multicolumn{7}{c}{CONCH}\\
\midrule
MeanMIL & 71.9$_{4.1}$ & 79.5$_{9.9}$ & 62.5$_{3.2}$ & 67.4$_{4.6}$ & 61.7$_{4.7}$ & 71.2$_{2.4}$ \\
MaxMIL & 73.1$_{6.1}$ & 83.3$_{5.4}$ & 61.6$_{4.9}$ & 61.7$_{6.3}$ & 60.7$_{4.7}$ & 65.4$_{2.2}$ \\
CLAM~\cite{lu2021clam} & 73.8$_{6.7}$ & 81.8$_{9.3}$ & 64.0$_{4.8}$ & 67.6$_{4.7}$ & 62.7$_{3.8}$ & 71.8$_{3.3}$ \\
ABMIL~\cite{ilse2018abmil} & 73.6$_{4.1}$ & 82.0$_{7.6}$ & 63.6$_{4.7}$ & 68.7$_{4.8}$ & 63.1$_{4.1}$ & 72.7$_{2.8}$ \\
MIHM-ABMIL~\cite{tang2023MHIM} &74.0$_{5.0}$ &84.1$_{6.9}$ &63.0$_{5.0}$ &67.4$_{3.6}$ &62.8$_{2.7}$ &71.6$_{3.9}$ \\
DSMIL~\cite{li2021dsmil} & 74.0$_{5.2}$ & 82.2$_{7.9}$ & 62.5$_{3.5}$ & 68.3$_{4.3}$ & 62.1$_{3.0}$ & 73.1$_{2.3}$ \\
DTFD~\cite{zhang2022dtfd} & \sbest{74.1$_{5.0}$} & \sbest{84.9$_{5.4}$} & \sbest{65.0$_{4.0}$} & 68.3$_{5.5}$ & 63.2$_{4.3}$ & 72.0$_{2.6}$ \\
TransMIL~\cite{shao2021transmil} & 71.6$_{5.6}$ & 81.0$_{8.0}$ & 62.9$_{2.4}$ & 65.1$_{3.6}$ & \best{64.7$_{3.9}$} & 68.6$_{4.0}$ \\
MIHM-TransMIL~\cite{tang2023MHIM} &73.1$_{3.6}$ &82.8$_{7.6}$ &64.2$_{1.4}$ &66.8$_{2.4}$ &63.0$_{4.4}$ &69.1$_{4.0}$ \\
RRTMIL~\cite{tang2024rrtmil} & 72.0$_{4.0}$ & 84.9$_{7.8}$ & 65.4$_{3.0}$ & 66.9$_{6.4}$ & 63.1$_{2.5}$ & 72.1$_{3.3}$ \\
PAM~\cite{huang2024pam} & 72.0$_{3.8}$ & 80.0$_{8.1}$ & 62.5$_{2.8}$ & 67.4$_{2.9}$ & 60.3$_{3.6}$ & 71.3$_{2.2}$ \\
MambaMIL & 72.1$_{5.6}$ & 80.7$_{5.6}$ & 62.3$_{4.7}$ & \sbest{69.9$_{3.6}$} & 60.9$_{3.3}$ & 72.5$_{4.1}$ \\
2DMambaMIL~\cite{zhang20252dmamba} & 71.1$_{5.1}$ & 79.7$_{9.5}$ & 63.5$_{4.3}$ & 69.6$_{3.7}$ & 62.4$_{3.7}$ & \sbest{73.1$_{4.2}$} \\
\ModelName & \best{74.4$_{3.9}$} & \best{85.1$_{5.7}$} & \best{65.4$_{6.0}$} & \best{70.2$_{3.5}$} & \sbest{63.2$_{3.2}$} & \best{73.6$_{4.0}$} \\

\bottomrule
\end{tabular}
}
\caption{Performance comparison of different MIL methods across cancer survival datasets. The values highlighted in \best{red} and \sbest{blue} denote the best and second-best performances, respectively.}
\label{tab:comparison_survival}
\end{table}

\section{Experiments}
\subsection{Datasets and Evaluation Protocol}
To comprehensively evaluate our method, we conduct experiments across three fundamental tasks in computational pathology: diagnostic classification, survival analysis, and molecular prediction.
\subsubsection*{Diagnostic Classification}
Accurate disease classification and subtyping are fundamental to computational pathology, offering essential insights for diagnosis, treatment selection, and prognosis evaluation.
We first describe the five challenging public datasets used for diagnostic classification.
\textbf{BRACS}~\cite{bracs} dataset contains WSIs of benign, atypical, in situ, and invasive breast lesions. 
On this dataset, we perform a 7-class fine-grained carcinoma subtyping task.
\textbf{BRCA-Subtyping} cohort is a curated two-class dataset derived from The Cancer Genome Atlas (TCGA)~\cite{tcga} to distinguish invasive ductal carcinoma from invasive lobular carcinoma based on diagnostic WSIs.
\textbf{Camelyon} dataset combines data from the CAMELYON16~\cite{c16} and CAMELYON17~\cite{c17} challenges, serving as a benchmark for binary metastasis detection in sentinel lymph node WSIs.
\textbf{NSCLC} is a two-class histologic cohort built from TCGA-LUAD and TCGA-LUSC WSIs to distinguish adenocarcinoma from squamous cell carcinoma. 
\textbf{UBC-OCEAN}~\cite{ubcocean} dataset is employed for five-class ovarian cancer histotype classification: high-grade serous, low-grade serous, endometrioid, clear cell, and mucinous.

\subsubsection*{Molecular Prediction} 
Accurate prediction of gene mutations and protein expression from histopathology images can accelerate decisions for precision treatment.
In this study, we perform comparative experiments on seven public datasets, four curated as subsets of TCGA (BRCA-Molecular, BRCA-TP53, CRC-Molecular, and CRC-KRAS) and three derived from the BCNB core-needle biopsy cohort~\cite{xu2021bcnb} (BCNB-ER, BCNB-HER2, and BCNB-PR).
We also include two in-house datasets: Fuzhou-CRC-KRAS and Zhujiang-CRC-KRAS.
\textbf{BRCA-Molecular} is a five-class task using PAM50 subtypes (Basal-like, HER2-enriched, Luminal B, Luminal A, Normal-like); subtype labels are taken from TCGA molecular annotations. 
\textbf{BRCA-TP53} is a binary task that predicts TP53 mutation status from WSIs using mutation calls recorded in TCGA; non-synonymous somatic alterations are treated as positive. 
\textbf{BCNB-ER}, \textbf{BCNB-HER2} and \textbf{BCNB-PR} are three binary biomarker expression prediction tasks from the BCNB cohort, where each slide is labeled positive or negative for ER, HER2, and PR status, respectively.
\textbf{CRC-Molecular} uses TCGA colorectal cohorts and frames a four-class task with consensus molecular subtypes (CMS1, CMS2, CMS3, CMS4), restricting to cases with reliable subtype calls. 
\textbf{CRC-KRAS}, \textbf{Fuzhou-CRC-KRAS} and \textbf{Zhujiang-CRC-KRAS} are binary tasks for KRAS pathogenic mutation status.

\subsubsection*{Survival Analysis} 
Survival prediction from histopathology images is clinically meaningful, as it aids risk stratification and guides treatment planning.
For survival analysis, we use six TCGA cohorts: \textbf{TCGA-BLCA}, \textbf{TCGA-BRCA}, \textbf{TCGA-CRC}, \textbf{TCGA-KIRC}, \textbf{TCGA-KIRP}, and \textbf{TCGA-LUAD}. Each cohort contains WSIs annotated with patient survival outcomes. 
To ensure a robust evaluation and mitigate bias from data partitioning, we employ a 5-fold cross-validation scheme. 
Critically, the splits are performed at the patient level to prevent data leakage between training and validation sets. In each fold, the data is partitioned into training and validation subsets at a 4:1 ratio.

Collectively, these three task categories represent key challenges across the computational pathology pipeline, from molecular characterization to clinical prognosis.

\subsubsection*{Evaluation Metrics}
For all classification tasks (i.e., diagnostic and molecular), we report the Area Under the Receiver Operating Characteristic Curve (AUC), Accuracy (ACC), and F1-Score, along with their standard deviations (std) over the cross-validation folds. AUC and F1-Score offer robust assessments, particularly in the presence of class imbalance. For the survival analysis task, model performance is evaluated using the cross-validated Concordance Index (C-Index)~\cite{cindex} and its standard deviation.

\subsection{Implementation Details}
We use ImageNet-pretrained ResNet-50~\cite{he2016resnet}, PLIP~\cite{huang2023visual} and CONCH~\cite{lu2024conch} to extract features for each instance. 
To ensure a fair comparison, all models are trained for 30 epochs with a batch size of 1, optimized using Adam with a weight decay of $1 \times 10^{-5}$. 
We set the learning rate to $2 \times 10^{-4}$ for baseline models with a cosine annealing schedule, while all Mamba-based models share a learning rate of $2 \times 10^{-5}$, ensuring that each configuration yields optimal performance.
The hyperparameter $r$ in Equation~\ref{eq:cts} for each dataset is documented in our released code.
Reported results, including means and standard deviations, are averaged over 5 independent runs of 5-fold cross-validation.

\section{Results}

\begin{table}[ht]
\centering
\resizebox{0.8\linewidth}{!}{
\setlength{\tabcolsep}{3pt}
\begin{tabular}{l|c|cc|cc}
\toprule
\multirow{3}{*}{Method} & \multirow{3}{*}{\#Params} & \multicolumn{2}{c|}{N=1000} & \multicolumn{2}{c}{N=5000} \\
&  & Time & Thro. & Time & Thro. \\ 
& (K) & (ms) & (K) & (ms) & (K) \\ 
\midrule
MeanMIL                  & 525.8   & 0.25   & 4.02   & 0.53   & 1.90   \\ 
MaxMIL                   & 525.8   & 0.19   & 5.22   & 0.51   & 1.97   \\ 
CLAM~\cite{lu2021clam}   & 790.8   & 0.82   & 1.22   & 0.94   & 1.07   \\ 
ABMIL~\cite{ilse2018abmil} & 591.6 & 0.71   & 1.41  & 0.80   & 1.25      \\ 
MHIM-ABMIL~\cite{tang2023MHIM} & 1183.2 & 0.78   & 1.28 & 0.88 & 1.14      \\ 
DSMIL~\cite{li2021dsmil} & 872.7   & 1.48   & 0.68  & 1.95   & 0.51      \\ 
DTFD~\cite{zhang2022dtfd} & 789.3  & 9.81   & 0.10   & 13.86  & 0.07   \\ 
TransMIL~\cite{shao2021transmil} & 2672.2 & 9.47 & 0.11   & 11.40  & 0.09   \\ 
MHIM-TransMIL~\cite{tang2023MHIM} & 5344.3 & 10.51 & 0.10  & 12.10  & 0.08   \\ 
RRTMIL~\cite{tang2024rrtmil} & 2697.5 & 2.97  & 0.34   & 4.47   & 0.22   \\ 
PAM~\cite{huang2024pam} & 2290.4  & 3.09   & 0.32   & 6.47   & 0.16   \\ 
MambaMIL~\cite{yang2024mambamil} & 2410.2 & 2.87 & 0.35   & 4.96   & 0.20   \\ 
2DMambaMIL~\cite{zhang20252dmamba} & 2288.4 & 2.46 & 0.41   & 6.44   & 0.16   \\ 
\ModelName & 2683.3 & 2.44   & 0.36   & 10.01  & 0.11   \\ 
\bottomrule
\end{tabular}}
\caption{Parameters, inference time, and throughput for different MIL methods. The sequence length used for \ModelName is four times longer than that used for the other methods during inference.}
\label{tab:inference_time_throughput}
\end{table}

\subsection{Comparison with SOTA Methods}
We evaluated and benchmarked \ModelName on 20 computational pathology tasks, comparing it against eleven state-of-the-art MIL methods and two traditional pooling baselines.
The compared MIL methods include:
1) \textbf{CLAM}~\cite{lu2021clam}, an attention-based framework with instance-level clustering to improve interpretability;
2) \textbf{ABMIL}~\cite{ilse2018abmil}, a classical attention-based framework that learns instance importance weights;
3) \textbf{DSMIL}~\cite{li2021dsmil}, an attention-based dual-stream framework integrating both instance-level and bag-level representations; 
4) \textbf{DTFD}~\cite{zhang2022dtfd}, a double-tier framework that introduces pseudo-bags to virtually enlarge the number of training bags and better exploit intrinsic features with limited samples; 
5) \textbf{TransMIL}~\cite{shao2021transmil}, a Transformer-based framework that captures long-range contextual relations among patches; 
6) \textbf{MHIM-ABMIL}~\cite{tang2023MHIM} and 7) \textbf{MHIM-TransMIL}~\cite{tang2023MHIM} extend ABMIL and TransMIL with masked hard instance mining.
8) \textbf{RRTMIL}~\cite{tang2024rrtmil}, a Transformer-based framework that introduces a re-embedded regional Transformer (R$^2$T) to re-embed instance features online via regional and cross-region attention; 
9) \textbf{MambaMIL}~\cite{yang2024mambamil}, a Mamba-based framework that introduces sequence reordering Mamba (SR-Mamba) to capture order-aware dependencies for efficient long-sequence modeling; 
10) \textbf{PAM}~\cite{huang2024pam}, a Mamba-based framework that integrates a bidirectional Mamba encoder to efficiently model large-scale WSIs, with a test-time importance resampling module to mitigate feature distribution shifts between training and testing; 
and 11) \textbf{2DMambaMIL}~\cite{zhang20252dmamba}, a Mamba-based framework that extends the 1D selective state-space model into 2D to capture spatial continuity with high computational efficiency.
For fairness, all experiments are implemented using our codebase under consistent settings.
As shown in Fig.~\ref{fig:vis_results}, \ModelName consistently outperforms existing MIL methods across all feature extractors, with improvements that remain statistically significant and generally yield larger relative gains than all competing approaches.

\subsubsection*{Diagnostic Classification} 
Tab.~\ref{tab:comparison_subtyping} summarizes the diagnostic classification performance of MIL methods across the BRACS-7, BRCA-Subtyping, Camelyon, NSCLC, and UBC-OCEAN benchmarks.
Averaged over five diagnostic classification tasks, \ModelName achieves the best overall performance, outperforming all competing MIL approaches across every feature extractor with statistically significant gains (Fig.~\ref{fig:vis_results}, row 3). 
Specifically, under ResNet-50 features, \ModelName attains average relative improvements of 1.5\% AUC, 2.9\% ACC, and 3.0\% F1 over the second-best method. 
With PLIP, the gains remain notable, 0.9\%, 1.1\%, and 0.9\% for AUC, ACC, and F1, respectively. 
Using CONCH, our model further improves performance by 0.5\% AUC, 2.0\% ACC, and 1.1\% F1.
Overall, \ModelName ranks first on almost all diagnostic classification tasks, demonstrating strong diagnostic capability across diverse disease cohorts. 
Notably, this performance advantage holds consistently across different feature extractors, indicating that \ModelName is feature-agnostic rather than dependent on a specific feature extractor.

\subsubsection*{Molecular Prediction} 
Tab.~\ref{tab:comparison_molecular_1} and Tab.~\ref{tab:comparison_molecular_2} present detailed results on molecular prediction tasks.
Averaged across all nine datasets, \ModelName consistently surpasses existing MIL methods under all three feature extractors, achieving statistically significant gains (Fig.~\ref{fig:vis_results}, row 4).
Across molecular subtyping, gene mutation, and expression prediction tasks, \ModelName exhibits a noticeably larger margin over other methods. 
Specifically, compared with the second-best competitor, it improves overall performance in AUC by 1.4–3.4\% across different extractors.
What is also noteworthy is that, unlike diagnostic classification tasks, where performance trends remain relatively small, molecular prediction tasks show substantial variation across feature extractors, highlighting the importance of a strong foundation model for such challenging settings. 
However, even in this setting, \ModelName remains the top performer: it ranks first on eight of the nine tasks and delivers consistent improvements on the strongest feature extractor (e.g., CONCH), with absolute gains of 1.3\%, 2.5\%, and 2.7\% in AUC, ACC, and F1, respectively.
Overall, the evidence shows that \ModelName maintains strong generalization across diverse molecular targets, successfully linking tissue morphology to molecular variation.


\subsubsection*{Survival Analysis} 
Tab.~\ref{tab:comparison_survival} summarizes the results on six survival prediction tasks (KIRC, KIRP, LUAD, BRCA, BLCA, CRC).
The performance margin among methods is relatively small compared with diagnostic classification or molecular prediction, yet \ModelName still delivers consistently stable gains across all datasets (Fig.~\ref{fig:vis_results}, row 5).
Averaged over the six survival prediction tasks, \ModelName obtains mean C-Index scores of 0.656, 0.681, and 0.720 for ResNet-50, PLIP, and CONCH features, respectively.
In contrast, the classical ABMIL baseline shows lower performance, with corresponding mean C-Indices of 0.597, 0.659, and 0.706.
Overall, \ModelName yields an average 1.5\% improvement over the strongest competing method with ResNet-50 features, while still maintaining 0.6\% and 0.7\% gains under PLIP and CONCH, demonstrating its stability and robustness even in this more challenging evaluation setting.

\begin{figure*}
    \centering
    \includegraphics[width=0.8\linewidth]{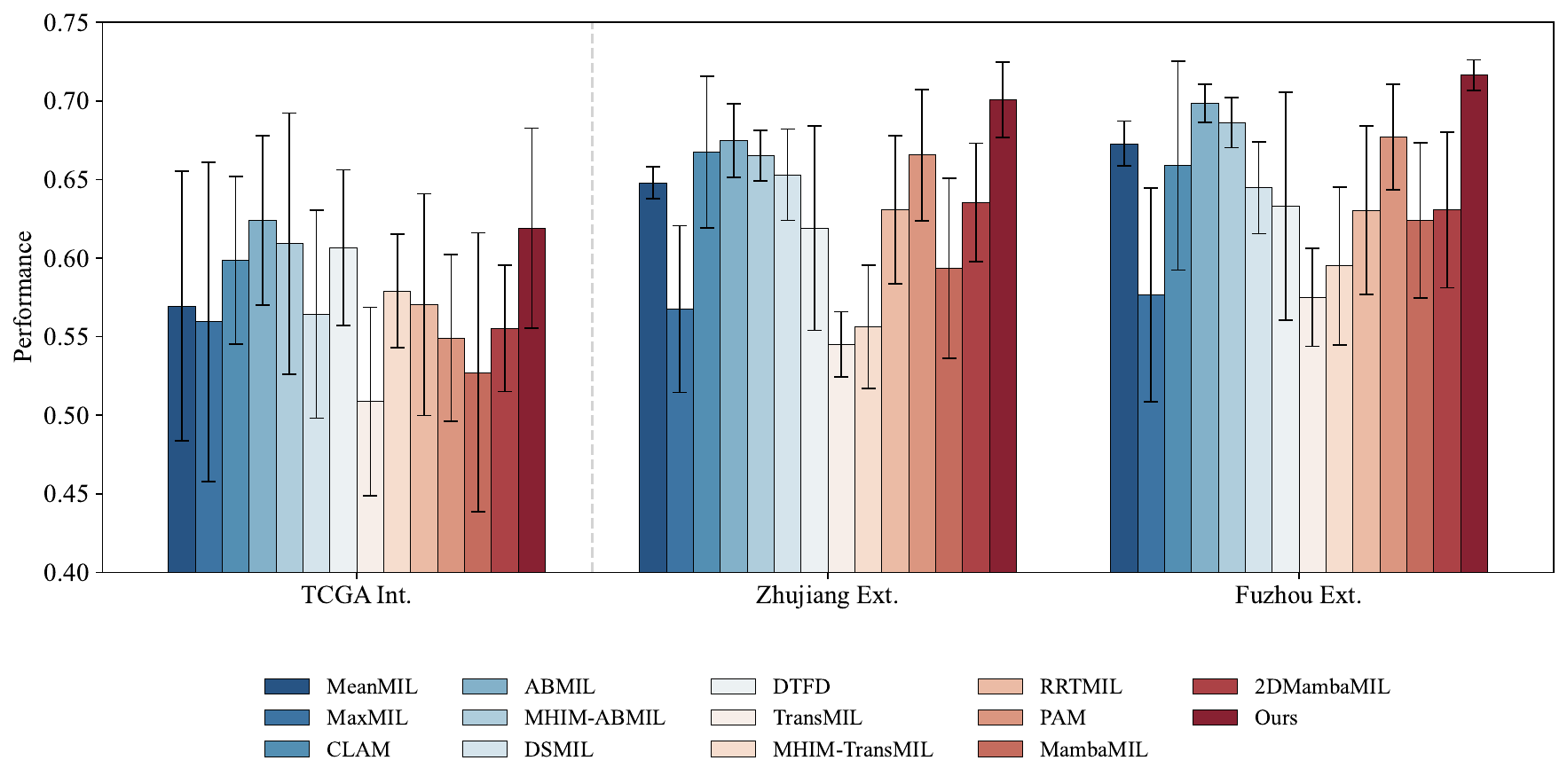}
    \caption{Performance comparison of \ModelName with other MIL methods on the CRC-KRAS mutation prediction tasks under the internal and external evaluation settings. Int., internal test set. Ext., external test set.}
    \label{fig:external_results}
\end{figure*}

\subsubsection*{External Evaluation}
In real-world applications, off-the-shelf models often fail to deploy effectively across clinical sites, largely due to distribution shifts arising from differences in fixation protocols, staining conditions, and whole-slide imaging pipelines.
Moreover, cross-cohort variation poses an even greater challenge for mutation prediction, as this task is inherently more challenging than routine diagnostic classification.
Therefore, to assess the generalizability of our method, we perform external evaluations on the two in-house gene mutation prediction datasets. 
Specifically, the model is trained on the public CRC-KRAS dataset from TCGA and tested on the Fuzhou-CRC-KRAS and Zhujiang-CRC-KRAS datasets, providing a strict external evaluation across independent cohorts from different institutions.
Considering the relatively limited performance of all models in this more challenging cross-cohort setting, we report results using CONCH as the feature extractor, which provides stronger overall baseline representation.
As shown in Fig.~\ref{fig:external_results}, our method achieves notably superior performance under this external evaluation protocol.
Models that perform well in internal evaluations, such as ABMIL and CLAM, exhibit substantial degradation when transferred to external datasets, whereas our approach maintains consistently high performance.
In particular, our method outperforms the second-best performing model by 1.8\% and 2.6\% in AUC on the Fuzhou and Zhujiang cohorts, respectively, underscoring its strong cross-domain generalizability and robustness.

From traditional feature extractors such as ResNet-50 to pathology foundation models like PLIP and CONCH, each generation has brought substantial gains in representation quality and downstream performance.
As these foundation models grow increasingly expressive however, the marginal improvements contributed by MIL aggregators naturally become smaller—an expected trend, given that much of the inter-instance representational gap is largely bridged by the foundation model itself.
Yet, even with highly capable foundation models, extracted features alone do not fully saturate performance. 
A well-designed MIL framework can still push the upper bound forward, unlocking fine-grained relational cues that are otherwise overlooked by single-stage encoders. 
While simple MIL models like ABMIL may perform reasonably well with strong feature extractors, their cross-cohort generalization remains limited, largely due to the two-stage paradigm in which feature learning and aggregation are decoupled. 
Without joint optimization, the resulting representations are not fully aligned with downstream objectives and therefore transfer poorly under distribution shift. 
This issue is even more pronounced in lightweight MIL methods, whose limited representational capacity leaves less room for refining features toward task-specific decision boundaries.
In contrast, a more carefully designed MIL aggregator, such as \ModelName, learns more resilient cross-cohort representations, yielding stronger robustness and transferability in real-world clinical settings.
In summary, the overall results highlight the strong versatility and robustness of \ModelName across a broad spectrum of WSI analysis tasks, providing compelling empirical evidence that Mamba can serve as a powerful and scalable backbone for computational pathology.

\begin{figure*}[ht]
    \centering
    \includegraphics[width=\linewidth]{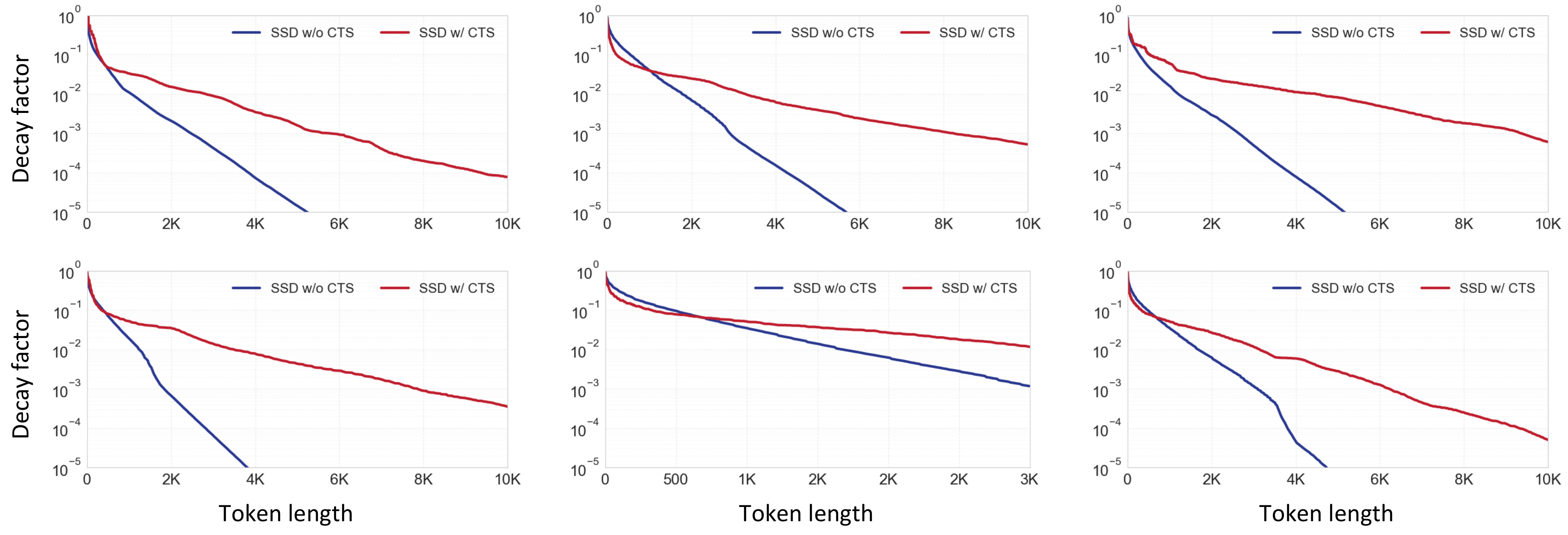}
    \caption{Visualization of memory decay on SSD with and without CTS across multiple cases from the Camelyon dataset. The zoomed-in regions highlight differences in long-range retention behaviors between the two models.}
    \label{fig:decay}
\end{figure*}

\subsection{Efficiency Analysis}
Efficiency is also an essential aspect in practical deployment.
We therefore evaluated the computational efficiency of \ModelName and compared it against all baselines. 
The results are summarized in Tab.~\ref{tab:inference_time_throughput}, reporting model parameters, inference time, and throughput. 
For fairness, the input sequence length of \ModelName is approximately four times that of other MIL methods due to its overlapping scanning design.
Under shorter to moderate input lengths (i.e., $N$=1,000), most MIL methods based on i.i.d. assumption require fewer parameters and exhibit fast inference with high throughput, but they come at the cost of limited ability to capture instance correlation. 
In contrast, Mamba-based MIL methods provide substantial efficiency gains over Transformer-based MIL approaches while still retaining the ability to model long-range correlations.
Despite processing sequences that are four times longer, \ModelName achieves lower inference time and higher throughput than Transformer-based MIL methods at $N$=1,000 (with \ModelName evaluated at $N$=4,000 for fairness).
As the sequence length increases (e.g., when it reaches 5,000, which corresponds to moderate- to long-range in WSI analysis), inference time naturally increases.
However, even at four times this length (i.e., $N$=20,000), \ModelName remains more efficient than conventional TransMIL and MHIM-TransMIL in both inference time and throughput.
Importantly, this efficiency is achieved alongside notable performance improvements, making the trade-off introduced by longer input sequences both justifiable and advantageous.

\subsection{Ablation Study}
We conducted ablation studies to assess the contribution of each proposed component.
As shown in Tab.~\ref{tab:ablation_all}, the joint combination of overlapping scanning, CTS, and S$^2$PE yields consistent gains across target domains, outperforming all partial configurations as well as the base model.
On average, relative to the base model without any components, our method achieves an improvement of +1.9\% in AUC, +3.6\% in ACC, and +3.6\% in F1 score on the diagnostic classification benchmark. 
This demonstrates that overlapping scanning, CTS, and S$^2$PE complement each other effectively, collectively enabling robust and stable slide-level prediction.

\begin{table*}[htbp]
\centering
\resizebox{\linewidth}{!}{
\setlength{\tabcolsep}{3pt}
\begin{tabular}{@{}ccc|*{3}{c}|*{3}{c}|*{3}{c}|*{3}{c}|*{3}{c}|*{3}{c}}
\toprule
\multirow{2}{*}{$X$} & \multirow{2}{*}{$Y$} & \multirow{2}{*}{$Z$} & \multicolumn{3}{c}{BRACS-7} & \multicolumn{3}{|c}{BRCA-Subtyping} & \multicolumn{3}{|c}{Camelyon} & \multicolumn{3}{|c}{NSCLC} & \multicolumn{3}{|c}{UBC-OCEAN} & \multicolumn{3}{|c}{AVERAGE} \\
&&& AUC & ACC & F1 & AUC & ACC & F1 & AUC & ACC & F1 & AUC & ACC & F1 & AUC & ACC & F1 & AUC & ACC & F1 \\
\midrule
\xmark & \xmark & \xmark & 78.8$_{4.3}$ & 38.1$_{7.9}$ & 36.0$_{8.6}$ & 84.8$_{5.5}$ & 69.8$_{5.6}$ & 72.2$_{4.2}$ & 89.6$_{1.0}$ & 84.9$_{2.4}$ & 85.5$_{2.3}$ &  94.4$_{0.8}$ & 86.6$_{1.7}$ & 86.5$_{1.8}$ & 91.7$_{1.7}$ & 69.7$_{3.1}$ & 70.6$_{3.2}$ & 87.9 & 69.9 & 70.2 \\
\cmark & \xmark & \xmark & 79.1$_{5.1}$ & 38.9$_{8.8}$ & 37.3$_{9.7}$ & 89.2$_{2.9}$ &  \sbest{77.1$_{3.7}$} & 78.7$_{2.4}$ & 89.0$_{1.5}$ & 84.0$_{1.3}$ & 84.6$_{1.8}$ & 93.8$_{1.4}$ & 86.8$_{0.9}$ & 86.8$_{1.0}$ & 92.4$_{2.0}$ & 69.3$_{4.8}$ & 70.0$_{4.8}$ & 88.7 & 71.1 & 71.5 \\
\xmark & \cmark & \xmark &  \sbest{81.5$_{2.2}$} & 41.1$_{6.1}$ & 39.9$_{6.7}$ & 85.2$_{2.6}$ & 73.6$_{4.2}$ & 74.2$_{5.0}$ &  \best{90.8$_{2.5}$} &  \sbest{86.5$_{3.2}$} &  \sbest{86.2$_{3.3}$} &  \best{94.7$_{0.9}$} & 86.2$_{3.2}$ & 85.8$_{3.5}$ & 92.5$_{2.0}$ & 69.5$_{2.7}$ & 71.0$_{2.8}$ & 89.0 & 71.3 & 71.5 \\
\xmark & \xmark & \cmark & 80.5$_{3.1}$ & 41.3$_{7.8}$ & 40.2$_{9.0}$ & 85.5$_{2.8}$ & 72.5$_{4.4}$ & 74.5$_{4.0}$ & 89.3$_{1.8}$ & 83.3$_{3.2}$ & 83.8$_{4.0}$ & 94.4$_{0.9}$ & 86.7$_{1.1}$ & 87.1$_{1.0}$ & 91.8$_{2.0}$ & 67.9$_{4.0}$ & 69.4$_{4.5}$ & 88.3 & 70.5 & 71.1 \\
\cmark & \cmark & \xmark & 81.1$_{3.1}$ &  \sbest{42.3$_{5.9}$} &  \sbest{40.3$_{6.9}$} &  \sbest{89.4$_{1.8}$} & 76.5$_{4.4}$ &  \sbest{79.1$_{3.4}$} &  89.4$_{2.7}$ & 84.8$_{3.4}$ & 86.2$_{3.0}$ & 94.0$_{1.3}$ & \sbest{86.9$_{2.9}$} & \sbest{87.0$_{3.0}$}  &  \sbest{92.6$_{2.0}$} &  \best{70.2$_{6.5}$} &  \best{71.3$_{7.3}$} &  \sbest{89.3} &  \sbest{72.4} &  \sbest{73.0} \\
\cmark & \cmark & \cmark &  \best{81.5$_{3.0}$} &  \best{43.2$_{4.9}$} &  \best{42.5$_{5.9}$} &  \best{90.0$_{3.3}$} &  \best{80.5$_{3.9}$} &  \best{81.8$_{4.0}$} &  \sbest{90.3$_{0.9}$} &  \best{86.7$_{2.0}$} &  \best{86.9$_{1.8}$} &  \sbest{94.6$_{1.4}$} & \best{87.3$_{1.5}$} & \best{87.2$_{1.5}$} &  \best{93.2$_{1.9}$} &  \sbest{69.7$_{3.6}$} &  \sbest{71.0$_{3.3}$} &  \best{89.9} &  \best{73.5} &  \best{73.8} \\
\bottomrule
\end{tabular}}
\caption{Ablation studies conducted on diagnostic classification datasets using pretrained ResNet-50 as the feature extractor. The values highlighted in \best{red} and \sbest{blue} denote the best and second-best performances, respectively. $X$: overlapping scanning; $Y$: contextual token selection; $Z$: selective stripe position encoder.}
\label{tab:ablation_all}
\end{table*}

\begin{figure}
    \centering
    \includegraphics[width=\linewidth]{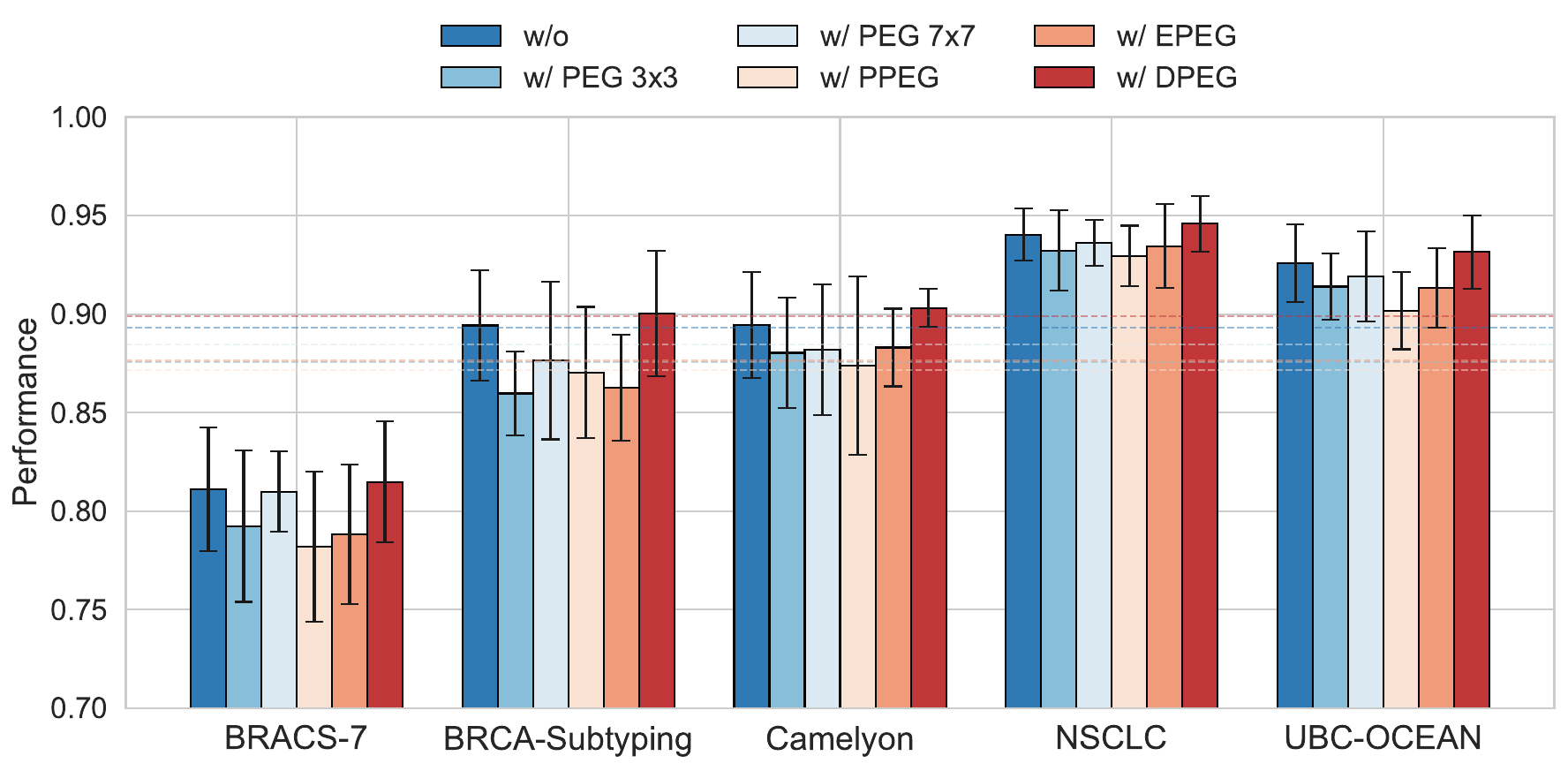}
    \caption{Ablation studies on different position encoders across diagnostic classification datasets using pretrained ResNet-50 as the feature extractor.}
    \label{fig:ablation_pe}
\end{figure}

\begin{table*}[ht]
\centering
\resizebox{\linewidth}{!}{
\setlength{\tabcolsep}{3pt}
\begin{tabular}{@{}l|*{3}{c}|*{3}{c}|*{3}{c}|*{3}{c}|*{3}{c}|*{3}{c}@{}}
\toprule
Method & \multicolumn{3}{c}{BRACS-7} & \multicolumn{3}{c}{BRCA-Subtyping} & \multicolumn{3}{c}{Camelyon} & \multicolumn{3}{c}{NSCLC} & \multicolumn{3}{c}{UBC-OCEAN} & \multicolumn{3}{c}{AVERAGE} \\
\midrule
Metric & AUC & ACC & F1 & AUC & ACC & F1 & AUC & ACC & F1 & AUC & ACC & F1 & AUC & ACC & F1 & AUC & ACC & F1 \\
\midrule
w/ CTS-R
& 80.5$_{3.4}$ & \sbest{41.7$_{8.2}$} & 40.3$_{9.8}$ 
& 89.1$_{3.1}$ & 77.5$_{3.2}$ & 79.8$_{3.2}$ 
& 88.6$_{3.1}$ & 83.7$_{3.4}$ & 83.7$_{4.6}$ 
& 94.3$_{1.6}$ & 86.4$_{1.3}$ & 86.4$_{1.4}$ 
& 92.6$_{1.9}$ & 69.0$_{7.5}$ & 70.3$_{7.4}$ 
& 89.0 & 71.6 & 72.1 \\
w/ CTS-U
& 80.8$_{3.6}$ & 41.4$_{7.7}$ & \sbest{40.6$_{9.3}$} 
& 89.7$_{3.5}$ & 76.3$_{6.2}$ & 78.9$_{5.2}$ 
& \sbest{88.8$_{3.3}$} & 83.9$_{2.9}$ & 84.6$_{3.0}$ 
& 94.4$_{1.7}$ & 87.2$_{1.8}$ & 87.2$_{1.8}$ 
& 92.7$_{1.5}$ & 69.5$_{3.6}$ & 70.2$_{3.0}$ 
& 89.3 & 71.7 & 72.3 \\
w/ CTS-A
& 81.1$_{3.7}$ & 38.5$_{7.3}$ & 37.3$_{8.7}$
& \sbest{89.9$_{3.8}$} & 76.8$_{5.2}$ & 79.7$_{4.7}$ 
& 88.6$_{4.0}$ & 83.1$_{4.2}$ & 82.8$_{5.2}$ 
& \sbest{94.5$_{1.9}$} & \best{87.4$_{2.2}$} & \sbest{87.3$_{2.2}$} 
& 92.4$_{1.4}$ & \sbest{69.9$_{6.5}$} & 70.6$_{5.9}$ 
& \sbest{89.3} & 71.1 & 71.5 \\
w/ CTS-$\Delta$ 
& \sbest{81.1$_{4.1}$} & 38.5$_{8.4}$ & 37.2$_{9.5}$ 
& 89.6$_{2.8}$ & \sbest{79.1$_{3.9}$} & \sbest{81.0$_{3.8}$} 
& 88.2$_{2.5}$ & \sbest{84.9$_{2.3}$} & \sbest{86.1$_{2.3}$} 
& 93.9$_{0.9}$ & 87.1$_{1.9}$ & 87.1$_{2.0}$ 
& 92.6$_{1.8}$ & \best{69.9$_{7.3}$} & \best{71.6$_{6.4}$} 
& 89.1 & \sbest{71.9} & \sbest{72.6}\\
w/ CTS-P & 79.7$_{3.2}$ & 39.0$_{5.6}$ & 37.2$_{6.2}$ 
& 89.8$_{4.6}$ & 76.9$_{7.5}$ & 78.6$_{6.8}$ 
& 77.5$_{4.2}$ & 74.8$_{6.2}$ & 75.7$_{7.6}$ 
& 93.7$_{2.4}$ & 86.1$_{3.5}$ & 86.1$_{3.5}$ 
& \sbest{93.1$_{1.1}$} & 69.2$_{3.6}$ & 70.7$_{4.6}$ 
& 86.8 & 69.2 & 69.7 \\
w/ CTS 
& \best{81.5$_{3.1}$} & \best{43.2$_{4.9}$} & \best{42.5$_{5.9}$} 
& \best{90.0$_{3.3}$} & \best{80.5$_{3.9}$} & \best{81.8$_{4.0}$} 
& \best{90.3$_{1.0}$} & \best{86.7$_{2.0}$} & \best{86.9$_{1.8}$} 
& \best{94.6$_{1.4}$} & \sbest{87.3$_{1.5}$} & \best{87.3$_{1.5}$} 
& \best{93.2$_{1.9}$} & 69.7$_{3.6}$ & \sbest{71.0$_{3.3}$} 
& \best{89.9} & \best{73.5} & \best{73.8} \\
\bottomrule
\end{tabular}}
\caption{Comparison results of different variants of the contextual token selection mechanism across diagnostic classification datasets using pretrained ResNet-50 as the feature extractor.}
\label{tab:ablation_cts}
\end{table*}

\begin{figure}[ht]
    \centering
    \includegraphics[width=0.65\linewidth]{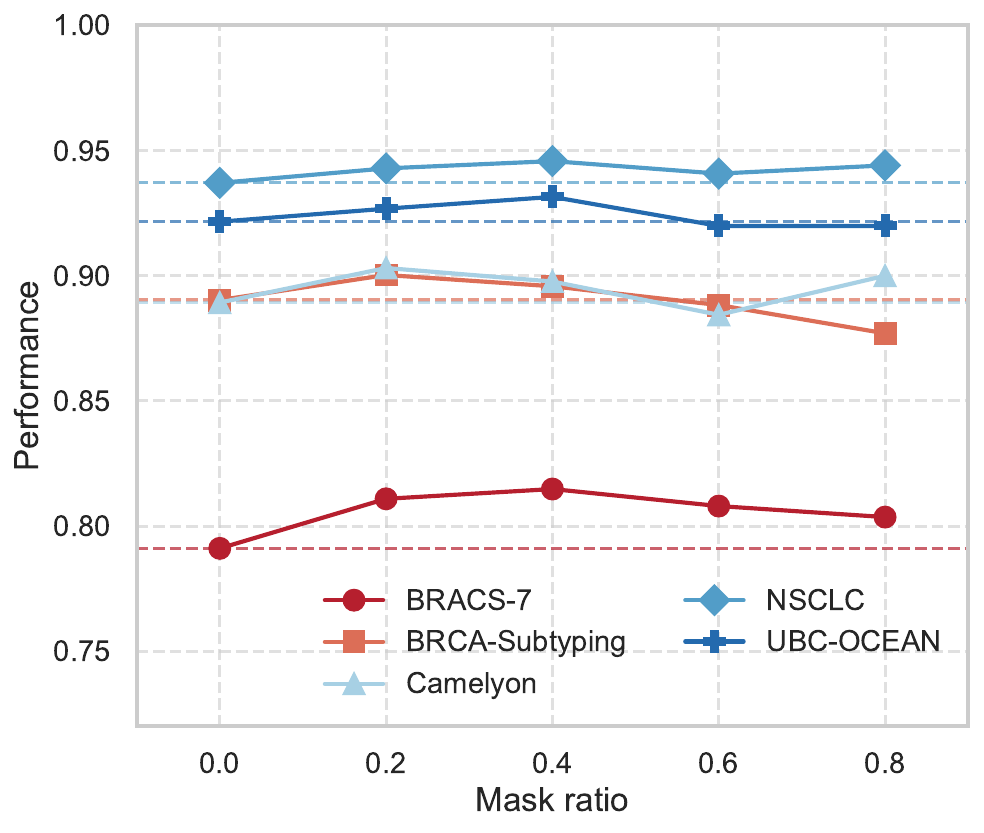}
    \caption{Effect of the mask ratio $r$ across diagnostic classification datasets using pretrained ResNet-50 as the feature extractor.}
    \label{fig:ablation_ratio}
\end{figure}

\begin{figure*}[ht]
    \centering
    \includegraphics[width=\linewidth]{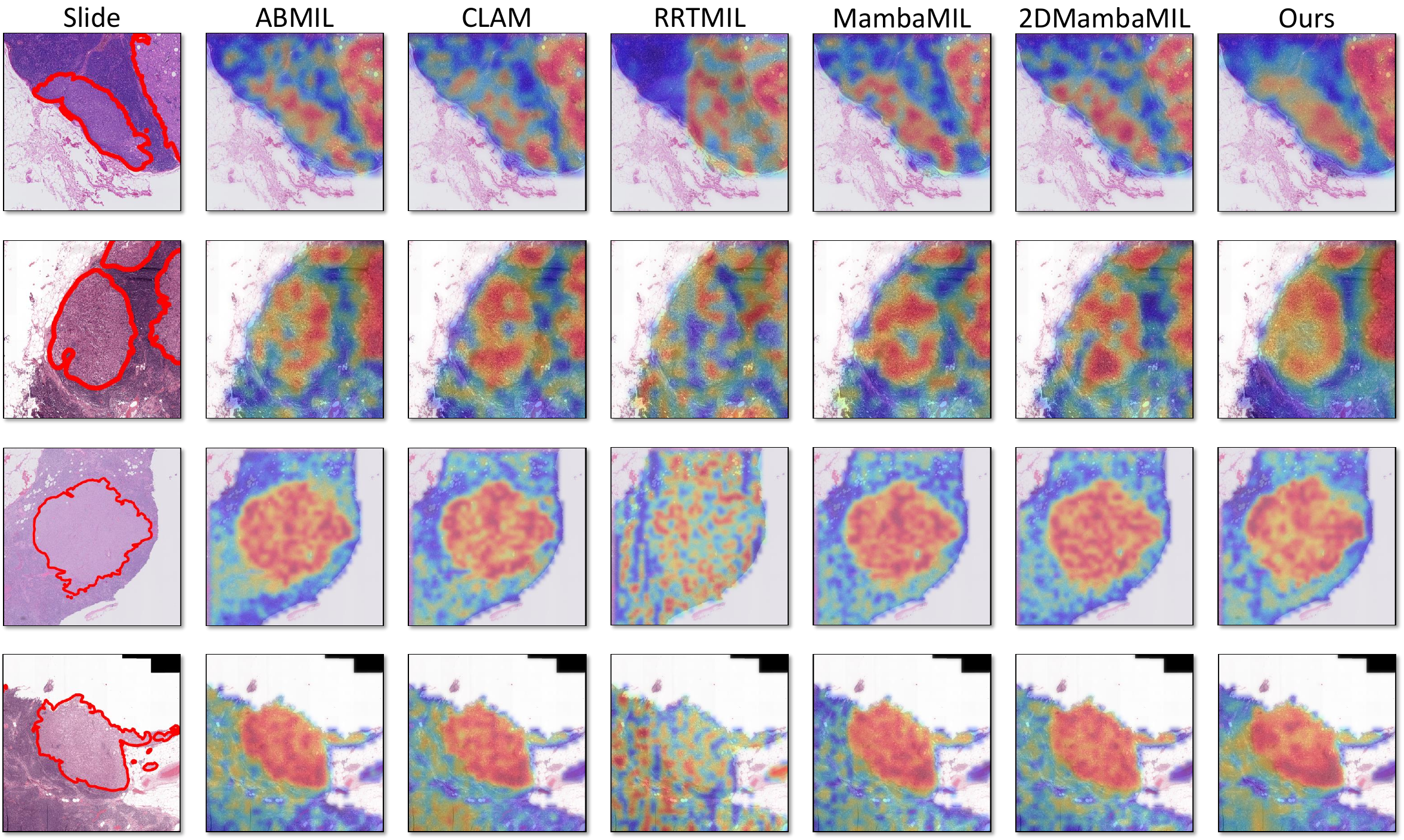}
    \caption{The attention visualization of \ModelName and five other methods on four Camelyon sample for diagnostic classification.
    }
    \label{fig:CAM_Attn_Visualization}
\end{figure*}

\subsubsection*{Further Analysis on $\text{S}^{2}\text{PE}$}
To further evaluate the effectiveness of our proposed positional encoding, we compared $\text{S}^{2}\text{PE}$ with several representative alternatives, including PEG~\cite{chu2021peg}, PPEG~\cite{shao2021transmil}, and EPEG~\cite{tang2024rrtmil}, as illustrated in Fig.~\ref{fig:ablation_pe}.
Experimental results show that although these positional encoders improve Transformer-based MIL models, they cause a noticeable performance decline when integrated into Mamba, with an average drop of 0.8–2.1\% in AUC across five diagnostic classification tasks compared to the baseline without any positional encoding. 
This behavior is expected, as the sequential modeling nature of Mamba already encodes positional dependencies, rendering external 2D positional priors largely redundant. 
Furthermore, conventional conditional encoding may amplify such redundancy and even introduce feature ambiguity, ultimately weakening discriminative power.
In contrast,  $\text{S}^{2}\text{PE}$ yields clear and steady improvements (+0.7\% AUC, +1.4\% ACC, +1.1\% F1; $P=0.0147<0.05$), demonstrating its compatibility with Mamba.

\subsubsection*{Further Analysis on CTS}
To validate the effectiveness of CTS, we attempted to visualize the memory decay phenomenon in the hidden states of Mamba. 
As illustrated in Fig.~\ref{fig:decay}, we plot the decay factor of hidden states with respect to local tokens on the Camelyon dataset, which can be interpreted as the influence of the first token on subsequent tokens along a sequence of a certain length. 
It is noted that in Mamba with no CTS, the decay factor drops by several orders of magnitude after only a few hundred tokens, which is consistent with our analysis of exponential memory decay. 
With the incorporation of CTS, however, the memory decay issue is substantially alleviated. 
More importantly, this alleviation does not compromise the utilization of contextual tokens; instead, CTS preserves long-range dependencies while mitigating the risk of rapid information fading, thereby ensuring richer and more stable long-term context modeling.

We further compare the CTS with five variants, as shown in Tab.~\ref{tab:ablation_cts}:
\begin{itemize}
    \item CTS-R: a random variant that selects contextual tokens randomly without considering spatial or semantic information.
    \item CTS-U: a uniform variant that selects contextual tokens evenly across spatial positions to ensure balanced coverage of the WSI.
    \item CTS-A: an attention-driven variant in which an attention branch ranks patches by learned attention scores; the top-ranked patches are chosen as contextual tokens. This approach relies purely on learnable attention weights rather than supervised instance predictions.
    \item CTS-$\Delta$: a heuristic variant that selects the top-ranked patches with the largest $\Delta t$ values as contextual tokens, since $\Delta t$ in Mamba can, to some extent, be interpreted as a form of gating strength.
    \item CTS-P: a simple variant that directly prunes tokens before feeding them into the Mamba blocks, instead of eliminating them within the Mamba memory.
\end{itemize}

As shown in Tab.~\ref{tab:ablation_cts}, our CTS mechanism consistently outperforms all comparison variants across five diagnostic classification tasks. 
Notably, the CTS-prune variant exhibits a clear performance drop of 3.1\% on average relative to the best configuration, as removing tokens prior to sequential modeling disrupts the temporal-state representations that Mamba relies on.
In contrast, CTS not only achieves higher average scores in AUC, ACC, and F1 (89.9\%, 73.4\%, and 73.8\%, respectively), but also does so with substantially lower performance variance, reflecting improved stability and robustness.
We attribute these gains to two factors: 
1) the supervised auxiliary branch provides explicit instance-level supervision that helps distinguish truly informative patches from noisy ones, and 
2) entropy-based filtering preferentially selects low-uncertainty tokens, reducing the chance of including misleading context. 
In contrast, both CTS-attn and CTS-$\Delta t$ do not incorporate supervision and can be misled by spurious signals, which explains their inferior average performance and larger variability.

\noindent \textbf{Hyperparameter analysis.}
We further analyzed the impact of the mask ratio $r$ in CTS on model performance.
As shown in Fig.~\ref{fig:ablation_ratio}, we report the relative improvements in AUC and F1 score when adopting different evenly spaced mask ratios on the BRACS-7, NSCLC, and UBC-OCEAN datasets. 
It is evident that when CTS is not applied, the model performs the worst across all metrics.
Surprisingly, introducing a small proportion of masked tokens leads to a substantial improvement. 
However, as the mask ratio continues to increase, the performance gain becomes limited and tends to plateau. 
This phenomenon may be attributed to the limited capability of the instance learner, which can mistakenly mask out certain key tokens that are critical for accurate prediction.

\noindent \textbf{How local channels affect CTS.}
It is also worth noting that, in the context of language modeling, representation channels in Mamba always follow a dual-role pattern: local channels are expected to retain localized context, whereas global channels should maintain long-range dependencies for as long as possible~\cite{ben2024decimamba,ye2025longmamba}.
An intriguing question is how this principle transfers to the WSIs. 
Following Eq.~\ref{eq:recursive_mamba_eq}, the contribution of the earlier token to the latter one can be naturally used as a locality indicator for each channel, expressed as:
\begin{equation}
\alpha_{i,j} = C_j \left( \prod_{k=j+1}^i \bar{A}_k \right) \bar{B}_j,
\label{eq:mamba_attention}
\end{equation}
To explore how localized channels interact with CTS, we quantify locality using $\alpha_{4N-1,0}$ and select the top-$K$ most localized channels accordingly.
Interestingly, the results across three challenging diagnostic classification tasks summarized in Tab.~\ref{tab:local_channel} diverge from the linguistic expectation: 
preserving these localized channels without enlarging the memory window yields minimal improvement, and in many cases, even impairs performance.
We attribute this behavior to a WSI-specific characteristic: instances within a slide often exhibit high redundancy and strong visual similarity, causing models to over-prioritize local regions while failing to integrate broader contextual signals.
This fundamentally contrasts with language modeling, where tokens naturally differ in semantic importance and a balance between local and global channels emerges without explicit intervention.
\begin{table}[ht]
\centering
\caption{Model performance under preserving different number of local channels.}
\label{tab:local_channel}
\resizebox{0.85\linewidth}{!}{
\setlength{\tabcolsep}{3pt}
\begin{tabular}{cccccccccccc}
\toprule
\multirow{2}{*}{$K$} 
& \multicolumn{3}{c}{BRACS-7} 
&& \multicolumn{3}{c}{BRCA-Subtyping}
&& \multicolumn{3}{c}{Camelyon} \\
& AUC & ACC & F1 && AUC & ACC & F1 && AUC & ACC & F1 \\
\cline{2-4} \cline{6-8} \cline{10-12}
0 & 81.5 & 43.2 & 42.5 && 90.0 & 80.5 & 81.8 && 90.3 & 86.7 & 86.9 \\
1 & 80.7 & 41.5 & 40.8 && 90.0 & 79.8 & 81.4 && 90.5 & 85.6 & 86.0 \\
2 & 80.8 & 41.3 & 40.9 && 90.1 & 80.9 & 81.8 && 88.4 & 82.7 & 83.6 \\
4 & 80.5 & 38.5 & 37.2 && 90.1 & 80.1 & 81.5 && 89.6 & 83.3 & 84.5 \\
8 & 80.0 & 39.7 & 38.6 && 90.2 & 81.8 & 82.6 && 90.5 & 85.9 & 86.7 \\
\bottomrule
\end{tabular}}
\vspace{-10pt}
\end{table}

\subsection{Visualization}
As illustrated in Fig.~\ref{fig:CAM_Attn_Visualization}, we qualitatively compare and visualize the attention map generated by ABMIL~\cite{ilse2018abmil}, CLAM~\cite{lu2021clam}, RRTMIL~\cite{tang2024rrtmil}, MambaMIL~\cite{yang2024mambamil}, 2DMambaMIL~\cite{zhang20252dmamba} and \ModelName.
From top to bottom, the corresponding slide IDs are \texttt{patient\_051\_node\_2}, \texttt{patient\_073\_node\_1}, \texttt{patient\_075\_node\_4}, and \texttt{patient\_092\_node\_1} from the Camelyon17 dataset.
The visualization results demonstrate the superiority of \ModelName in localizing malignant regions, as its attention heatmaps exhibit significantly sharper and more definitive boundaries around the tumor areas compared to the other MIL methods. 
Specifically, RRTMIL generates nearly uniformly distributed attention scores, failing to effectively differentiate tumor regions from the background.
Other models, including ABMIL, MambaMIL, and 2DMambaMIL, do identify the tumor areas but struggle with precise delineation, often assigning attention to surrounding normal tissue, which results in ambiguous and blurred boundaries. 
In contrast, \ModelName consistently focuses its highest attention exclusively on the tumor regions while assigning lower attention to non-tumor areas. 
This result provides strong qualitative evidence that \ModelName has successfully learned robust, discriminative tumor features, enabling it to accurately localize and delineate malignant regions from normal tissue.

\section{Conclusion}

In this work, we propose \ModelName, a novel Mamba-based MIL framework designed to model ultra-long WSI sequences with explicit spatial awareness and enhanced contextual memory capabilities.
To better understand the role of spatial context in WSIs, we explore the significance of spatial correlations in WSI modeling.
Building on our observation, \ModelName adopts the overlapping scanning scheme to incorporate spatial context while maintaining linear complexity.
To further enhance positional cues, \ModelName integrates a selective stripe position encoder for improved spatial encoding.
Given the exponential memory decay problem formalized in our work, \ModelName introduces the contextual token selection mechanism to expand the memory capacity for more effective contextual modeling.
Extensive experiments across 20 benchmark datasets, 11 SOTA methods, and three feature extractors demonstrate that \ModelName consistently outperforms existing methods, validating its effectiveness for large-scale computational pathology and establishing Mamba as a powerful backbone for WSI analysis.





\bibliographystyle{ieeetr}
\end{document}